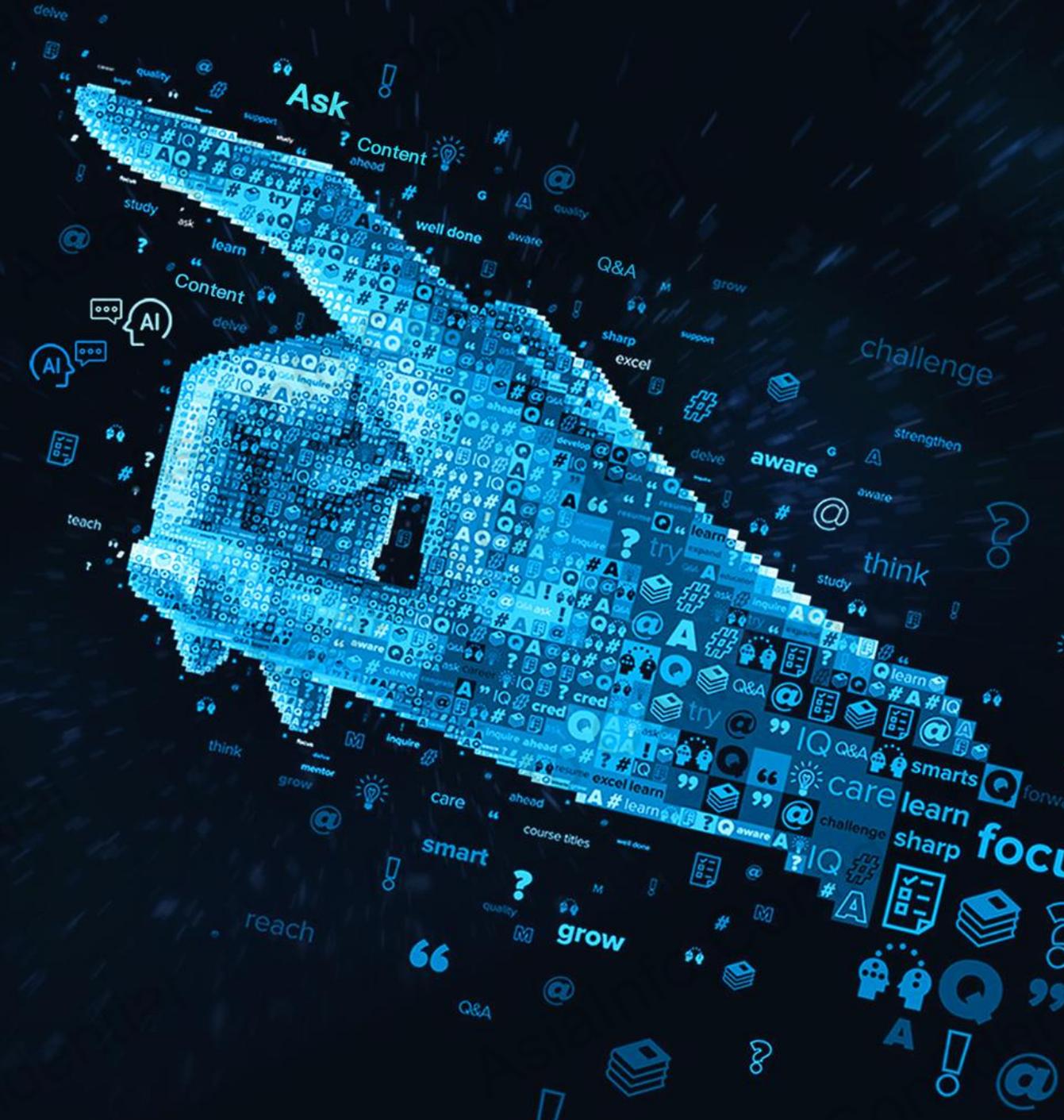

# AIGC（GPT）赋能通信行业应用白皮书 2.0

## A White Paper of AIGC (GPT) Empowering Telecom Sector V2.0

联合发布：

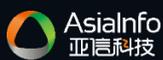 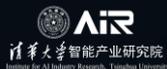 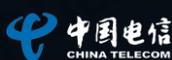 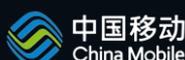 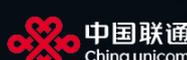 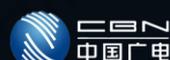 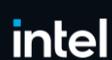

## 联合作者

亚信科技

清华大学智能产业研究院

中国电信研究院

中国移动信息技术中心

中国移动研究院

中国联通软件研究院

中广电移动网络有限公司

英特尔（中国）有限公司

## 引用本白皮书

《AIGC（GPT）赋能通信行业应用 2.0》，欧阳晔，张亚勤，叶晓舟，刘云新，宋勇，刘洋，边森，刘志勇等，2023 年 7 月。

版本说明：《AIGC（GPT）赋能通信行业应用 2.0》在第一版白皮书的基础上进行了修订，基于业务实践重点更新了构建通信业务 TelcoGPT 的相关内容。

# 目录

# 图目录

# 一. 前言

2022 年被认为是人工智能生成内容（AIGC）元年。作为 AIGC 在自然语言领域的代表，ChatGPT 在 2022 年年底一经推出，就掀起了一场可能涉及所有人和所有行业的"大火"，2023 年 3 月 GPT-4 的发布则进一步推动了"态势升级"。由 ChatGPT/GPT-4 引发的全球关注，令许多人回忆起 2016 年 AlphaGo 战胜人类围棋世界冠军的时刻。如果说 AlphaGo 代表了 AI 在专业领域战胜人类的起点，ChatGPT/GPT-4 似乎迈出了通用人工智能的第一步。这是第三次 AI 浪潮以来所有积累产生的硕果，AI 技术到了一个即将大规模产业化的临界点。

ChatGPT 将单调呆板的人机通信演进到前所未有的自然、高效、有创造力的人机协作，代表了生产力的提升。ChatGPT 有多重要？埃隆·马斯克评价："好得吓人，我们离危险的强人工智能不远了"。黄仁勋评价："这是 AI 界的 iPhone 时刻"。比尔.盖茨表示"和个人电脑和互联网的出现一样重要。是当前最重要的创新，将改变世界"。在 ChatGPT 的基础上，GPT-4 进一步在各种专业和学术基准测试中表现出"人类水平"的性能，在事实性、可引导性和可控制方面取得了"史上最佳结果"。当然，GPT-4 在不是无所不能的，仍具有与早期 GPT 模型相似的局限性，如：犯常识性的错误、缺乏对新世界知识的了解、存在社会偏见、产生幻觉、推理错误等。但是我们看到其技术背后却具有一直优化、学习和进步的能力。

在 GPT 大模型席卷全球的热潮中，人们已经深刻认识到人工智能作为经济社会发展中一项变革性技术与关键性力量，将为全球产业带来的巨大飞跃和突破式发展，深刻影响未来世界竞争格局。通信行业作为信息通信基础设施的建设者和运营者，既为 AI 的发展提供基础设施支撑，又是 AI 应用落地的领先者。AIGC

（GPT）如何赋能通信行业应用，通信行业如何落地 AIGC，这是通信业者必须要思考和回答的问题。

本白皮书通过对 AIGC 的典型代表 GPT 大模型的研究，以场景化的形式对 GPT 大模型如何赋能通信行业进行了分析，探讨了当前通用大模型与通信业务之间存在的鸿沟，首次提出了通信认知增强能力体系，针对通信行业如何构建通信业务 GPT 进行了解答并开展多方面实践。期待行业同仁围绕通信和人工智能协同创新，构建开放共享的创新生态，促进人工智能与通信产业的深度融合，加速构建下一代信息基础设施，助力经济社会数字化转型。

# 二. GPT 大模型研究

## 2.1 AIGC 与 GPT

AIGC 是继专业生成内容（PGC）和用户生成内容（UGC）之后，利用人工智能技术自动生成内容的新型生产方式。AIGC 根据其内容模态不同可分为文本、图像、音频、视频与跨模态生成。文本方面，例如文本创作、代码生成、问答对话等；图像方面，例如图片编辑、图片生成、3D 图像生成等；音频方面，例如文本合成语音、语音克隆、音乐生成等；视频方面，例如视频画质增强、视频内容创作、视频风格迁移等；跨模态方面，如文字生成图片、文字合成视频、图像描述等，而且在不同内容模态的技术应用场景也有着各自的细分品类。

ChatGPT 是 AIGC 发展的第一个丰碑。ChatGPT 是由人工智能研究公司 OpenAI 在 2022 年 11 月发布的一个对话型大语言模型，是人工智能技术驱动的自然语言处理工具和应用。ChatGPT 的全称是 Chat Generative Pre-trained Transformer，顾名思义，就是以 Transformer 为基础架构，采用预训练和生成式方式构建的面向对话的大语言模型，是 AIGC 在文本方面的典型代表。ChatGPT 的主要用途是生成对话，它能够通过学习和理解人类的语言来进行对话，根据聊天的上下文进行自然、流畅的互动，还能完成邮件撰写、文案编写、文本翻译、代码生成等任务。

ChatGPT 提供了前所未有的高效、自然的人机交互体验和极富创造力的内容生成能力，成为了 AI 时代的第一个"杀手级"应用。以 ChatGPT 为代表的生成式 AI 工具将使机器可以大规模参与知识类工作和创造性工作，极大提升生产力，涉及数十亿人方方面面的工作，可能产生数万亿美元的经济价值。

ChatGPT 覆盖了 NLP 所有领域，其所代表的大规模预训练语言模型（LLM）或基础模型已成为工业界和学术界最为关注的研究热点，并引领近期自然语言处理（NLP）乃至人工智能领域的研究范式的转变，对人工智能的技术发展可能产生重大影响。

相距 ChatGPT 发布仅仅 4 个月，OpenAI 在 2023 年 3 月正式发布了多模态预训练大模型 GPT-4。GPT-4 支持图片和文本输入并生成文本输出，相比 ChatGPT 增加了识图能力，并能够生成歌词、创意文本、学习用户的写作风格等，更具创造力和协作性。GPT-4 的输入限制提升至 2.5 万字，处理能力是 ChatGPT 的八倍，可用于长篇内容创作、扩展对话以及文档搜索和分析等应用场景，并可以用所有流行的编程语言写代码。 GPT-4 的回答准确性大幅提高，性能优于现有的大型语言模型、以及当前最先进（SOTA，State Of The Arts）模型，虽然在许多现实场景中不太聪明，但在各种专业和学术基准上表现出了人类水平。

ChatGPT 本身存在一定局限，如逻辑处理能力不够强，对上下文的理解仍然有限，多轮对话可能失控，缺乏一些通识能力，可能会出现创造不存在的知识，或者主观猜测提问者的意图，对 2021 年后的世界和事件了解有限等等。GPT-4 相较于 ChatGPT，不合规内容的响应请求降低了 82%，响应准确度提高了 40%，具备了更强的理解能力。但是 GPT-4 仍然具有与早期 GPT 模型相似的局限性，如：犯常识性的错误、缺乏对新世界知识的了解、存在社会偏见、产生幻觉、推理错误等。为解决这些问题，ChatGPT 引入插件（Plugins）功能让其升级为一个功能丰富的多元化平台。通过支持第三方开发者为 ChatGPT 开发的丰富的插件，用户可以根据需求挑选自己喜欢的插件，实现更专业、更个性化的体验。截至目前，ChatGPT 已提供超过 60 个插件，覆盖了网购商品搜索、优惠券搜索、

比价，旅游订机票、订酒店、租车、订餐，以及数学计算、股票分析，联网搜索、网页内容浏览和总结、文生图、文转音频等多项功能。总的来说，GPT 大模型是 NLP 和人工智能领域划时代的标志，也昭示着人类向通用人工智能或强人工智能迈出了尝试性的一步。

GPT-4 基于 Transformer 架构演进发展，其发展历程如下：

- 2017 年 6 月，Google 发布论文《Attention is all you need》[1]，首次提出 Transformer 模型，成为 GPT 发展的基础；

- 2018 年 6 月，OpenAI 发布论文《Improving Language Understanding by Generative Pre-Training》[2](通过生成式预训练提升语言理解能力)，首次提出 GPT-1 模型 (Generative Pre-Training)；

- 2019 年 2 月，OpenAI 发布论文《Language Models are Unsupervised Multitask Learners》[3]（语言模型应该是一个无监督多任务学习者），提出 GPT-2 模型；

- 2020 年 5 月，OpenAI 发布论文《Language Models are Few-Shot Learners》[4](语言模型应是一个少量样本(few-shot)学习者)，提出 GPT-3 模型；

- 2022 年 2 月底，OpenAI 发布论文《Training language models to follow instructions with human feedback》[5]（使用人类反馈指令流来训练语言模型），公布 Instruction GPT 模型；

- 2022 年 11 月 30 日，OpenAI 推出 ChatGPT 模型并提供试用，自发布两个月内月活跃用户数已达 1 亿，成为史上用户数增长最快的消费者应用；

- 2023 年 3 月 14 日，OpenAI 发布 GPT-4 模型，并在微软的新版搜索引擎必应（Bing）上运行。

- 2023 年 3 月 23 日，OpenAIs 在 ChatGPT 中实现了对插件的初步支持，同步开放了 Expedia、FiscalNote 等 11 个插件。。微软和 OpenAI 将通过互操作性来支持和发展 AI 插件生态系统，包括 Dynamics 365 Copilot、Microsoft Dev Box、Copilot Stack、Azure AI Studio、 Windows Copilot 等，重塑整个微软应用生态。

- 2023 年 5 月 18 日，ChatGPT app 上线苹果应用商店。在推出的短短 6 天时间内，该应用程序的下载量已经超过了 50 万次。

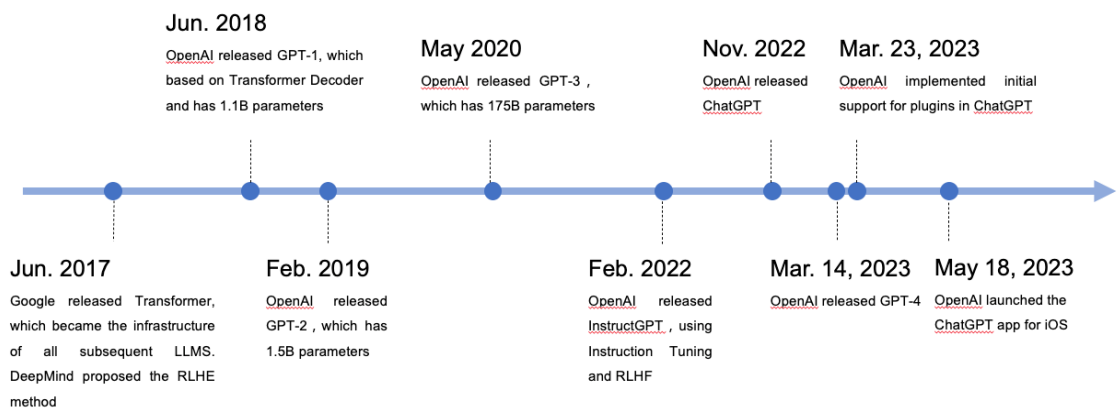

图 2-1 GPT 发展历程

## 2.2 GPT 大模型的能力分析

NLP 领域研究包括了众多子领域，从任务角度看可分为两大类：一类是中间任务，如中文分词、词性标注、句法分析、指代消解等；另一类是最终任务，

如文本分类、文本摘要、机器翻译、文本对话等，最终任务的完成往往依赖中间任务的结果。随着 GPT 等 LLM 的出现，中间任务逐渐退出，LLM 可以直接端到端实现最终任务。而 GPT 大模型则用自然对话的人类表达的方式，提供了 LLM 与人的新型交互接口，以统一的接口方式支撑 NLP 的所有领域端到端最终任务。在日常对话中，ChatGPT 和 GPT-4 之间的差异很小。但是当任务的复杂性超过了某一阈值时，GPT-4 的表现则要优于 ChatGPT。目前，通过与 OpenAI 合作，一些组织利用 GPT-4 开发了一系列创新产品如 Duolingo (GPT-4 deepens the conversation)、Be My Eyes (use GPT-4 to transform visual accessibility)、Stripe (leverage GPT-4 to streamline user experience and combat fraud)等[6]。下面以 GPT-4 对话接口的方式，通过示例进行 GPT-4 端到端任务能力分析。

### 多种风格解释复杂名词

多风格复杂名词解释指人工智能在解释一个名词时使用多种不同的风格或方式，包含严谨、科技、幽默、简单易懂等风格。这种能力可以在不同的场景中得到应用，例如学术知识科普、复杂概念简化等。GPT 大模型能够以各种风格来扮演各种角色，向各种类型的用户解释一些复杂的名词。例如，向 GPT 大模型询问通信行业的 5G 网络技术时，它所产生的回答如下图 2-2。

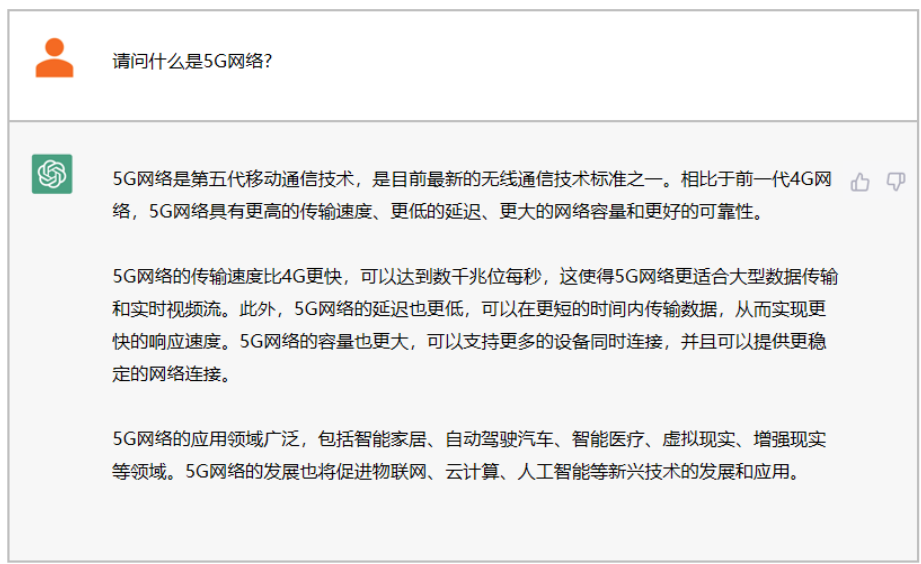

**图 2-2 GPT 大模型解释复杂名词**

上述 GPT 大模型对于 5G 网络的解释，可能很难让一个 5 岁的孩子快速理解。当告诉 GPT 大模型，需要 GPT 大模型简化概念，向一个 5 岁的孩子介绍时：

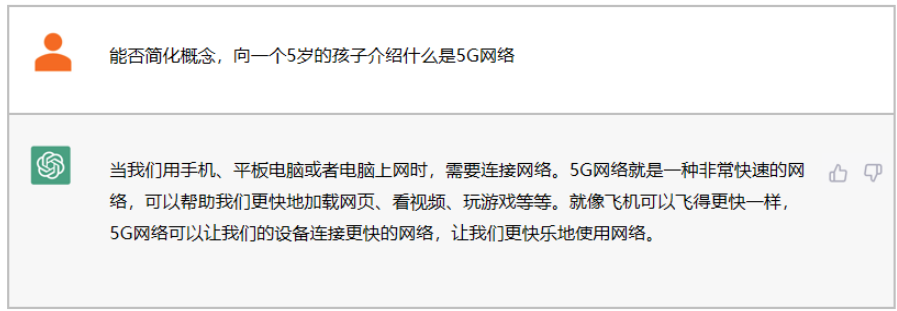

**图 2-3 GPT 大模型简化概念，解释复杂名词**

而对于一些想深入了解 5G 网络其中所运用到的技术的用户而言，GPT 大模型能够以一个通信网络工程师的身份，向用户介绍 5G 网络的概念：

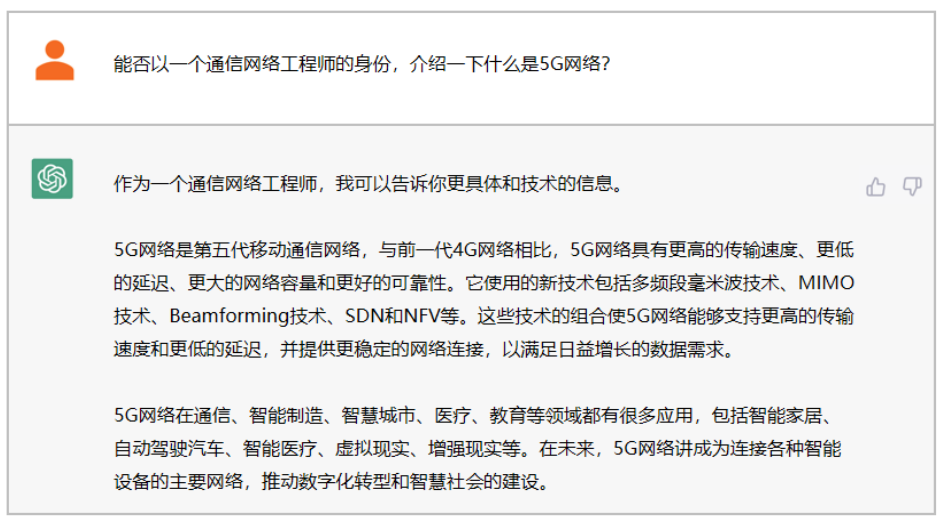

**图 2-4 GPT 大模型以专业人士身份，解释复杂名词**

## 特定要求或主题创作文章

AI 写作根据用户给定的输入生成符合语法和语义要求的文本内容，包括新闻报道、技术介绍、小说、商业信函等等。在 AI 写作的应用场景中，人们通常会提供一些初始的输入，例如主题、文章结构等等特定要求，AI 会根据这些输入自动生成对应的文本内容。与传统的人工撰写相比，AI 写作可以大大提高工作效率，同时还可以生成更加多样化和精准的文本内容，因此越来越受到各行各业的关注和应用，图 2-5 为 GPT 大模型所撰写的关于 AIGC 的文章：

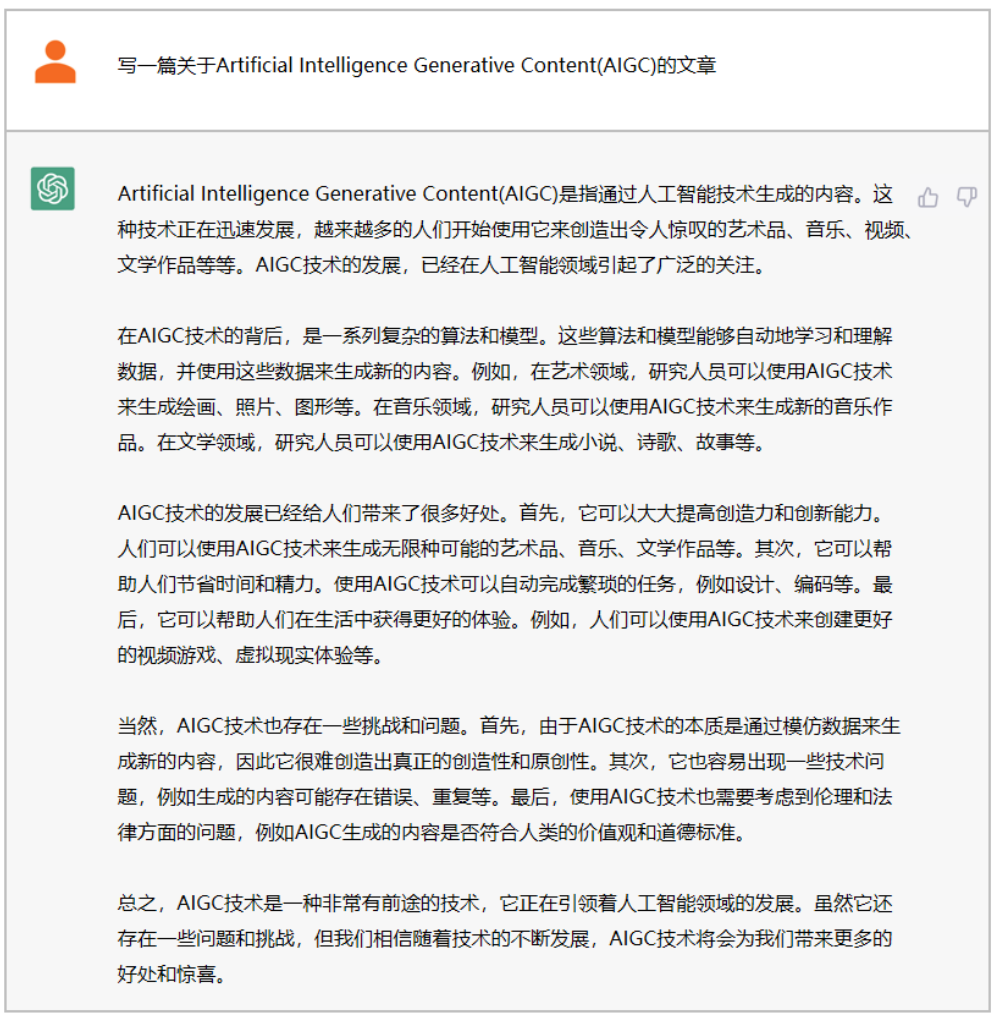

**图 2-5 GPT 大模型根据特定要求或主题创作文章**

## 对话机器人

对话机器人要求人工智能理解用户的语言输入，并根据用户需求和问题生成相应的回答。用户可以通过对话机器人进行语音或文字交互，就像与真人交谈一样，从而获得所需的信息和服务。对话机器人可以应用于多种场景，例如客户服务、虚拟助手等。在客户服务领域，对话机器人可自动回答客户的常见问题，解决客户的问题，减少客服工作量，提高客户满意度；在虚拟助手领域，对话机器人可以帮助用户完成日常任务，如提醒、查询、推荐、预定等，图 2-6 为 GPT 大模型根据用户的具体需求而推荐的购机指南：

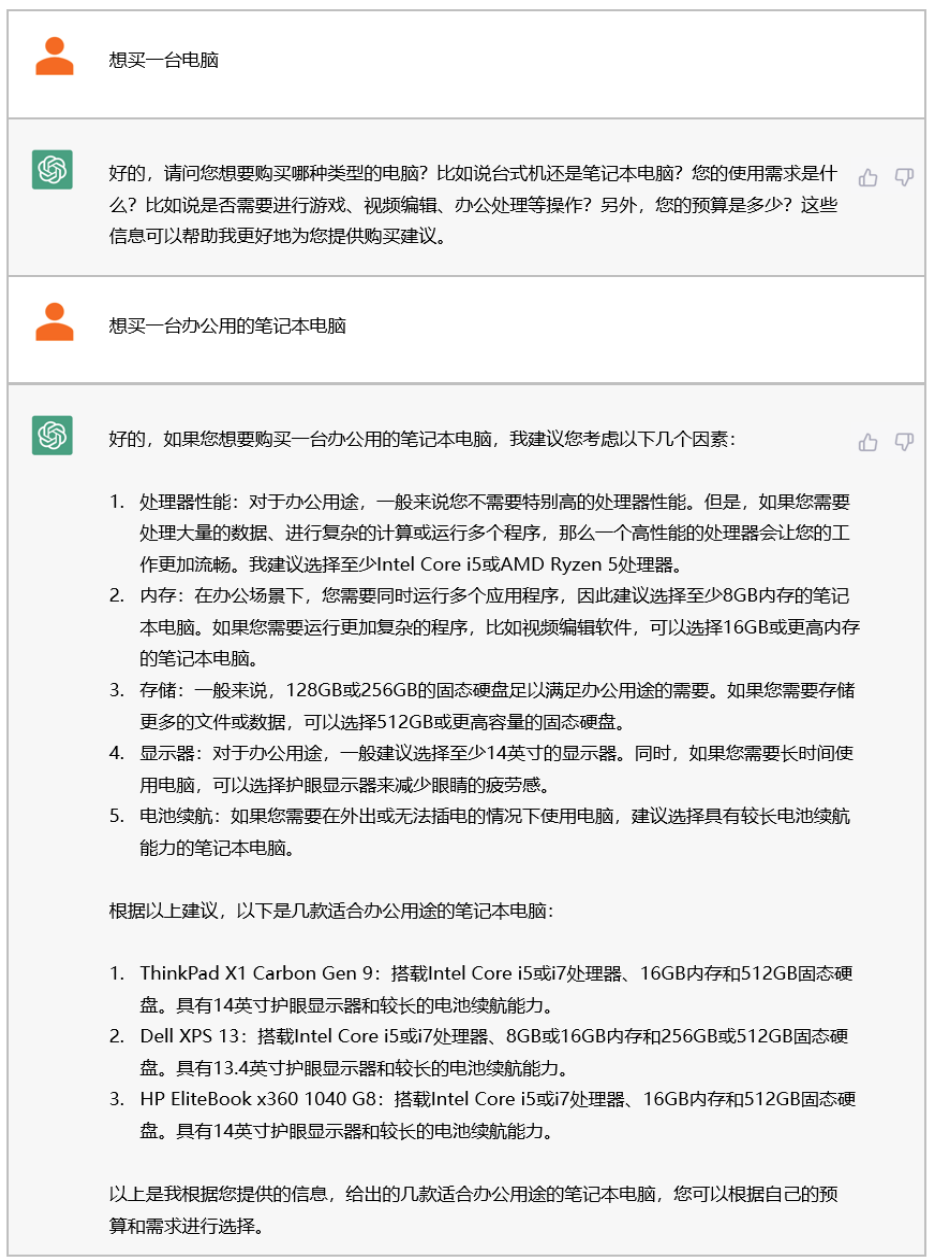

**图 2-6 GPT 大模型与用户多轮对话进行推荐**

## 辅助进行代码编写、Debug 及解释

GPT 大模型具有非常强大的代码生成能力，可以生成、修正各种编程语言的代码片段。具体而言，GPT 大模型的代码生成能力主要基于两个方面：一是其能够理解并生成自然语言描述的程序逻辑；二是其能够生成符合编程语言语法规则的代码。GPT 大模型通过对大量程序源代码和自然语言描述的语料进行训

练，学习到程序的逻辑结构和语义含义。当输入一段自然语言描述的程序逻辑时，GPT 大模型可以理解其中的含义并生成对应的代码。GPT 大模型的训练数据也包括大量的符合编程语言语法规则的代码片段，因此它可以很好地遵循编程语言的语法规则生成代码。GPT 大模型还可以结合常见的编程库和框架生成相应的代码，比如生成使用 Scikit-learn 进行机器学习训练的 Python 代码。图 2-7 为GPT 大模型使用 Python 编程语言和机器学习框架 Scikit-learn 进行机器学习训练的代码：

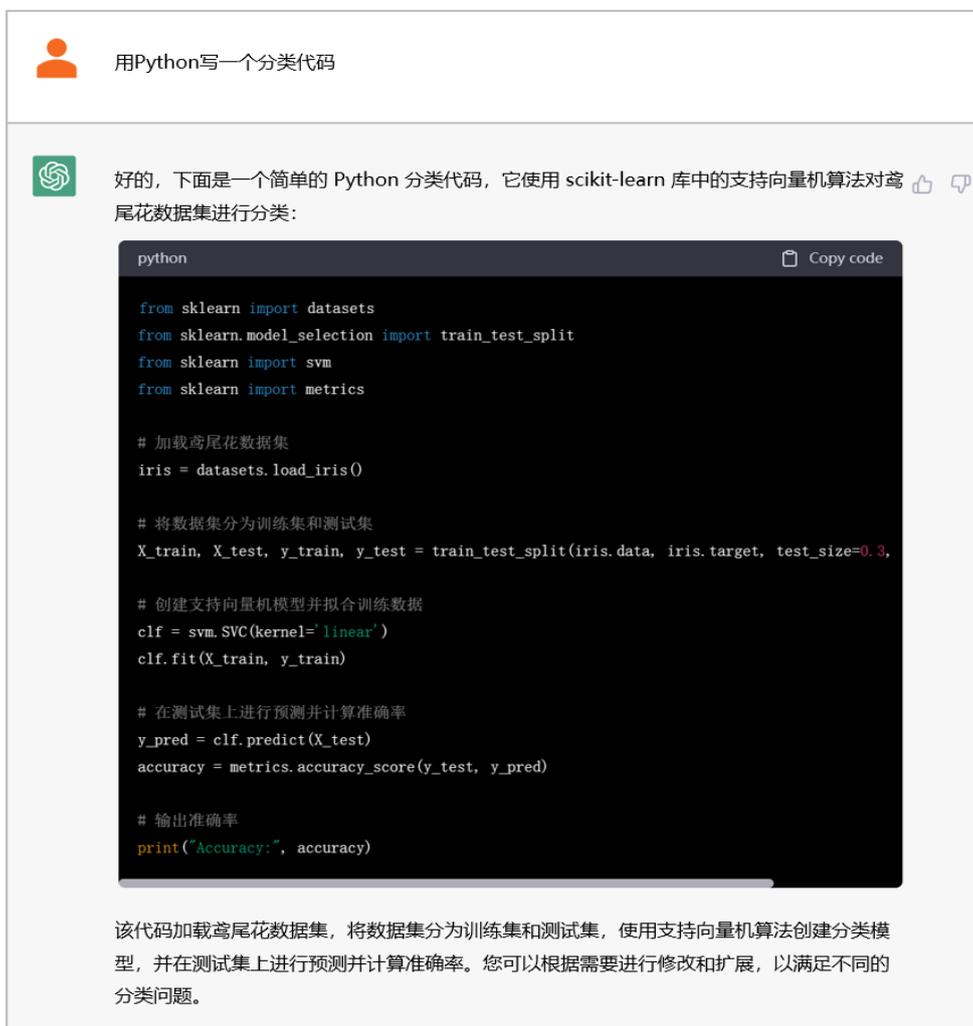

**图 2-7 GPT 大模型智能生成代码**

GPT 大模型具备代码解释能力。对于一些缺少注释的代码，GPT 大模型也能够自动识别代码块的编程语言，代码块的缩进等，以代码注释的方式向用户解释每行代码的具体作用：

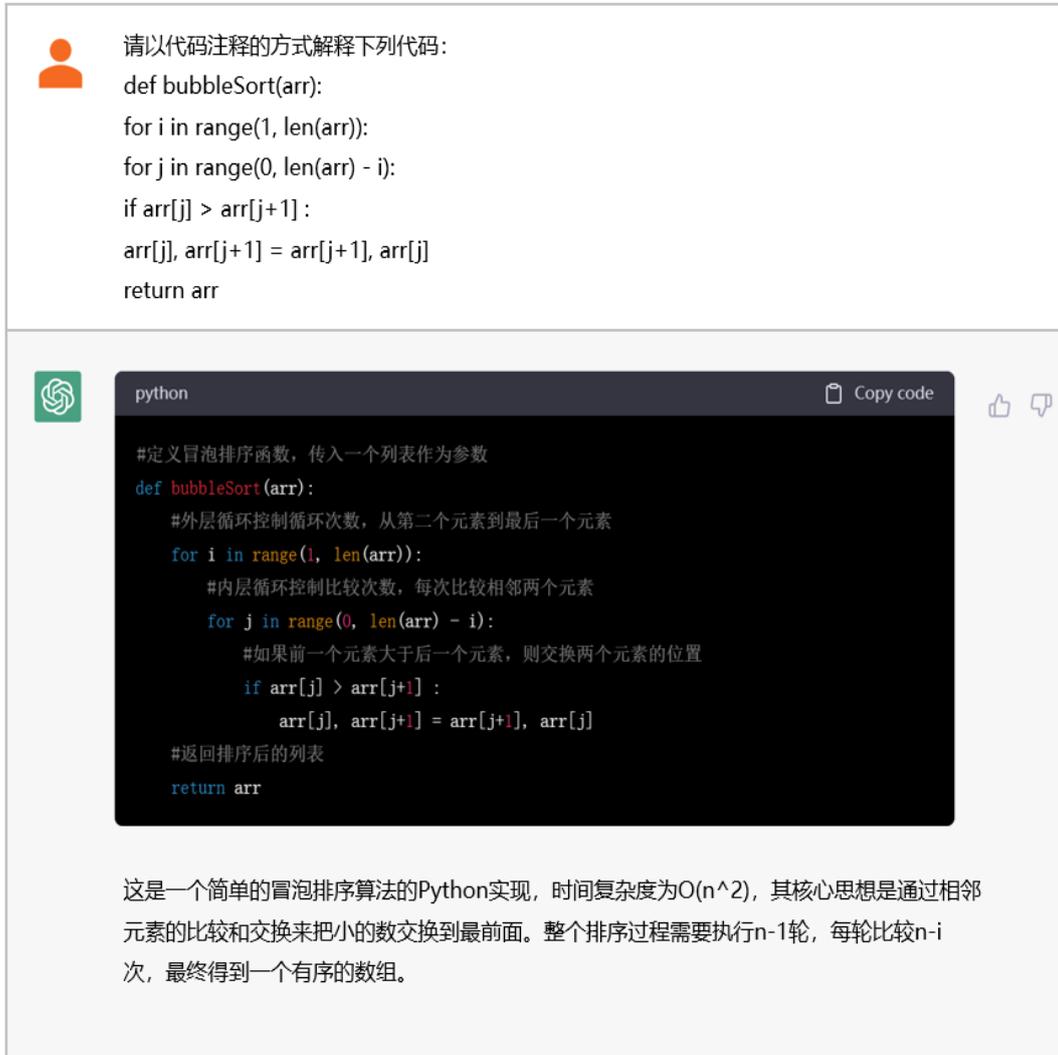

图 2-8 GPT 大模型智能解释代码

GPT 大模型具备代码修正能力。GPT 大模型能够自动识别代码块中所存在的错误并给出合理的解释并修正：

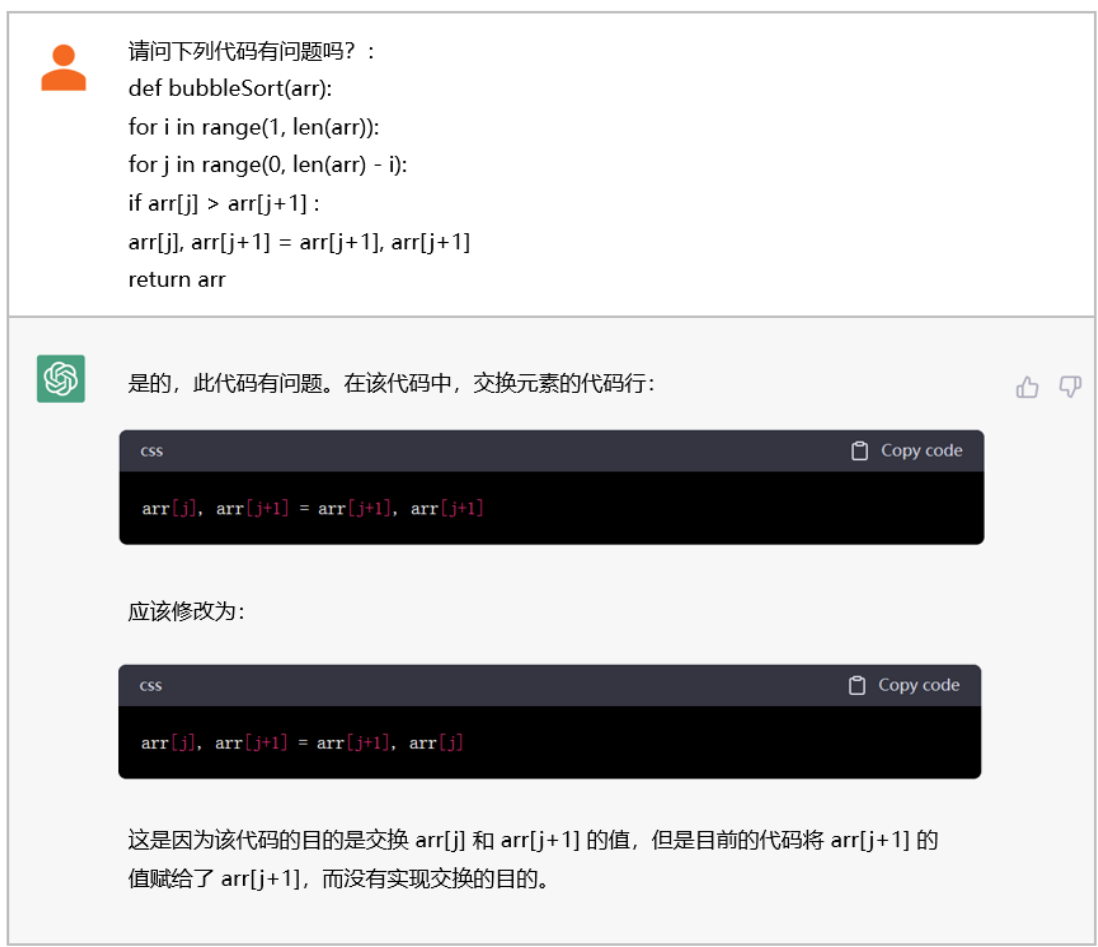

**图 2-9 GPT 大模型智能修正代码**

文本知识抽取

文本知识抽取是指人工智能从自然语言文本中提取结构化或半结构化的信息或知识。对于 GPT 大模型而言，用户仅需要输入特定的文本抽取任务和待抽取的文本，GPT 大模型就能够抽取出文本中的关键性实体，对于苹果即将上市的 iPhoneSE4 智能手机配置信息，GPT 大模型的抽取结果如下图 2-10：

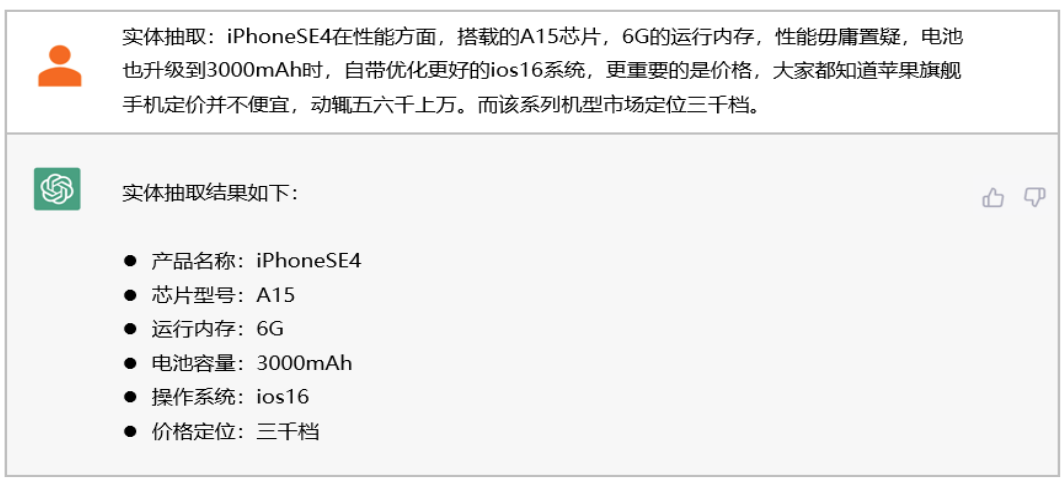

**图 2-10 GPT 大模型抽取文本中关键信息**

## 阅读理解与分析

阅读理解与分析任务要求人工智能不仅仅能理解和抽取文本中的关键字词，还能够推理出更加深入的信息和数据。对于 2022 年亚信科技公司的年中财报数据，GPT 大模型不仅仅能够将文本中的关键数字提取并列表，还能够推理出 2021 年的三新业务收入、三新业务收入占比等：

学习下面的内容，用表格的形式，对比一下亚信科技2022年上半年和去年同期的营收：近日，亚信科技公布了公司截至2022年6月30日中期综合业绩。报告期内，公司新签订单同比增长超过30%；收入达到约人民币31.09亿元，同比上升14.5%。其中三新业务保持高速增长，收入达到约人民币10.04亿元，同比上升81.4 %，三新业务收入占收入比重提升至32.3 %。

| 经济指标 | 2022年上半年 | 2021年上半年 | 同比变化 |
|---|---|---|---|
| 收入（亿元） | 31.09 | 27.13 | 上升14.5% |
| 三新业务收入（亿元） | 10.04 | 5.53 | 上升81.4% |
| 三新业务收入占比（%） | 32.3 | 20.4 | 上升11.9个百分点 |

根据亚信科技公布的中期综合业绩，可以看出亚信科技2022年上半年的营收同比去年同期增长了14.5%。其中，三新业务收入同比增长了81.4%，三新业务收入占比也从去年同期的20.4%提升至32.3%。

**图 2-11 GPT 大模型智能分析文本和数据并制表**

### 复杂问题分析

GPT 大模型具有渊博的知识和强大的解决问题的能力，它可实现更高的精度分析解决复杂的问题：

> 在使用生成对抗网络（GANs）进行图像生成时，如何解决模式崩溃（mode collapse）问题，同时避免生成器（generator）和判别器（discriminator）之间的训练不稳定性，以提高生成图像的质量和多样性？讨论可能的解决方案和相应的优缺点

在使用生成对抗网络（GANs）进行图像生成时，解决模式崩溃和训练不稳定性的问题至关重要。以下是一些可能的解决方案及其相应的优缺点：

1. 使用迷你批次（minibatch）判别：
   在判别器中，而非针对单个样本进行判断，使用小批量样本的统计信息来评估真实性。这有助于生成器产生更多样化的样本。
   优点：有助于提高多样性。
   缺点：可能会增加训练时间。
2. 修改损失函数：
   使用Wasserstein GAN（WGAN）或其他替代损失函数，可以减轻训练不稳定性问题。
   优点：更稳定的训练过程。
   缺点：需要调整超参数以获得最佳性能。
3. 梯度惩罚（Gradient Penalty）
   在WGAN的基础上，引入梯度惩罚，确保判别器的梯度在局部保持合适的范围，有助于避免模式崩溃。
   优点：提高稳定性和生成样本质量。
   缺点：计算成本较高。
4. 自动编码器-生成对抗网络（AAE-GAN）
   结合自动编码器（AE）和生成对抗网络（GAN）的优点，学习生成更多样化的样本。
   优点：提高多样性，缓解模式崩溃问题。
   缺点：需要训练额外的自动编码器。
5. 实例噪声（Instance Noise）
   在训练期间，在判别器的输入数据中添加随机噪声。这有助于平滑判别器的决策边界。提高生成图像的多样性。
   优点：提高多样性，降低训练不稳定性。
   缺点：可能导致生成图像的质量降低。
6. 使用卷积神经网络（CNN）
   使用卷积神经网络作为生成器和判别器，有助于捕获图像的局部特征，并改善生成图像的质量。
   优点：提高生成图像的质量。
   缺点：计算复杂度较高。
7. 结合多个生成器和/或判别器：
   使用多个生成器和/或判别器进行训练，以提高生成图像的多样性。
   优点：提高多样性，降低模式崩溃的风险。
   缺点：计算成本较高，需要协调多个生成器和判别器。

综上所述，解决GAN中的模式崩溃和训练不稳定性问题需要综合考虑各种方法。在实际应用中，可以根据具体任务和需求，结合多种技巧来提高生成图像的质量和多样性。需要注意的是，每种方法都有其优缺点，可能需要对超参数和网络结构进行调整，以找到最合适的解决方案。

**图 2-12 GPT 大模型分析解决复杂问题**

### 图文分析

　　GPT 大模型可以同时接受文本和图片输入，通过分析图片内容，结合输入文本作出合理回复[6]：

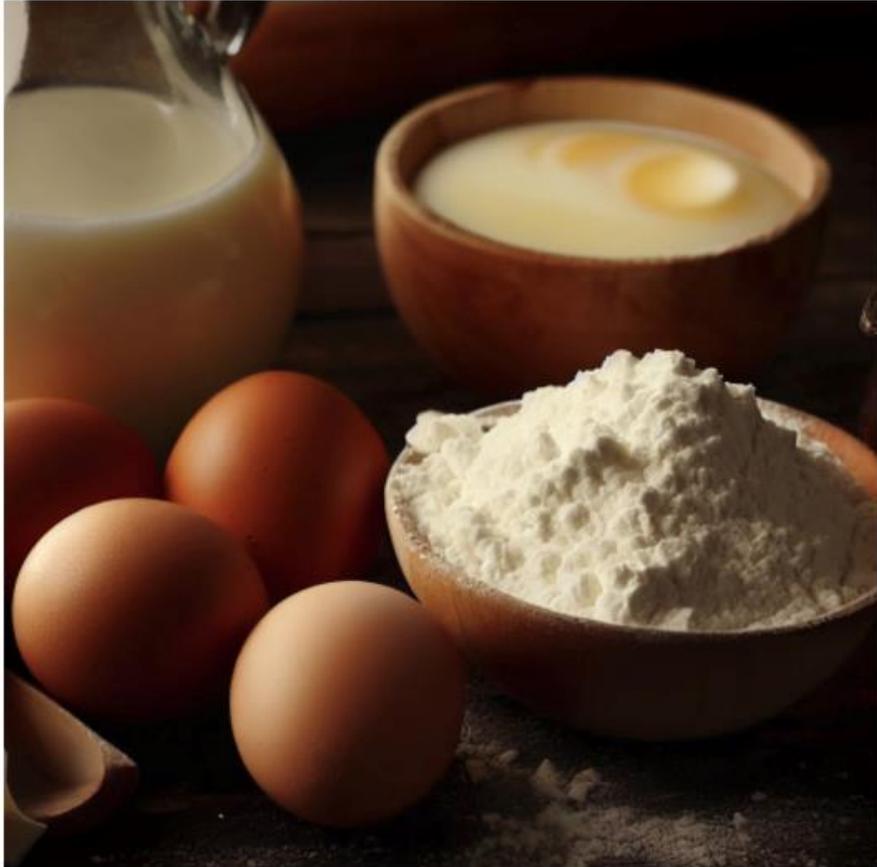

**图 2-13 GPT 大模型处理文本图片混合输入**

从上述分析示例看，GPT 大模型展示了四个极为重要的能力：

- **为人类提供世界知识**：GPT 大模型拥有庞大的世界知识，包括事实性知识和常识，相比之前的 LLM 覆盖了更多的主题和知识。不仅可以提供知识的查询和检索，而且可以进行知识的归纳和总结，甚至提供简单的知识推理和证明能力。

- **为人类与 LLM 之间提供了符合人类习惯的交互方式**：以遵循提示（prompt）并生成补全提示词的句子的语言生成方式，通过上下文学习保持对话一致性，能理解人类意图并用自然语言回答问题、生成内容和解决问题，改变了现有人机互动方式及人类获取世界知识的方式。

- **为人类提供了跨语言及多模态交流能力**：不仅可以处理多种人类语言，还可以理解图片内容，为人们提供更加便捷的交流，此外还可以将人类语言与机器语言进行相互翻译，促进了人机物三元世界的融合。

- **具备了自我学习，自我进化的能力**：能够自动从海量数据和人类指令中学习到其中的世界知识，学习过程不需要人的介入，而且可以自行检查学习成果并不断优化迭代，能灵活应用所学知识来解决实际问题。

当然，我们也看到，由于 GPT 大模型是基于通用公开数据进行训练，缺少各行业领域的非公开的专业数据。这就造成其世界知识中缺乏深度的专业知识。因此，虽然在与人类进行沟通中，可以进行百科全书式的回答，但距离真正满足专业场景，还有相当大的差距。

## 2.3 GPT 大模型的技术原理

通过当前最先进的 GPT-4 介绍 GPT 大模型的技术原理。GPT-4 是 OpenAI 第四代 GPT 模型，其技术原理与 GPT-3.5 版本的 ChatGPT 较为接近。ChatGPT 的技术特征已经包含在其全称 Chat Generative Pre-trained Transformer 中，是针对对话（Chat）进行的专门优化，GPT-4 保留了上述全部技术特征。Transformer 是 GPT 系列的基础架构，预训练和生成式是其成功之匙。本节通过 Transformer、GPT-1、GPT-2、GPT-3、InstructGPT、ChatGPT 等一系列 GPT 技术的演进脉络分析，阐述 GPT 大模型的技术原理。

2017 年，谷歌大脑团队在机器学习与人工智能领域顶会 NeurIPS 发表了 "Attention is all you need"论文，首次提出了基于自注意力机制（self-attention）的变换器（Transformer）模型并用于自然语言处理。相比于传统的循环神经网络（RNN）和卷积神经网络（CNN），Transformer 采用自注意力机制、多头注意力机制、前馈网络、残差连接等技术，具有更好的长距离特征捕获能力、特征提取能力和并行运行效率[1]。这些能力使 Transformer 可以处理更大量的数据，形成更大规模的网络，从而引发了预训练模型的形成，并进一步使其在自然语言处理任务中获得出色表现。Transformer 的诞生，深刻地影响了接下来几年人工智能领域的发展，标志着预训练大模型的开始[7]。在不同的场景中，Transformer 衍生出了不同架构。例如，在生成式任务中，GPT-3 是基于自回归序列思路的模型[8]；在自然语言理解任务中，BERT 是基于 Transformer+Mask 的自编码序列思路的模型[9]；在条件文本生成任务中，T5 使用 Encoder-decoder 架构，利用双向或单向 attention 来解决条件文本生成任务 [10]。

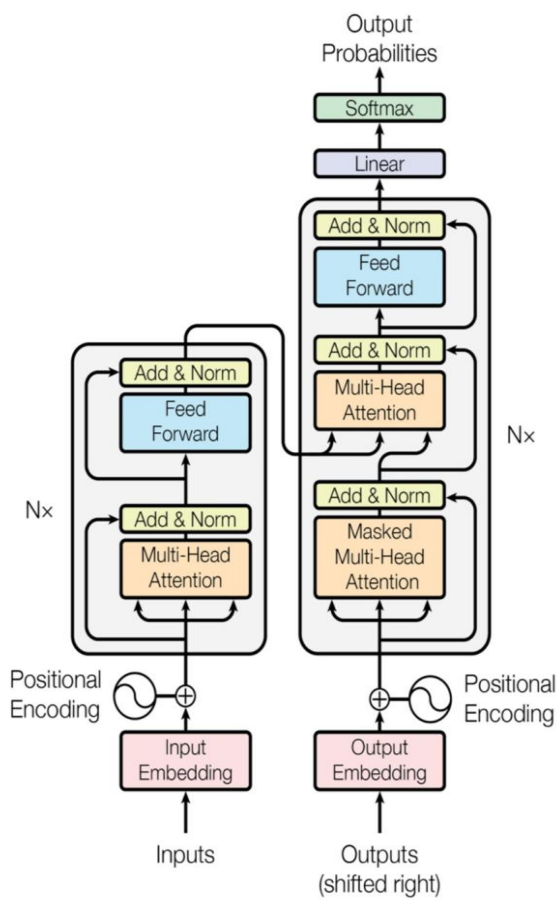

**图 2-14 Transformer 架构**

GPT-1 模型基于 Transformer 解决了顺序关联和依赖性的问题，并采用生成式方式，重点考虑了从原始文本中有效学习的能力。在 GPT-1 中，采用了 12 个 Transformer 块结构作为解码器，每个 Transformer 块是一个多头的自注意力机制，然后通过全连接得到输出的概率分布[2]。GPT-1 的训练分为无监督的预训练和有监督的模型微调，这对于减轻自然语言处理中对监督学习的依赖至关重要。但是，GPT-1 采用的是自回归序列的架构，难以处理过长的文本。在生成长文本时，GPT-1 会出现信息遗忘和重复等问题。在某些特定任务上，如机器翻译等，其表现并不如其他特定领域的模型表现出色。

GPT-2 的目标是训练一个泛化能力更强的词向量模型，通过使用无监督的预训练模型做有监督的任务，尝试解决 zero-shot(零次学习问题)[3]。GPT-2 没有对

GPT-1 的网络进行过多的结构创新与设计，只是使用了更多的网络参数和更大的数据集。GPT-2 的最大贡献是验证了通过海量数据和大量参数训练出来的词向量模型可以迁移到其他任务中，而不需要额外的训练。然而实验表明，GPT-2 的无监督学习和零次学习结果未达预期。

GPT-3 在 GPT-2 追求无监督和零次学习的特征基础上进行了改进，转而追求无监督模式下的 few-shot(少量学习)。GPT-3 在网络容量方面有了巨大提升，基于 45TB 的文本数据训练，在多个 NLP 数据集上实现了出色的性能[4]，具体来说：采用了 96 层的多头 Transformer，每层有 96 个头；词向量长度为 12888；上下文窗口大小提升至 2048 个 token；采用 alternating dense 和 locally banded sparse attention。GPT-3 能够生成高质量的文本，但有时在编写长句并反复重复文本序列时可能会失去连贯性。而且无监督模式下的 few-shot 效果仍然略逊于 fine-tuning 监督微调的方式。

InstructGPT 是基于 GPT-3 的一轮增强优化，所以也被称为 GPT-3.5。InstructGPT 采用了人类反馈的强化学习方案（RHLF），对 GPT-3 增加了监督微调，进一步增加了奖励模型，通过训练出的奖励模型指导语言模型的学习即增强学习优化[5]。这个过程是可以多次迭代优化的，因此 InstructGPT 自动化的实现了比 GPT-3 更好的性能，同时减少了参数的数量。

ChatGPT 采用 InstructGPT 相同结构的模型，针对 Chat 进行了专门的优化，同时开放到公众测试训练，以便产生更多有效标注数据。基于人类反馈的强化学习（RLHF）方法的使用是ChatGPT区别于其他生成类模型的最主要原因，该方法帮助模型尽量减少有害的、不真实的及有偏见的输出，提升自然沟通效果。同时，为了更好地支持多轮对话，ChatGPT 引入了一种基于堆栈的上下文管理的机制，帮助 ChatGPT 跟踪和管理多轮对话中的上下文信息[11]。ChatGPT 将上下

文压入堆栈，并通过堆栈中的上下文信息使用注意力机制对齐用户的意图和之前的对话内容，并根据对话历史生成响应，从而在多轮对话中生成连贯自然的回复。

与之前的 GPT 模型一样，GPT-4 使用公开数据（如互联网数据）和第三方提供商许可的数据来训练模型预测下一个词的能力，这些训练数据包括数学问题的正确和错误解答，弱推理和强推理，自相矛盾及前后一致的陈述，以及各种各样的意识形态和想法。因此，当用户提出一个问题时，模型可能以各种各样的方式响应，其中一些方式可能会偏离用户的意图。为了使模型更好的理解用户的意图，GPT-4 使用人类反馈强化学习（RLHF）对模型进行微调。为了解决 GPT-4 对特定任务微调不可行的问题，OpenAI 建立了一个可预测扩展的深度学习堆栈[6]。此外，OpenAI 团队还开发了基础设施和优化方法，通过使用自回归生成模型的缩放定律[12]，对大规模语言模型最终损失进行预测，增加可预测扩展，减少计算量。

GPT-4 涉及到技术重点包括：指令微调（Instruction Fine-Tuning，IFT）、有监督微调（Supervised Fine-tuning，SFT）、人类反馈强化学习 (Reinforcement Learning From Human Feedback，RLHF)、思维链 (Chain-of-thought，CoT) 等，下述对这些关键技术做进一步介绍：

IFT 是为了解决基础模型未遵循用户的指令时，对基础模型的微调。在基础模型的语言建模目标不足以让模型学会以有用的方式遵循用户的指令，IFT 方法除了使用情感分析、文本分类、摘要抽取等经典 NLP 任务来微调模型外，还在非常多样化的数据集上向基础模型示范各种指令及输出，从而实现对基础模型的微调[14]。指令示范主要由指令、输入和输出组成，输入是可选的，一些任务只

需要指令，例如 ChatGPT 和 GPT-4 做开放式文本生成任务，当存在输入时，输入和输出组成一个实例，给定指令可以有多个输入和输出实例。IFT 的训练数据通常是人工编写的指令或用语言模型生成的实例，在语言模型生成中，先使用少样本技术输入一些样本给 LM 用于提示它，然后要求 LM 生成新的指令、输入和输出，每一轮都会从人工编写的样本和模型生成的样本中各选择一些送给模型。

经过 IFT 的 LM 生成的结果并不一定是有帮助的或安全的，比如通过给出诸如"对不起，我不明白"的模棱两可的或者无益的结果，为了减轻这种行为，需要在高质量的人类标注数据上微调基础的语言模型，以提高有用性和无害性。SFT 和 IFT 联系非常紧密，IFT 可以看作是 SFT 的一个子集，模型开发人员在开发过程中，常把 SFT 阶段用于提高生成的安全性，而不是接在 IFT 后面提高指令响应的具体性[13]。

RLHF 是 InstructGPT、ChatGPT、GPT-4 成功的关键之一。在 RLHF 中，根据人类反馈来对模型的响应进行排序标注 (如，根据人类偏好选择文本简介)。然后，用这些带标注的响应来训练偏好模型，该模型用于返回 RL 优化器的标量奖励。最后，通过强化学习训练对话代理来模拟偏好模型。基于 RLHF 的 GPT-4 训练方法由三个不同的步骤组成[11]，如图 2-15 所示。

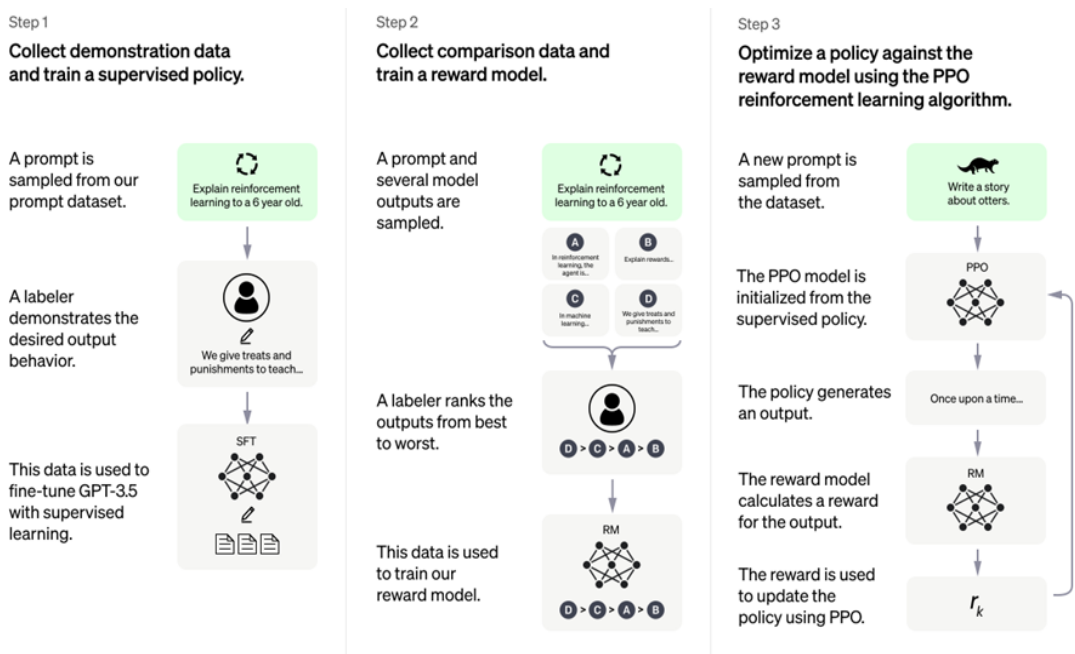

**图 2-15 RLHF 训练方法**

步骤 1：对基于监督学习获取的 GPT-3.5 模型进行微调，得到 SFT 模型。微调是使用问题/答案对完成的。首先会在数据集中随机抽取问题，由人类标注人员，给出高质量答案，然后用这些人工标注好的数据来微调 GPT-3.5 模型，通过对输入和输出数据的拟合来改进网络，调整模型的参数。此时的 SFT 模型在遵循指令/对话方面已经优于 GPT-3，但不一定符合人类偏好。

步骤 2：标注排序数据集，训练奖励模型 RM：这个阶段的主要是通过人工标注训练数据，来训练回报模型。在数据集中随机抽取问题，使用第一阶段生成的模型，对于每个问题，生成多个不同的回答。人类标注者对这些结果综合考虑给出排名顺序。这一过程类似于教练或老师辅导。接下来，使用这个排序结果数据来训练奖励模型。对多个排序结果，两两组合，形成多个训练数据对。RM 模型接受一个输入，给出评价回答质量的分数。这样，对于一对训练数据，调节参数使得高质量回答的打分比低质量的打分要高。

步骤 3：使用强化学习方法 PPO 进一步微调 SFT：PPO 的核心思路在于将 Policy Gradient 中 On-policy 的训练过程转化为 Off-policy，即将在线学习转化为离线学习，这个转化过程被称之为 Importance Sampling。这一阶段利用第二阶段训练好的奖励模型，靠奖励打分来更新预训练模型参数。在数据集中随机抽取问题，使用 PPO 模型生成回答，并用上一阶段训练好的 RM 模型给出质量分数。把回报分数依次传递，由此产生策略梯度，通过强化学习的方式以更新 PPO 模型参数。

其中，步骤 1 只进行一次，步骤 2 和步骤 3 可以进行多次迭代以得到效果更好的 RM 和 SFT 模型。

CoT 提示[15]是指令示范的一种特殊情况，它通过引发对话代理的逐步推理来生成输出。使用 CoT 微调的模型在涉及常识、算术和符号推理的任务上表现得更好[14]，CoT 微调也显示出对无害性非常有效 (有时比 RLHF 做得更好)，而且对于敏感提示，模型不会回避并生成"抱歉，我无法回答这个问题"这样的无益回答[13]。

总的来看，GPT-4 的世界知识能力来源于大数据和大模型，预训练大模型和指示学习的泛化能力，带来了模型强大的理解能力，可以显著降低人工智能应用门槛。同时，IFT、SFT、RLHF、CoT 的极致应用使 GPT-4 的输出与人类的意图对齐，带来了自然、客观、翔实的交互。

## 2.4 GPT 大模型的应用

在 ChatGPT 和 GPT-4 相继公布后，这一波由深度学习驱动的人工智能技术长驱直进，对人类社会产生了巨大影响。据报道[16][17]，89% 的美国大学生已经

在用 ChatGPT 写作业，ChatGPT 通过职位年薪 18 万美元的谷歌 L3 工程师入职测试，也通过了美国执业医师资格考试。微软已经迅速在 Bing 搜索中集成了 GPT-4。在为人类设计的学术专业考试上，测试模型理解能力，GPT-4 能排到 TOP10%，GPT-3.5 排在尾部 10%[6]。在传统 NLP 测评数据集上，GPT-4 超越了所有语言模型与大部分 SOTA 模型，在其他语言上测试结果：26 个测试语言中，24 种语言 GPT-4 优于 GPT3.5 与其他语言模型[6]。可以预见，基于前述分析展现出来的 GPT 大模型的强大生产力和巨大潜力，GPT 大模型与各行各业应用相结合，将在多领域、多功能应用中密集落地[18]。

- **传媒领域**：提升新闻效率，促进网络媒体安全

  GPT 大模型可以帮助新闻媒体工作者智能生成新闻报道，通过智能方式优化重复劳动性的采编工作。过往新闻智能写作的成功案例众多，例如中国地震网的写稿机器人在九寨沟地震发生后 7 秒内就完成了相关信息的编发；美联社使用的智能写稿平台 Wordsmith 可以每秒写出 2000 篇报道。同时 GPT 大模型强大的看图说话能力，整合互联网信息采集技术及信息智能处理技术，可以形成简报、报告、图表等分析结果，为全面掌握传媒思想动态，做出正确舆论引导，提供分析依据。

- **影视领域**：拓宽创作素材，降低后期优化成本

  GPT 大模型可以为剧本素材的创作提供新思路，创作者可根据大众兴趣使用 GPT 大模型生成内容，再进行筛选和二次加工，从而激发创作者的灵感，开拓创作思路，缩短创作周期。还可以根据长篇小说、传记、诗歌定制影视内容，从而更有可能吸引爱好者的注意力，获得更好的收视

率、票房和口碑。2016 年，纽约大学利用人工智能编写剧本《Sunspring》，经拍摄制作后入围伦敦科幻电影 48 小时前十强；国内海马轻帆科技公司推出的"小说转剧本"智能写作功能，服务了包括《你好，李焕英》《流浪地球》等爆款作品在内的剧集剧本 30000 多集、电影/网络电影剧本 8000 多部、网络小说超过 500 万部。

● **营销领域**：打造虚拟客服，提升品牌形象和服务态度

GPT 大模型可以打造虚拟客服，助力产品营销，GPT 大模型虚拟客服可以 24 小时不间断的提供服务，而且不会像人类客服一样产生疲劳。同时，虚拟人客服可以同时处理多个用户的问题，不会像人类客服一样需要等待处理完一个用户的问题后才能接待下一个用户。这使得虚拟客服在高峰期能够更好地应对客户的需求。此外，虚拟客服还可以自我进化和优化自己的服务，提高自己的准确性和服务质量。同时随着 GPT 大模型在多模态、复杂计算、多语言理解等方面的提升，使得虚拟客服的服务水平可以为客户提供更加优质的服务。

● **文娱领域**：加强人机互动，丰富虚拟现实趣味

GPT 大模型可以讲故事、创作小说；可以生成歌词，增加音乐创作的灵感；可以解读图片，满足探索未知渴望；可以作为闲聊机器人，尤其图文交互，增加趣味性；可以作为知识检索库，增强对未知的理解，尤其在游戏方面，GPT 大模型作为一种基于自然语言处理的人工智能技术，

可以帮助游戏开发人员制作出更加智能的游戏。它可以在游戏中生成任务描述，并帮助玩家解决游戏中的问题。此外，GPT 大模型还可以用于生成 NPC 的对话，从而让游戏的故事更加生动。

- **教育领域**：激发教材活力，释放孩子探索天性

  GPT 大模型能够帮助研究人员撰写连贯、（部分）准确、信息丰富且系统的论文，写作效率极高（2-3 小时），而且不需要作者有丰富的专业知识。即使作者改变了主题，GPT 大模型也能给出完全相同格式的回答。除了直接作用于文案本身，在接入 GPT 大模型以后，在线教育的模式和质量也会有明显的变化。比如，在教学过程中，老师往往会面临跨学科教学，学生的一些问题也需要老师长时间思考才能解答，但如果使用 GPT 大模型，就可以第一时间为学生提供快速简便的问题答案；一些重复性的问题，GPT 大模型就可以直接代替老师回答，让老师腾出大量时间在思想创造力上发挥更大价值；尤其是对不知道如何描述问题或者问题本身就是一些图片，GPT 大模型可以从多个维度给出问答答案。

- **金融领域**：促进风控安全，保驾金融科技稳步发展

  GPT 大模型在金融领域的应用范围非常广泛，包括且不限于：欺诈检测和风险管理、客户服务与支持、客户意图识别及精准营销，对金融市场的信息处理和决策辅助等。反欺诈 GPT 大模型可以快速识别和分析怀疑欺诈的数据，并在接收到异常请求时进行快速反应，以确保客户在资金交易过程中的安全；金融风险控制金融机构需要不断地进行风险评估和

风险控制，GPT 大模型可以通过对历史数据的学习，生成预测模型，帮助机构进行风险评估和风险控制；金融舆情分析金融机构需要对市场和客户的情绪进行分析，以便做出正确的投资决策，GPT 大模型可以通过对大量社交媒体和新闻数据的学习，生成情感分析模型，帮助机构了解市场和客户的情绪变化；充当研究助理 GPT 大模型可以结合金融行业的信息，构建知识库问答系统。对于一些初级的资料信息和数据分析提取，使用者通过询问可以快速找到相关问题的答案。

● **医疗领域**：提高诊断能力和效率，减负医护人员

GPT 大模型的复杂名词解释、知识抽取能力、多轮对话能力、复杂问题分析、图文分析能力等，可以在医学影像诊断、临床辅助决策、精准医疗辅助、健康管理、医疗信息化、医药研发以及医疗机器人等细分场景中全面开花，以助力降本增效、提升医疗水平、改善患者体验、降低患病风险等为目的，全面赋能院前、院中、院后各个环节。哈佛大学一项研究表明：在不同严重性的 45 个真实病例中，ChatGPT 能够对 39 个病例（87%）做出正确的诊断，而同样的病例，人类医生的准确性可达 84%，其他工具准确率只有 51%，GPT 大模型预计会达到更高的诊断精确率。

# 2.5 下一个 GPT 是什么?

ChatGPT 属于 AIGC 的范畴，代表了文本生成领域的最新进展。GPT 大模型则增加了图到文的生成，具备了图片识别的能力。下一个 GPT 预计将支持更多模态的 AIGC，即除文本生成之外，还可以生成图片、视频、音频等模态。与仅使用语言或仅使用视觉相比，构建良好的多模态模型更是一项极具挑战性的任务。但如同 Transformer 架构从语言走向视觉领域，下一个 GPT 或者 LLM 模型，必将向文本、图片、音频、视频等跨领域多模态 AIGC 不断发展演进。

Google、Stability AI、OpenAI、百度、阿里等公司已经开始多模态内容生成模型研发，并开源部分模型，比如图片生成的 Stable Diffusion[19]、音乐生成的 MusicLM[20]、视频生成的 Dreamix[21]等。

在文本生成场景中，可以分为非交互式生成和交互式生成。非交互式文本生成场景中，结构化写作其形式相对固定，生成难度较小，商业化应用较为广泛，例如公司财报简讯、新闻简讯、专业文书等，而创作型写作开放性较大，在长文本生成中难度较大，仍需技术进一步发展，相比于结构化的文本，创作型文本具有更高的开放度与自由度，需要一定的创意与个性化[22]，如营销文案、广告语、客服话术等。交互式文本生成场景主要以聊天机器人为代表，在基于上下文中信息进行文本交互，如客服问答、闲聊机器人、知识检索库等。

在图像生成场景中，模型结构不断进化提高了 AIGC 生成图像的多样性，但要求较高的功能实现还有待于技术的进一步提升。"图像编辑"难度低于"图像生成"与"2D-3D"转换，目前已存在多款产品支持"图像编辑"，而对于"图像生成"任务，由于图片相较包含更多元素，其生成效果仍存在不稳定性，对于要求较高的功能类图像生成仍需要技术上的提升[23]。

音频生成场景中，文本到语音任务已比较成熟，语音质量已达到自然的标准，未来将朝更富情感、富韵律的语音合成以及小样本语音学习方向发展。但音乐生

成任务中仍需解决音乐数据难以标注的问题，数据标注的颗粒度大小影响音乐生成任务的可控性。若可控性得以解决，则可指定风格、情绪等因素的音乐生成任务有希望在影视、游戏等场景下得到大量应用[24]。

视频生成场景中，视频生成本质上与图片生成类似，通过对视频进行帧数级别的切割，实现对每一帧的处理。视频生成过程包括三个阶段：数据的提取、训练和转换，当前技术正在着重提升视频修改精准度和实时性两个维度。鉴于视频本身的文本、图像和音频的综合属性，视频生成也是跨模态生成领域的重要应用场景。

跨模态生成场景中，现实世界的信息是文本、音频、视觉、传感器以及人类各种触觉的综合体系，要更为精准地模拟现实世界，就需要将各种模态能力之间打通，例如文字-图像、文字-视频等跨模态生成能力 大型预训练模型的发展使得跨模态逐步成熟，"文本-图像"生成正在快速落地，"文字-视频"的实验效果也已较为理想（视频时长、清晰程度、逻辑等还有较大提升空间）。

随着 chatGPT/GPT-4 的火爆，中国 GPT 大模型呈现涌现态势，"百模大战"正在上演，打造跨行业通用化人工智能能力平台，应用行业在从办公、生活、娱乐向医疗、工业、教育等加速渗透。但总体来看，中国大模型水平距离 GPT-4 仍有较大差距，一定程度影响了实际应用效果。

# 三. GPT 大模型赋能通信行业应用

通信技术与人工智能技术在 3G 阶段开始融合发展，到 5G 阶段开始深度融合发展。"5G 与人工智能"已被业界视为一组最新的通用目的技术组合，对垂直行业的发展起到提振生产力与赋能的作用。亚信科技携手中外运营商、清华大学智能产业研究院等工业和学术界伙伴，陆续发布了《通信人工智能的下一个十年》[25]、《通信人工智能赋能自智网络》[26]、《算力网络的第一个十年》、《BSS/OSS 融合的下一站》[27]、《下一代电信 IT 架构初探：业务功能虚拟化（BFV）》[28]等著述，介绍了通信人工智能及其方法主旨、路径与框架。人工智能赋能通信行业已在产业界的努力下取得了丰富的实践成果。随着 GPT 大模型带来更加先进的生产力，人工智能将进一步从各个层面深度赋能通信业，推动智能通信网络的演进升级和落地应用。

本章将重点通过 GPT 大模型所采用的对话框方式进行其赋能通信行业应用的场景化示例分析，以下场景分析以"文到文"方式为主进行展现。可以想象，后续多模态 GPT 大模型将为通信行业应用带来更加丰富的赋能场景。

## 3.1 通信 IT 领域的典型应用场景

通信 IT 领域覆盖范围非常广泛，以下将从智能开发、智能运维、智能营销、智能计费、智能客服、智能客户关系管理、商业智能等方面，针对 GPT 大模型如何赋能通信 IT 领域进行分析。

### 3.1.1 智能开发

每个 IT 组织都不遗余力的关注研发效能提升，十多年来智能开发不断迭代演进，对研发生命周期的需求、设计、开发、集成、测试、发布等关键阶段进行了不断实践。随着人工智能技术的发展，研发工作已经不再是传统的代码编写和测试工作，越来越多的研发人员需要处理大量的文本、语言数据和自然语言处理的任务。在研发领域中，GPT 大模型可以作为一种辅助工具，帮助研发人员更快速、更高效地进行开发和测试工作，从而提高研发效率和质量。此外，GPT 大模型还可以用于智能化的文档生成、自动化测试用例生成、智能化代码审查和优化、智能化项目管理等领域，为研发团队提供更全面、更高效的支持和帮助，从而提高整个研发流程的效率和质量。借着 GPT 大模型的东风，通过其对软件研发各个阶段的全面赋能，软件研发也将进入一个新的智能研发时代。

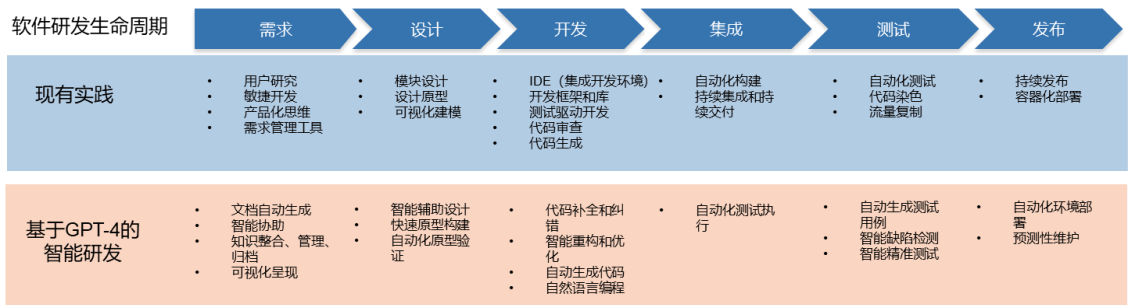

**图 3-1 GPT 大模型对于软件研发生命周期的变革**

#### 场景一：GPT 大模型赋能需求研究、分析和处理

在需求阶段，为了更好地理解和管理需求，优化研发流程，减少重复工作和错误，提高研发效率和质量，相关研发效能提升适用的场景主要有：

1. 用户研究：在需求阶段进行用户研究，了解用户需求和痛点，以此来确定产品或系统的功能和特性，从而确保产品或系统能够满足用户需求，避免开发出无用或者不受欢迎的功能。GPT 大模型引入可以很大程度的

提升用户研究的效率，提升需求的理解；可以更好地处理自然语言输入，使得需求文档的编写更加自然和简单；可以理解文本中的上下文、语义和逻辑关系，从而更好地理解需求。

2. 需求文档：采用产品化思维，将需求阶段看作产品开发的一部分，强调需求的用户价值和商业价值，以此来提高产品的市场竞争力和用户满意度，从而提高研发效能。GPT 大模型可以在自动化生成、智能协助方面提升需求文档的质量和编写效率。基于 GPT 大模型的技术，可以通过输入简单的文本或问题，自动生成相关的需求文档、用例和其他文档。这样可以节省研发人员的时间和精力，同时减少文档编写的错误。GPT 大模型可以智能地协助需求编写和管理。例如，它可以根据之前编写的需求文档或历史数据，提供自动建议或纠错建议，使得需求文档的编写更加准确和高效。

3. 需求管理：使用需求管理工具，对需求进行统一管理、跟踪和分析，以便研发团队能够更好地理解需求，掌握需求状态和变化，从而更好地管理需求和优化研发流程。GPT 大模型的引入，更好的管理需求相关的知识。GPT 大模型可以将需求阶段的知识整合、管理和归档，使得知识能够得到更好的传承和沉淀。这样可以提高研发团队的知识共享和协同效率，同时也能够更好地应对团队流动性和知识遗漏的问题。此外，还可以提升可视化呈现的效果。GPT 大模型可以将需求文档和其他相关文档进行可视化呈现，例如生成流程图、状态图等，使得需求文档更加易于

理解和使用。这样可以提高需求文档的可读性和易用性，从而更好地支持研发流程。

需求分析人员，借助 GPT 大模型，可以快速沟通澄清需求，需求文档可读性大大增强，需求变更管理将变得更加轻松。集成协同音视频的能力，拉通上下游的沟通，随时随地的发起快速沟通。需求分析，从原先以周为单位，降低到以小时为单位，人员效能提升 100 倍以上。

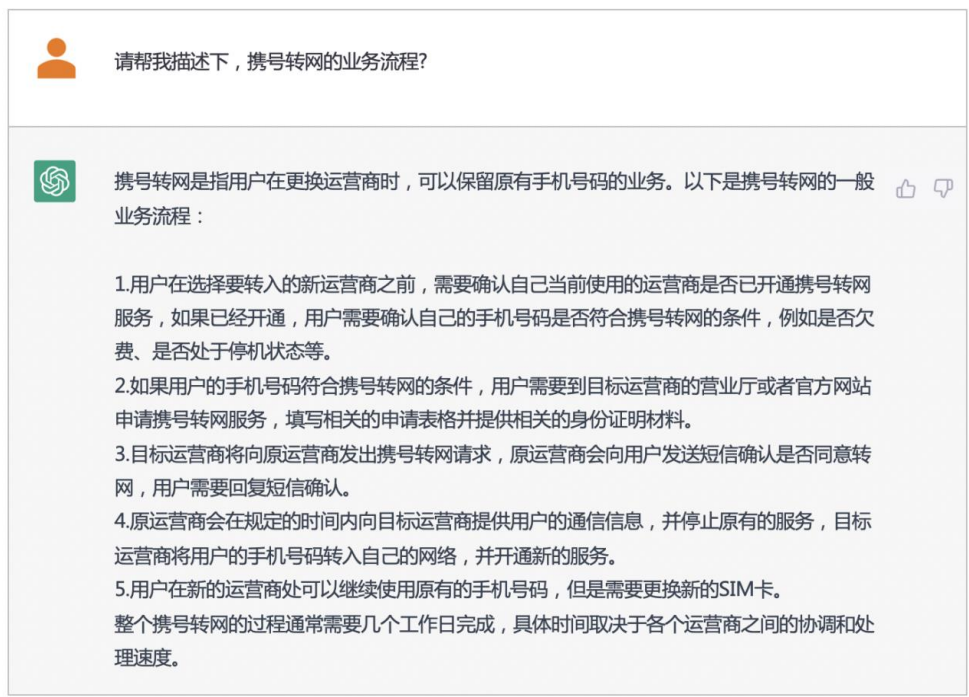

**图 3-2 GPT 大模型帮助分析业务需求**

**场景二： GPT 大模型辅助原型设计**

在设计阶段，帮助设计师快速构建设计原型和模型，可适用的场景主要有：

1. 设计原型：采用设计原型工具，如 Sketch、Figma、Axure 等，快速构建交互原型，提高设计效率和设计质量。设计原型可以在设计初期帮助设计师与相关方进行充分的沟通和交流，减少后期的设计修改，提高设计效率。引入 GPT 大模型可以提升快速原型构建的能力。GPT 大模型

可以生成交互原型，提高原型构建效率。设计师可以将自然语言描述输入 GPT 大模型，生成相应的交互原型，避免手动构建原型的繁琐和耗时，同时提高原型的质量和准确性。同时，GPT 大模型也可以实现智能化辅助设计，通过理解自然语言描述，生成相应的设计方案或 UI 元素，提升设计效率和准确性。设计师可以通过与 GPT 大模型交互，快速得到可行的设计方案或 UI 元素，避免重复的设计过程。

2. 可视化建模：采用可视化建模工具，如 UML，快速建立模型，简化建模过程，提高建模效率。采用可视化建模可以将抽象的概念用可视化的方式呈现出来，便于沟通和交流，同时可以减少重复的设计和提高设计效率。GPT 大模型可以实现自动化原型验证。设计师可以将设计方案或 UI 元素交给 GPT 大模型进行测试用例的生成，避免手动编写测试用例的繁琐和耗时，同时提高测试的覆盖率和准确性。

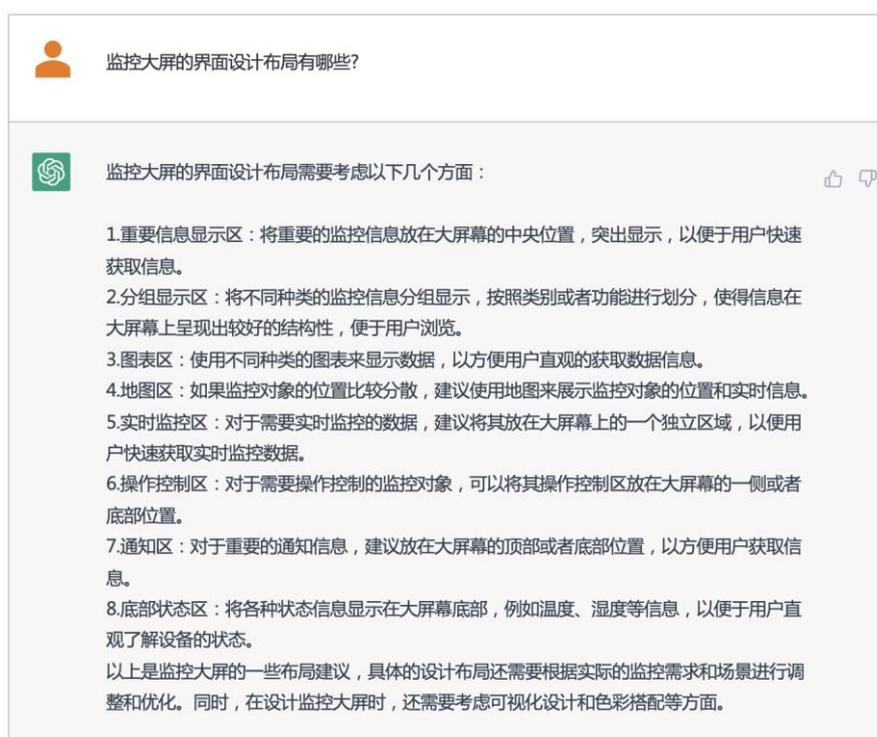

**图 3-3 GPT 大模型辅助界面设计**

设计人员，借助 GPT 大模型，可以根据输入的需求，快速生成界面原型，原型快速验证，确保需求场景覆盖完善。从原先以周为单位，降低到以小时为单位，人员效能提升 100 倍以上。

**场景三：GPT 大模型助力智能编码**

在开发编码阶段，当前通过采用集成开发环境、使用开发框架和库、采用代码审查等措施，能够提高研发效率，减少重复工作，降低代码出错率，提高代码质量和可维护性。可适用的场景主要有：

1. 使用集成开发环境（IDE）和开发框架：使用集成开发环境可以提高开发效率。IDE 通常集成了代码编辑器、调试器、自动补全、版本控制等工具，使得开发人员可以更加高效地编写代码、测试代码和维护代码。使用开发框架和库可以减少开发人员编写重复代码的工作量，提高开发效率。开发框架和库如 Spring、React、Vue 等，可以提供一系列的组件和函数，帮助开发人员更快地构建应用程序。GPT 大模型未来甚至可以颠覆开发编码的工作方式，即通过自然语言编程接口，让开发人员使用自然语言描述所需功能，然后将其转化为代码实现，降低代码实现的门槛和难度，提高开发效率。

2. 代码审查：采用代码审查可以帮助开发人员发现代码中的问题，提高代码质量和开发效率。代码审查可以由其他开发人员或专业的代码审查工具进行，发现问题后及时进行修改和优化。GPT 大模型可以自动代码补全和纠错、智能重构和优化，大大提升代码审查的效率。GPT 大模型可

以根据已有的代码和上下文，提供自动代码补全和纠错的功能，帮助开发人员更快速地编写代码，并避免常见的语法错误和拼写错误。GPT 大模型可以分析代码并提供智能重构和优化的建议，帮助开发人员优化代码结构、提高代码可读性和可维护性，从而减少后期的维护成本。

3. 代码生成：作为代码生成工具。Copilot 是由 OpenAI 和 Github 合作研发并商用的基于 GPT 模型的代码自动补全工具，可以根据上下文和用户输入自动生成代码片段。Copilot 可以快捷的集成到自己的 IDE 中，以便更高效地编写代码。GPT 大模型可以根据给定的场景或需求，自动生成相关的代码文本，例如自动生成 API 文档、自动生成注释等，减少手动编写文档的时间和精力。与 Copilot 不同，GPT 大模型更注重用户和模型的交互，它可以与用户进行多轮对话，通过询问和解释来更好地理解用户的意图。在代码开发方面，GPT 大模型可以根据用户的需求和上下文生成相应的代码，并可以与用户进行交互以获取更多信息和反馈。GPT 大模型的优势在于可以生成更加个性化和灵活的代码，但是也需要更多的交互和反馈，对用户的技能和经验要求更高。

开发人员，通过云 IDE 快速拉起预置的场景化的应用开发环境，开发小组内通过工作空间的共享，实现开发的快速协作，如结对编程、远程联调、单元测试等。GPT 大模型智能辅助编程，结合代码辅助生成的能力，通过聊天的方式，轻松完成代码编写，输出高质量的业务代码。懂代码，更懂业务。数字人助手和鼓励师，结合程序员的编码习惯，情绪等，不定时的通过鼓励的方式，调动程序

员工作的积极性。编码效率从以天为单位，降低到以小时为单位，甚至分钟级的完成开发，开发人员更聚焦核心代码编写，效能提升 100 倍以上。

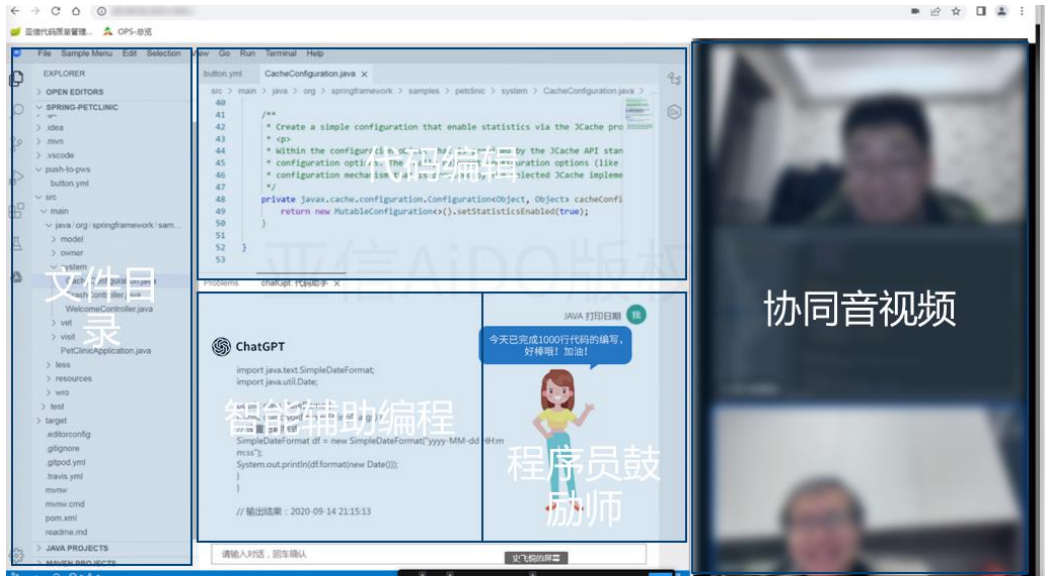

**图 3-4 引入 GPT 大模型后开发人员日常工作场景**

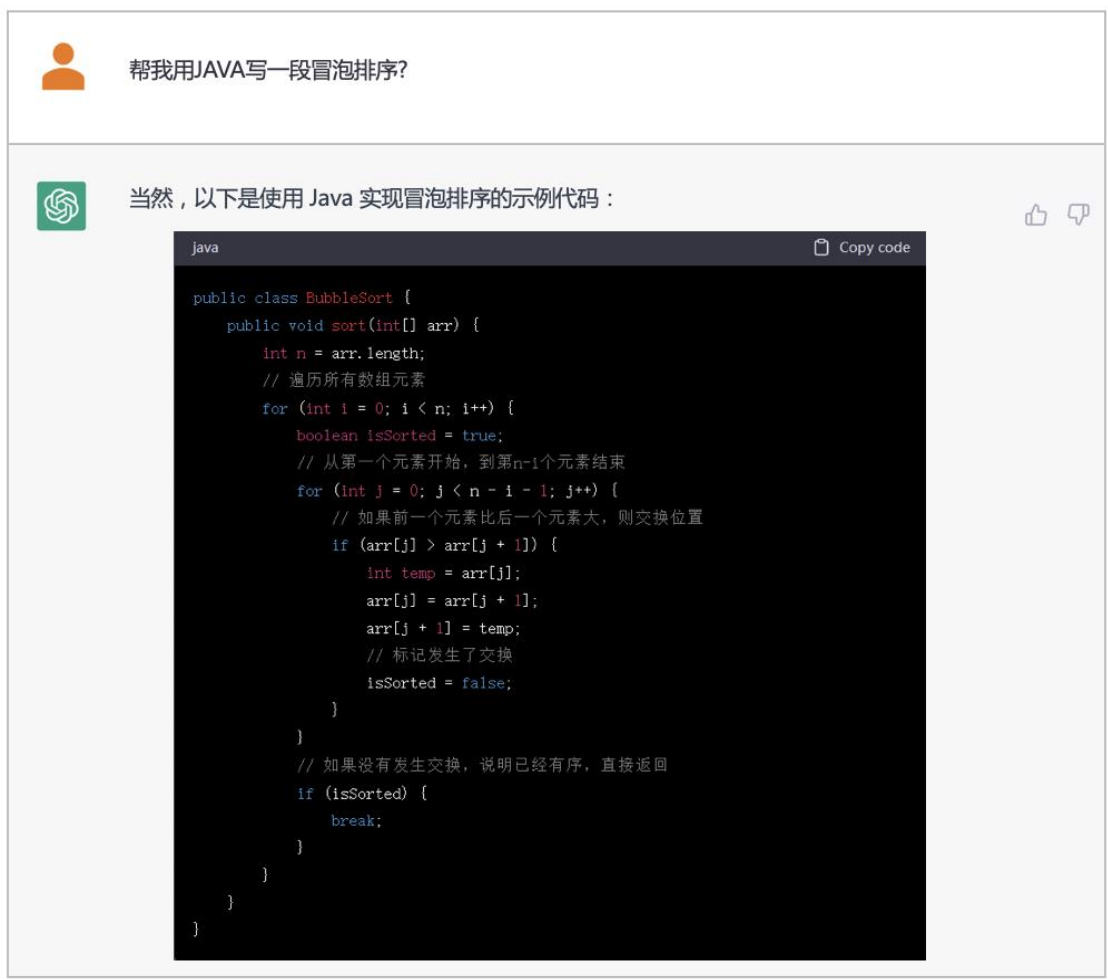

**图 3-5 GPT 大模型智能辅助编码**

### 场景四：GPT 大模型赋能持续集成与持续交付

在集成与发布阶段，当前借助自动化构建和测试、持续集成和持续交付、持续部署等措施可以在集成和发布阶段提高研发效能，从而加快软件交付的速度和质量。引入 GPT 大模型在可以带来进一步的提升。可适用的场景主要有：

1. 自动化构建和测试：通过使用自动化构建和测试工具，可以快速、可靠地构建和测试软件，减少手动操作和人为错误，从而提高软件交付的质量和效率。GPT 大模型可以用于自动化测试中，自动生成测试用例和测试数据，并自动执行测试用例，从而减少测试人员的工作量，提高测试的覆盖率和准确性。

2. 持续集成和持续交付：通过持续集成和持续交付的实践，可以实现快速、频繁地将软件集成和交付给用户，从而及时地获取反馈，提高软件的质量和响应能力。GPT 大模型可以用于自动化部署中，自动生成配置文件和部署脚本，并自动执行部署过程，从而减少手动操作和人为错误，提高部署的速度和可靠性。

3. 持续部署：通过使用容器化部署技术，可以将软件打包成轻量级、可移植的容器，并在不同的环境中快速、一致地部署和运行，从而提高软件的可靠性和可移植性。GPT 大模型可以用于预测性维护中，通过分析日志和监控数据，预测系统可能出现的问题并提出解决方案，从而避免系统故障和停机时间的损失，提高系统的可靠性和稳定性。

平台人员，借助 GPT 大模型，无需进行复杂的构建、部署配置，集成和发布活动变得更加轻松。应用的构建部署配置从以天为单位，降低到以小时为单位，集成和部署效率提升 10 倍以上。

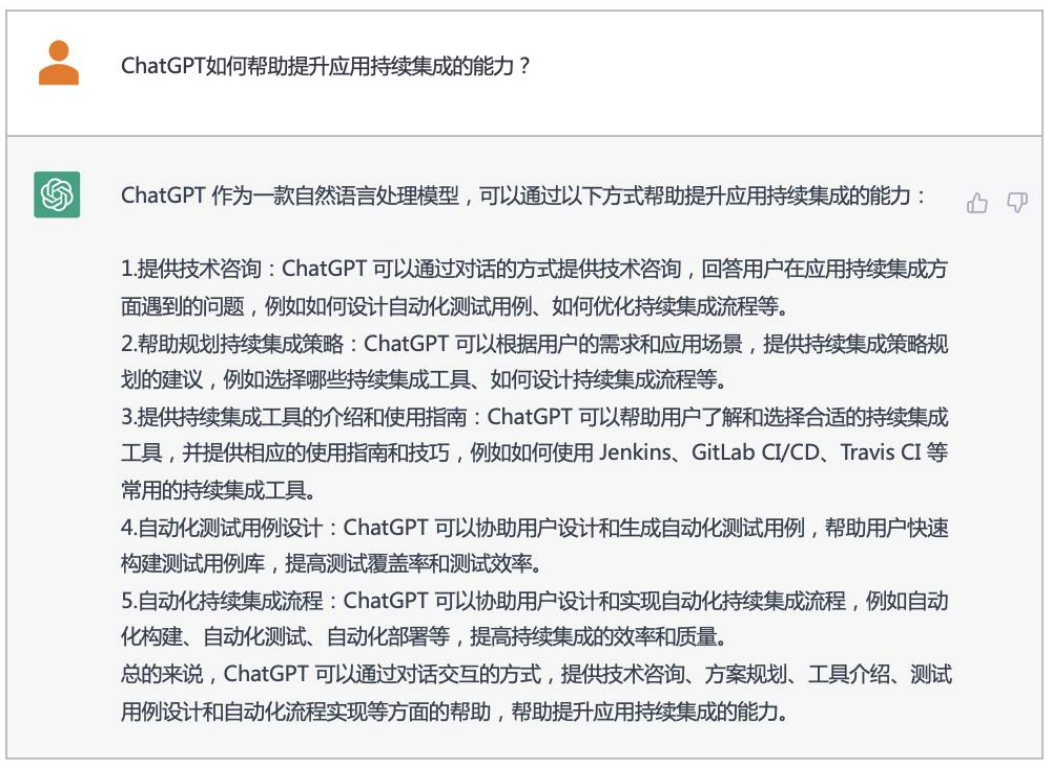

**图 3-6 GPT 大模型提升持续集成效能**

**场景五：GPT 大模型提升自动化测试效能**

在测试阶段，自动化测试工具（例如接口、UI、性能等），以及代码染色、流量复制等自动化测试能力，一定程度上提升了测试的效能。引入 GPT 大模型可以通过自动生成测试用例、智能缺陷检测、智能精准测试、以及智能代码染色，大幅提升测试效能。可适用的场景主要有：

1. 自动化测试：使用自动化测试工具进行自动化测试，包括单元测试、集成测试、功能测试、性能测试、接口测试、UI 测试等，可以减少测试时间，提高测试覆盖率和准确性。GPT 大模型可以帮助自动生成测试用

例，特别是在一些较为简单的场景中，可以避免手动编写重复的测试用例，提高测试的自动化程度，从而提高测试效率。GPT 大模型可以通过学习历史缺陷数据，识别并智能检测出潜在的缺陷，减少漏测和误测的情况。同时，GPT 大模型还可以根据缺陷类型和严重程度进行排序，提高缺陷修复的优先级和效率。GPT 大模型可以基于领域知识和历史测试数据，辅助测试人员制定更加精准的测试计划，包括测试范围、测试用例设计、测试覆盖率等方面，以提高测试的效率和覆盖度，减少漏测和误测的情况。

2. 代码染色：指利用静态代码分析工具，对代码进行自动化检查和诊断，以发现潜在的问题和漏洞。静态代码分析工具可以帮助开发人员快速发现代码中的错误和安全隐患，同时可以提供详细的代码质量报告，帮助开发人员优化代码质量和可维护性。此外，代码染色还可以结合自动化测试工具，自动化执行测试用例，并对测试结果进行反馈和调整，从而提高测试效率和准确性。GPT 大模型可以分析代码库中的代码结构、变量名、注释等信息，为测试人员提供更智能的代码染色功能，相比于基于传统静态代码扫描的代码染色，可以让代码可读性更高，使得测试人员更容易理解和编写测试用例，提高测试效率。

测试人员，借助 GPT 大模型，实现自动生成测试用例，智能代码染色等，让测试人员更容易理解和编写测试用例。

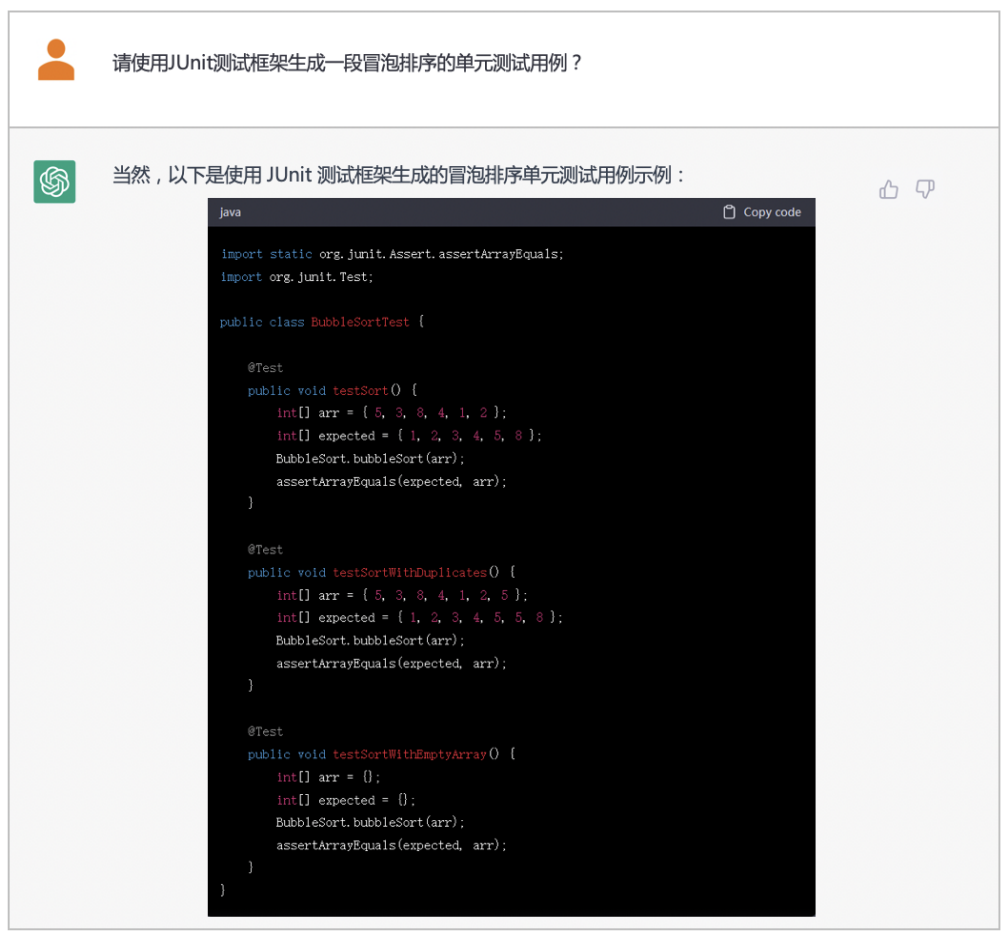

**图 3-7 GPT 大模型快速生成单元测试用例**

综上所述，通过引入 GPT 大模型，可以对软件研发全生命周期全面赋能，无论是需求分析阶段的需求沟通，还是设计阶段的原型设计，亦或开发人员的编码，测试人员的测试执行，都能够有较大程度的效能提升。

## 3.1.2 智能运维

智能运维 AIOps（Artificial Intelligence for IT Operations）是一种基于人工智能技术的运维自动化解决方案，它利用机器学习、深度学习、模式识别等人工智能技术，对 IT 系统的运行数据（包括日志、指标、事件）进行分析和处理，自动识别和分析故障，快速定位问题，并自动推荐或执行解决方案，以提高 IT 运维团队的效率和响应速度。

智能运维 AIOps 作为一个新兴的技术领域，目前正在快速发展，越来越多的研究者将机器学习、自然语言处理、数据挖掘等技术应用于智能运维 AIOps 领域。学件是一种性能良好的预训练机器学习模型，其具有一套解释模型意图和/或特性的规约，亚信科技 AIOps 运维学件已经在智能运维领域得到了广泛应用和快速发展，在通信领域已经沉淀出了像指标异常检测、日志异常检测等可复用、可演进的成熟运维学件。此外，早在 10 年前 Github 提出了 ChatOps 的概念，它是一种实时聊天驱动型的运维模型，可以"用聊天的方式来做运维工作"，通过将机器人植入聊天会话，形成人、机器、数据的自动化、透明化的联动，使运维团队能够高效执行任务和沟通协作，可以认为 ChatOps 是 GPT 大模型在运维领域的一个雏形。

目前随着 GPT 大模型及其相关技术的出现和普及，可以在智能运维领域进一步加强"ChatOps"的运维能力。下图提供了智能运维 AIOps 工作台中的运维协作、智能运维相关问题的快速问答以及运维人员确认后的快速执行等场景。GPT 大模型与智能运维 AIOps 可以协同，面向运维领域提供运维知识问答和智能工单分析，助力运维人员快速识别系统故障类型、高效提供故障处置策略并生成自动化脚本实现故障自愈。

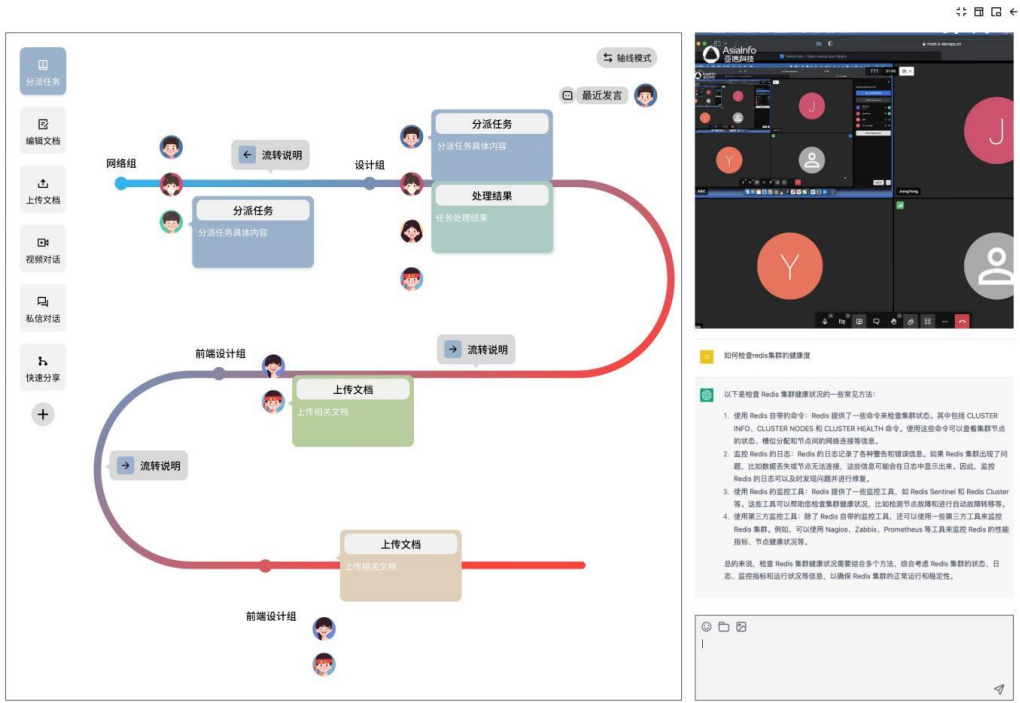

**图 3-8 智能运维工作台运维协作场景**

### 场景一：快速识别故障类型

当系统出现故障时一般会伴有很多错误类型的日志输出。在智能运维领域，目前主要是通过机器学习、聚类、频繁模式挖掘、NLP 等技术对日志的模式进行识别以及检测日志量的异动情况来进行日志的异常检测。

GPT 大模型基于对日志模式的识别能力，可以帮助运维人员提供系统故障关键日志的分析。运维人员可以将系统故障对应的关键日志和相关信息发送给 GPT 大模型，GPT 大模型通过 NLP 自然语言处理能力尝试识别和分析日志的问题，并给出日志对应的故障类型及原因。如将一段 Kafka 的错误日志：

```
"2022-01-25 09:09:14,930 | FATAL | [main] |  | kafka.Kafka$ (Log-
ging.scala:113)
java.lang.IllegalArgumentException: requirement failed: rep-
lica.fetch.max.bytes should be equal or greater than message.max.bytes
        at scala.Predef$.require(Predef.scala:233)
        at kafka.server.KafkaConfig.validateValues(KafkaConfig.scala:959)
        at kafka.server.KafkaConfig.<init>(KafkaConfig.scala:944)
        at kafka.server.KafkaConfig$.fromProps(KafkaConfig.scala:701)
        at kafka.server.KafkaConfig$.fromProps(KafkaConfig.scala:698)
```

```
        at kafka.server.KafkaServerStartable$.fromProps(KafkaServerStarta-
ble.scala:28)
        at kafka.Kafka$.main(Kafka.scala:60)
        at kafka.Kafka.main(Kafka.scala)"
```

发给 GPT 大模型，经 GPT 大模型识别分析后发现是 Kafka 配置的错误并给出错误的原因，帮助运维人员快速发现定位该类型的问题。进一步的 GPT 大模型可以与智能运维 AIOps 中的日志异常检测等运维学件能力进行结合，如将 GPT 大模型识别的故障类型作为运维专家经验与学件中智能算法检测的结果进行融合，以数、知双驱动能力提升运维效率。

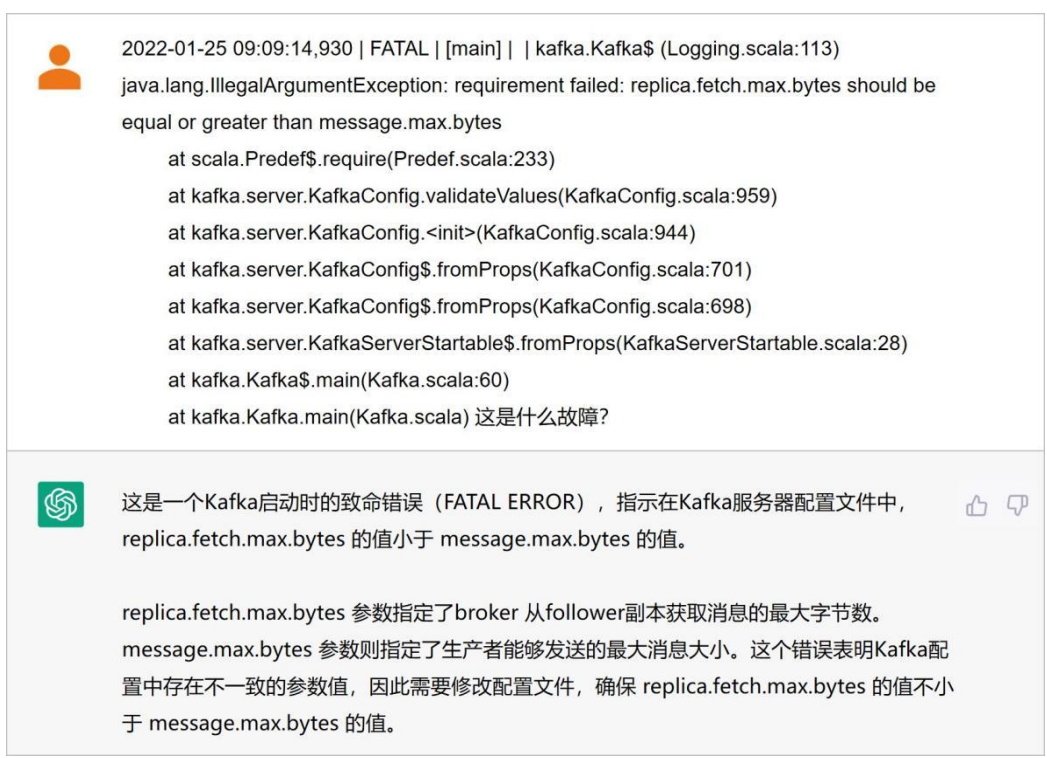

图 3-9 GPT 大模型帮助分析故障

### 场景二：高效提供处置策略

运维过程中当故障发生后，先经过分析诊断定位故障的根因，然后如何快速的找到故障修复的方案是至关重要的。在智能运维领域目前的实现思路是希望提供行业知识图谱或运维知识库，但由于运维经验相对分散不集中导致运维知识的

沉淀不足，大多数情况下还需要依靠专业的运维人员通过自身的经验来提供故障的解决方案，对人员的依赖程度较高。

GPT 大模型基于大量运维知识的学习与沉淀，可面向运维人员提供运维行业知识库能力，运维人员可以将故障的类型或故障的描述发送给 GPT 大模型，GPT 大模型通过其 NLP 自然语言理解能力和丰富的运维知识库能力提供相应故障 的 解 决 方 案 或 修 复 策 略， 如 可 以 直 接 提 问"Kafka 配 置 错 误 replica.fetch.max.bytes should be equal or greater than message.max.bytes，如何修复"， GPT 大模型会给出解决该问题的方法及相关步骤，以帮助运维人员对该类故障作出快速响应。另外 GPT 大模型支持多国语言，对一些中文运维知识库不能覆盖的部分，可以查找英文或其他语言的运维知识库。

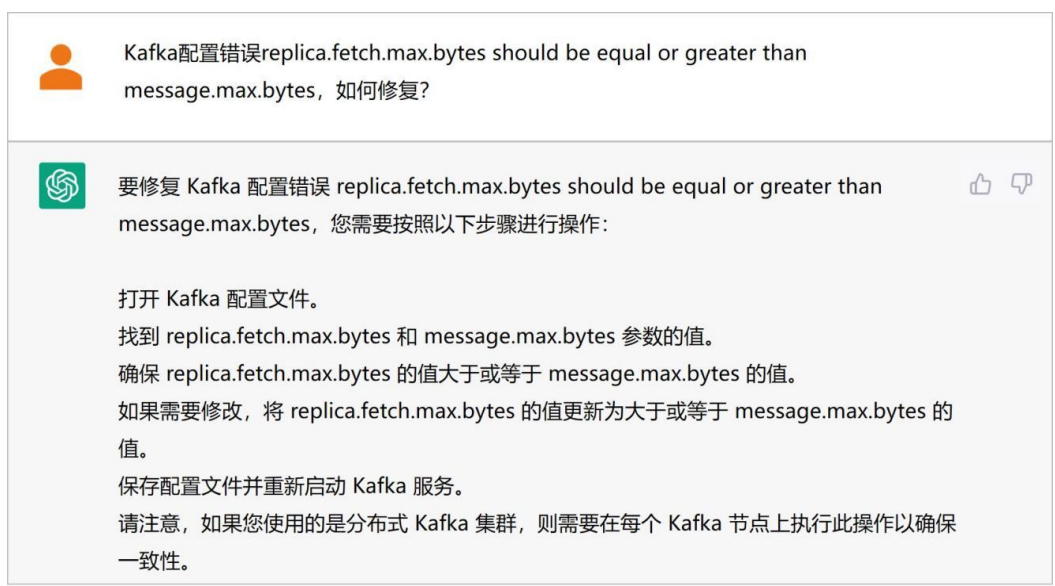

**图 3-10 GPT 大模型智能修复 Kafka 配置错误**

**场景三：故障自愈**

在智能运维领域，当发现并识别出故障类型后，需进一步结合专家经验或运维知识库等匹配对应故障的处置策略，然后如何能快速的达到故障自愈的目的，一般来说需要运维人员编写一些程序或脚本来完成故障的自动修复。

GPT 大模型可以自动生成脚本以及 ansible 等自动化运维工具需要的程序代码。如结合上述 Kafka 的错误信息自动生成 ansible 自愈程序的代码，帮助运维人员能快速恢复故障，实现故障自愈。

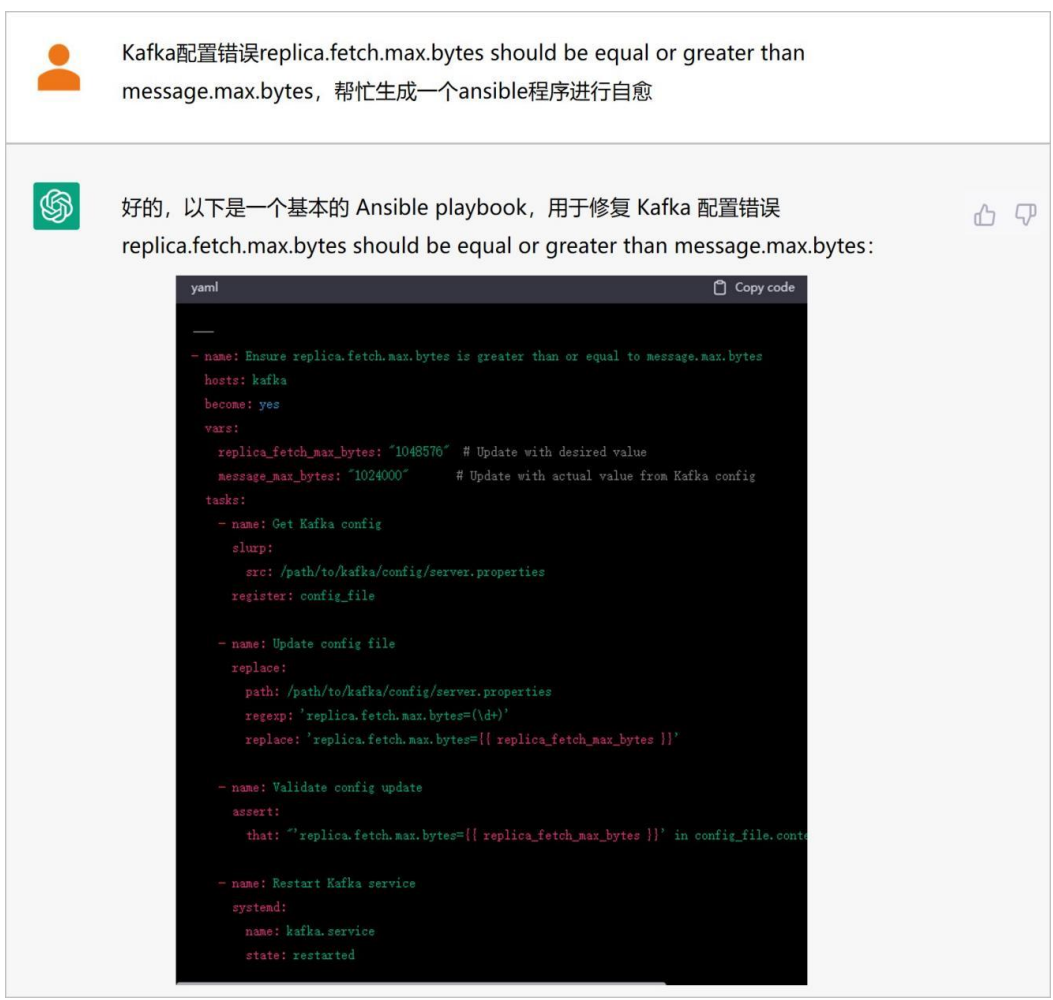

**图 3-11 GPT 大模型智能生成 ansible 程序代码**

**场景四：运维知识问答**

运维过程中积累了大量的运维数据，GPT 大模型可快速学习并融合运维知识，推理总结问题形成运维知识库，运维人员将运维相关问题输入到 GPT 大模型大模型，大模型通过文本识别准确理解用户意图，并智能化生成相关应答内容，提升运维效率。如针对通用运维组件 kubernets 出现的问题时，kubernets pod

出现 CreateContainerConfigError 错误，如何调试和解决?GPT 大模型会给出相应运维组件问题的答案。

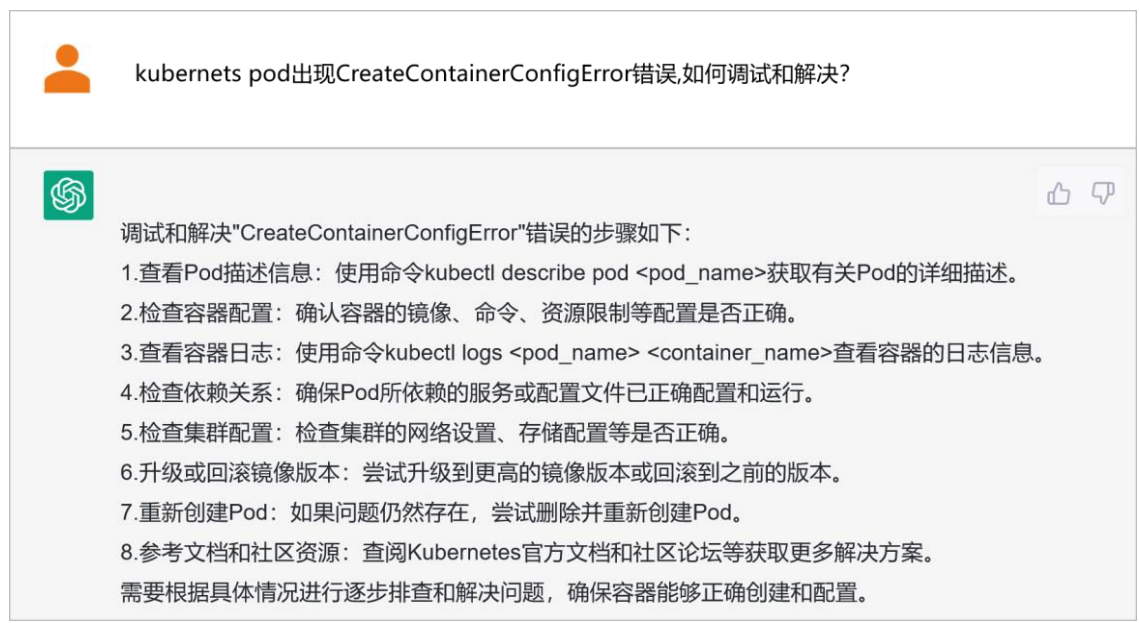

**图 3-12 GPT 大模型运维知识问答**

## 场景五：智能工单分析

运维过程中运维人员需要花费大量的时间对各类工单进行分析和处理，GPT 大模型大模型可以对历史工单数据进行知识抽取、挖掘、总结为工单知识网络，从而智能化实现类似工单推荐、工单合并推荐、工单解决方法等场景，提高工单处理效率。

如针对新的工单 "当 AntDB 节点空间超过 90%告警时，应该怎么解决？"，GPT 大模型会基于历史上类似的工单数据给出该工单的处理建议或策略。

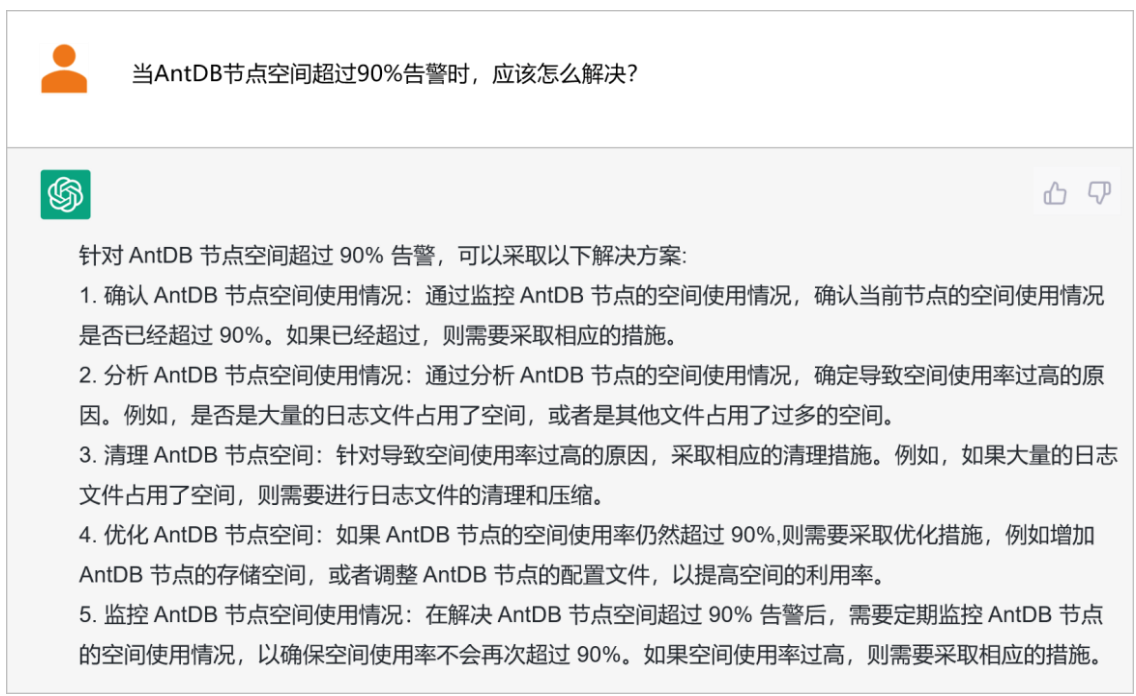

**图 3-13 GPT 大模型智能工单分析**

## 3.1.3 智能营销

营销系统是指一套针对企业销售和市场营销活动进行规划、执行和监控的系统化工具和流程。营销系统的目标是提高营销效率、减少浪费、提高收益。通过数据分析和自动化流程，营销系统可以帮助企业更好地了解其客户，定制个性化的营销方案，并跟踪和分析营销活动的结果，以便随时调整和改进营销策略。智能营销是利用 AI 技术和大数据分析来优化市场营销策略的方式，将智能的技术与营销策略相结合，实现更高效、精准和个性化的营销。

GPT 大模型之前，关于智能营销方面的问题，如果要寻求网络的帮助，只能就一些关键词进行搜索，再对根据关键词搜索到的结果进行甄别分析，看是否有些有助于制定或实施营销的内容。GPT 大模型可以理解自然语言，能够理解输入的背景信息，并准确了解整体需求，因而可以结合背景和需求，针对性地找到解决方案或建议。可以在产品策划、营销方案策划、主动营销、产品订购、营

销分析等营销相关的整个闭环场景中帮助通信企业实现更高效、更精准和个性化的营销。

### 场景一： 基于 GPT 大模型的产品策划增强

GPT 大模型可以帮助收集和分析市场数据：在产品策划过程中，了解市场需求是至关重要的。GPT 大模型可以帮助企业收集和分析大量的市场数据，包括竞争情况、目标用户、市场趋势等等。这些数据可以为企业的产品决策提供基础和参考。

GPT 大模型可以提供创意和灵感：在产品策划过程中，创意和灵感是非常重要的。GPT 大模型可以生成有关市场和用户的文本信息，这些信息可以激发企业的创意和灵感，帮助企业创造出更有吸引力的产品。

产品经理根据初步的产品创意想法，让 GPT 大模型列出业界相关的产品及特点，结合相关统计数据（GPT 大模型同时也可参与数据统计分析），不断加入自己的要求，和 GPT 大模型一起头脑风暴，调整优化产品的创意，形成产品策划的初步方案。

泛在智能感知被广泛认为是 6G 的一个重要特性。在居家环境中，可以用无线信号非侵扰地获悉人员的健康状况和生活规律：如老人跌倒不起等异常事件发生，都可以通过无线信号识别后再发送给家人、医院、社区护理人员等。运营商如果想根据 6G 和 Wi-Fi 智能感知能力，策划一个老人跌倒监控相关的产品，可以让 GPT 大模型提供产品策划方面的一些思路：

泛在智能感知被广泛认为是 6G 的一个重要特性，随着 Wi-Fi 感知新标准和 6G 的到来，无线感知应用将走进千家万户，Wi-Fi、6G 都同时具备通信、感知的功能。在居家环境中，无论是手机还是家电都有通信感知的功能，可以用无线信号非侵扰地获悉人员的健康状况和生活规律：如老人何时起床，何时三餐，何时外出和回家，每天是否生活规律，在家活动量够吗？有无呼吸心跳异常、睡眠不好、跌倒不起等异常事件发生？都可以通过无线信号识别后再发送给家人、医院、社区护理人员等。如果我是一个产品经理，现在想采用wifi和6G的通信感知能力，策划老人跌倒监控报警方面的产品，请问该产品起一个什么名字比较好，如何策划该产品呢？

针对您的需求，我能为您提供以下一些建议：

1、产品名称：您可以考虑一个简短、易记、具有代表性的名称。例如，"智能老年监护"、"康安家居"、"智能家庭护士"等。

2、产品定位：针对老年人跌倒监测的市场需求，您可以将产品定位为一款便捷、实用的老年人跌倒监控报警器，重点突出其对老年人的健康和安全提供可靠保障的功能。同时，强调产品采用无线信号识别和6G通信感知技术，实现无缝监控老年人的日常生活状态，并通过智能报警功能及时通知家人或护理人员。

3、产品功能：该产品的主要功能可以包括跌倒检测、心率、呼吸、体温等生理参数监测、智能报警、远程视频监控等。同时，您还可以考虑为老年人提供一些贴心服务，如定时提醒、健康咨询、社交互动等，为老年人提供全方位的健康保障和生活支持。

4、市场推广：您可以考虑在家庭保健、老年健康等领域寻找合适的推广渠道，如医疗机构、社区服务中心、家庭护理公司等，同时在社交媒体、线上商城等渠道上推广产品，吸引潜在用户的关注。

5、用户体验：为了让老年人更方便地使用该产品，您可以考虑为其设计用户友好的界面和操作流程，采用语音提示、远程遥控等交互方式，使老年人更容易接受和使用。同时，您还可以为用户提供优质的售后服务和技术支持，满足用户的需求和反馈，提高产品的用户满意度和口碑。

图 3-14 如何策划产品的回答

根据 GPT 大模型提供的建议，可以进一步沟通，让 GPT 大模型对产品功能设计提供进一步建议。GPT 大模型在下图 3-15 回答中，提到了智能学习用户习惯、用户隐私保护等在产品设计中容易忽略的问题，可以帮助产品经理完善产品的设计。

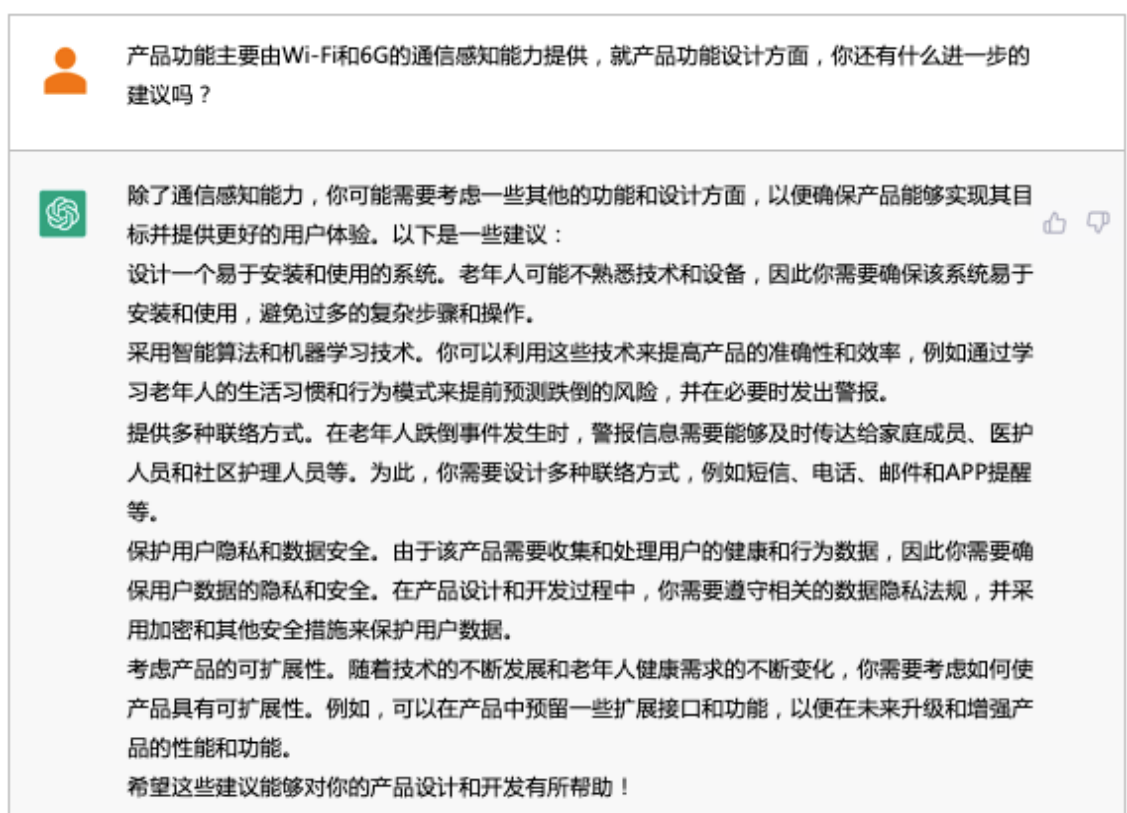

**图 3-15 产品功能设计的进一步建议**

**场景二：基于 GPT 大模型的营销方案策划增强**

类似于产品策划场景，营销经理让 GPT 大模型列出业界关于此类产品的营销特点与优选的方案列表，结合相关产品销售和使用统计数据（GPT 大模型同时也可参与数据统计分析），分析比较各种营销方案的优劣，并和 GPT 大模型一起头脑风暴，形成营销方案策划的初步方案。GPT 大模型也可以生成文本和语音内容，包括文章，视频，音频等，为企业提供内容营销的基础。

比如可以请 GPT 大模型帮助设计与产品相关的广告词：

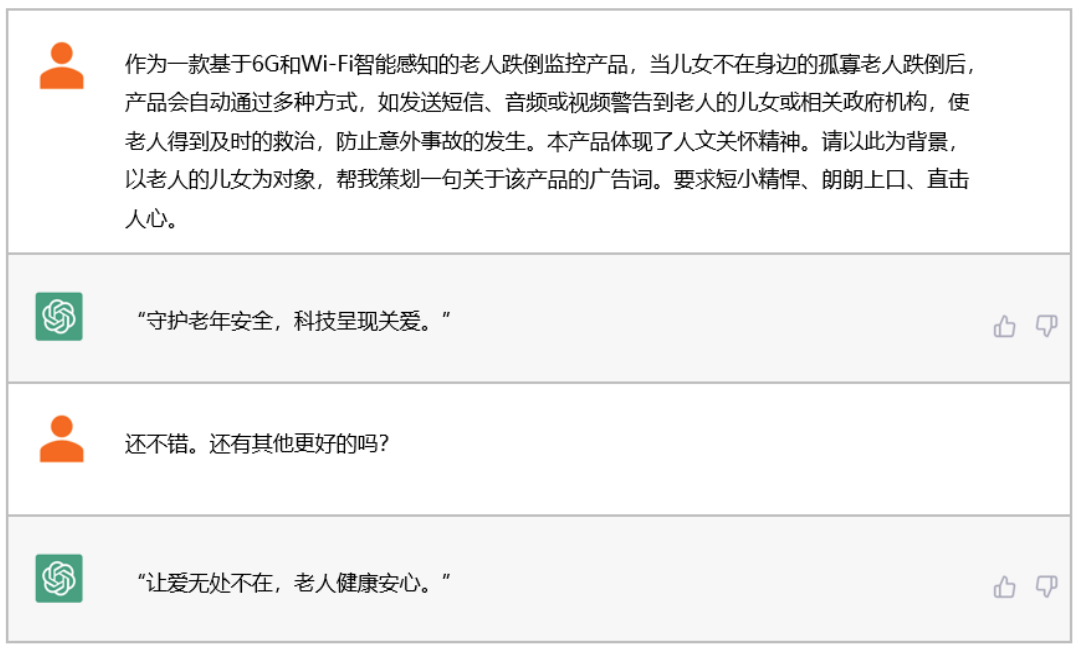

**图 3-16 关于产品广告词的问答**

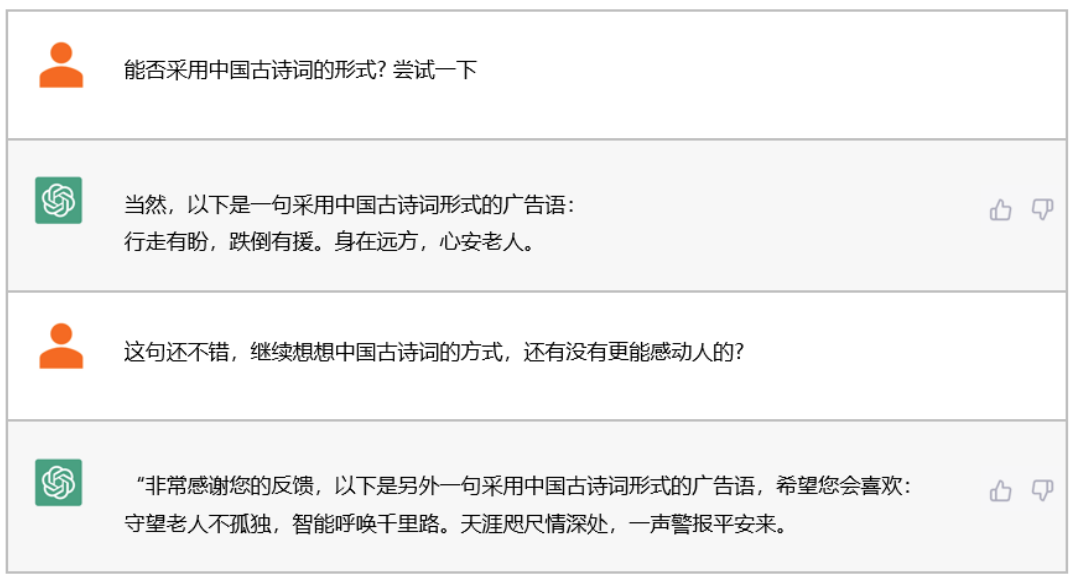

**图 3-17 关于产品广告词的古诗词形式的问答**

**场景三：基于 GPT 大模型的主动营销增强**

GPT 大模型可以处理大量的文本数据，并通过自然语言处理和机器学习技术提取有用的信息和趋势。这可以帮助企业了解客户需求和行为。GPT 大模型

可以集成到营销自动化平台中，根据用户行为和需求提供个性化的推荐和建议，增强客户参与度，提高转化率。客户和 GPT 大模型在聊天过程中，可结合以往的聊天记录，根据客户和运营商的交互历史，综合分析客户需求，主动寻找商机，并适时向客户推荐合适的产品：

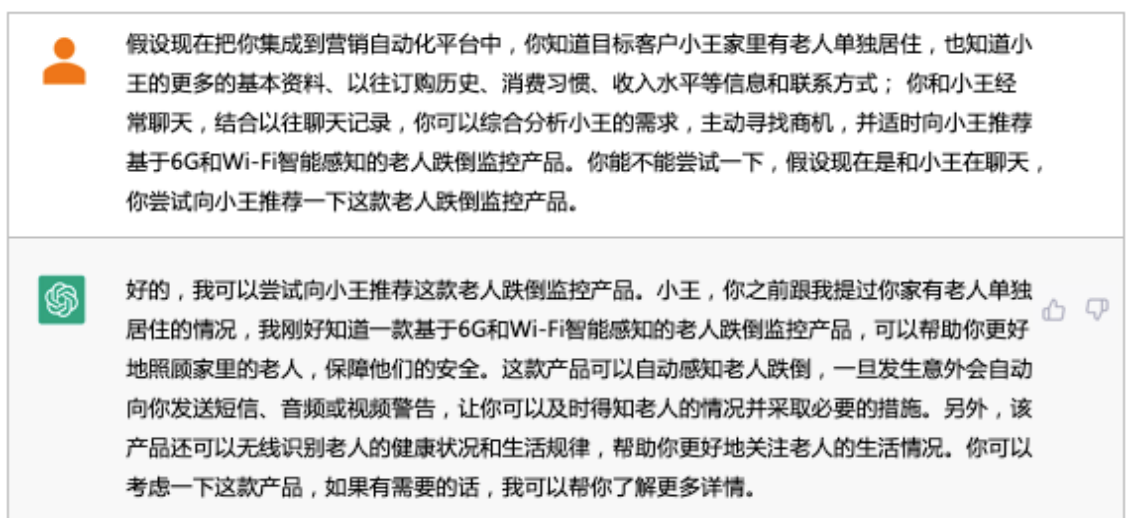

**图 3-18 关于主动营销推荐的问答**

**场景四：基于 GPT 大模型的产品订购增强**

客户不再需要和客户经理、话务员、营业前台等各个渠道的不同运营商员工打交道，GPT 大模型成为客户在运营商的专职助理。客户用自然语言向 GPT 大模型提出产品需求，就像和朋友聊天一样，聊聊当前手机的最新机型，有什么特点，有什么好的应用，网络流量需求等等，GPT 大模型会根据客户需求提出可选的方案，比较每种方案的优缺点，并根据客户实际情况，推荐最优方案。客户确认后 GPT 大模型可以帮助下单订购，并随时询问订单进展。下图是在销售过程中 GPT 大模型帮助客户进行产品比较、推荐最佳方案、帮助订购的示例。可以看到，GPT 大模型抓住了运营商这款产品的非侵入式、实时监控、自动报警等其他产品不具备的特点。

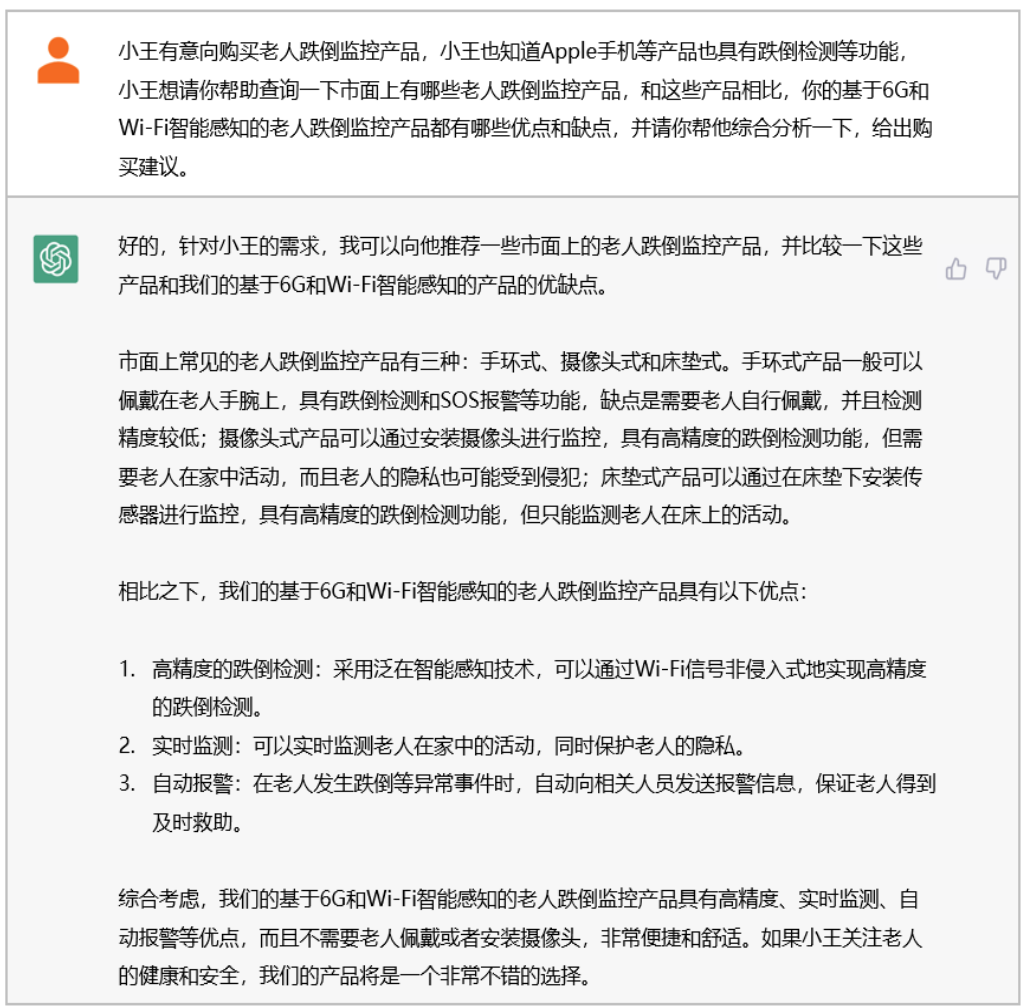

**图 3-19 关于不同产品特性分析比较的问答**

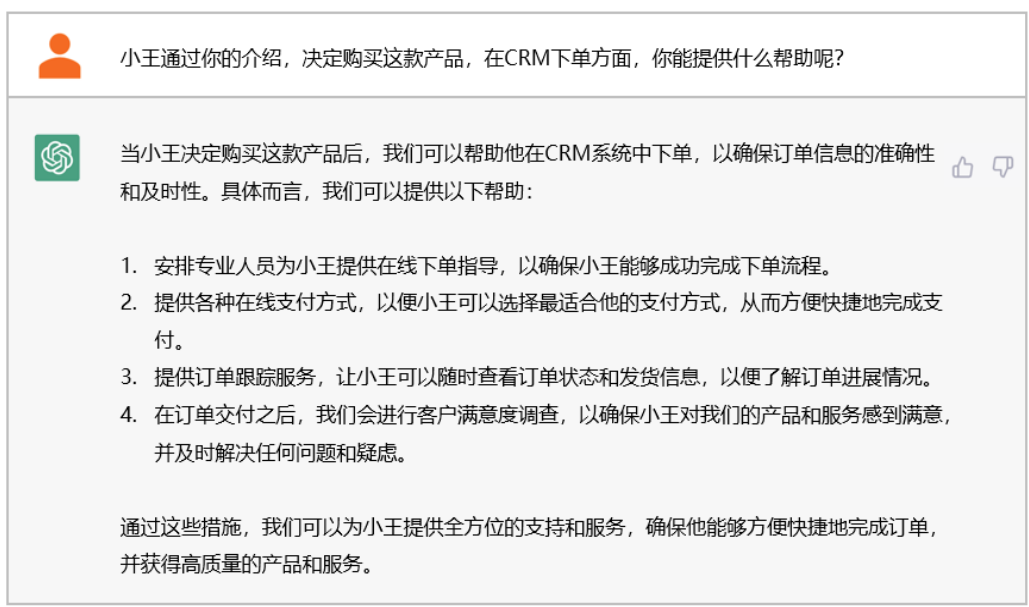

**图 3-20 关于帮助客户产品下单的建议**

**场景五：基于 GPT 大模型的营销分析增强**

GPT 大模型可以在以下方面帮助进行营销分析：

- **收集数据**：搜集产品销售数据和其他相关数据，如推广方式、渠道、竞争对手的营销活动等。

- **分析数据**：对收集到的数据进行分析，包括统计销售数据、计算 ROI、分析销售趋势、比较销售数据与预期等。

- **评估营销策略**：根据分析结果，评估各个营销策略的效果，看哪些策略比较成功，哪些策略需要优化或调整。

- **制定新的营销计划**：根据评估结果，制定新的营销计划和战略，包括改进现有的营销策略、增加新的营销渠道和方式等。

- **迭代优化**：根据新的营销计划，再次进行营销活动，并收集数据和分析结果，以此不断优化营销策略，从而提升产品的销售情况。

通过以上的分析方法，可以对产品的销售情况进行全面的评估和优化，从而提升产品的市场占有率和竞争力。

以下是请 GPT 大模型帮助进行营销分析的示例：

这款基于6G和Wi-Fi智能感知的老人跌倒监控产品，经过最近一个月的销售，发现商机到订单的转化率不高。假设你能够拿到所有的营销数据，你也能进行统计分析。请你帮助分析一下，最近的营销策划和销售策略可能在哪些方面出现了问题，如何提升商机到订单的转化率。

为了更好地分析最近营销策划和销售策略可能出现的问题，我们需要了解一些关键指标，例如：

1、营销活动的曝光率和点击率：这可以帮助我们确定营销活动是否能够吸引潜在客户的注意力，并让他们进一步了解我们的产品。

2、销售团队的拜访率和转化率：这可以帮助我们了解销售团队是否足够努力，以及他们的销售技巧是否足够高超，以便将商机转化为订单。

3、销售流程的效率和客户体验：这可以帮助我们确定我们的销售流程是否顺畅，以及我们的客户是否在购买过程中遇到了任何障碍。

4、竞争对手的价格和功能：这可以帮助我们确定我们的产品在市场上的竞争力，并确定是否需要调整价格或添加新功能。

通过分析以上指标，我们可能会发现以下一些问题和解决方案：

1、营销活动的曝光率和点击率低：我们可以通过使用不同的营销渠道（例如社交媒体，广告，SEO等）来增加曝光率，并优化营销材料以提高点击率。

2、销售团队的拜访率和转化率低：我们可以为销售团队提供更多的培训和支持，以帮助他们提高销售技巧，同时我们也可以考虑提供更具吸引力的销售奖励和福利，以激励销售人员更努力地工作。

3、销售流程的效率和客户体验低：我们可以优化销售流程，简化购买流程，并提供更好的客户支持和服务，以提高客户满意度和忠诚度。

4、竞争对手的价格和功能更具吸引力：我们可以考虑调整产品价格或增加新功能，以提高产品的竞争力。我们也可以针对竞争对手的优势进行针对性的营销宣传，以突出我们的产品在某些方面的优势。

通过上述解决方案，我们可以逐步提高商机到订单的转化率，并将产品销售情况提高到更好的水平。

**图 3-21 关于产品营销分析的问答**

综上所述，GPT 大模型可以在产品策划、营销方案策划、主动营销、产品订购、营销分析等整个闭环中帮助通信企业实现更高效、更精准和更个性化的营销。

### 3.1.4 智能计费

计费系统作为电信运营企业核心 IT 系统，承担了从用户使用记录采集与预处理、计费、帐务处理与管理、多方结算、风险控制，以及业财合拢等核心工作，

是确保电信企业健康运营的基础设施。从计费系统服务的业务角度来看，计费系统需要满足电信自有按量计价的语音、短信、流量等业务的费用处理，订阅制与点播制的云计算、长途租用、内容与服务等业务的费用处理，还要逐步探索面向算力网络等未来业务的方案推荐与费用处理，是助力电信企业打造可持续发展的引擎。

电信运营企业从传统 ToC 通信业务的利基市场逐步向未来算力网络的多边平台市场发展过程来看，随着业务发展，可以逐步在计费系统的应用场景中引入 GPT 大模型相关的技术，以使得计费系统更加智能。

## 场景一：计费系统的可观测性与可解释性提升

电信运营企业从其国企定位来看，有较强的社会责任与企业责任。让客户每一笔通话、每一笔订阅都能做到明明白白消费是非常重要的。而随着业务复杂程度的不断提升，用户的资费、优惠等越来越复杂，进而造成计费系统在进行批价、优惠的过程中需要花费很大的成本来为客户提供使用与消费过程的说明。引入 GPT 大模型相关技术，将用户的计费数据（使用记录）、优惠数据，以及资费信息等数据建立分析，同时配合 NLP 实现自然语言交互情景下的消费情况解释，有助提升客户消费的满意度。

例如：传统上，当客户对消费情况提出质疑的情况下，通常通过 IVR 等渠道录入客户服务工单，经专家分析、批价过程重现等手段分析消费记录后，再向客户提供解释说明。引入 GPT 大模型相关技术，可以为客户提供基于自然语言的解释手段，降低成本的同时，提升客户满意度。

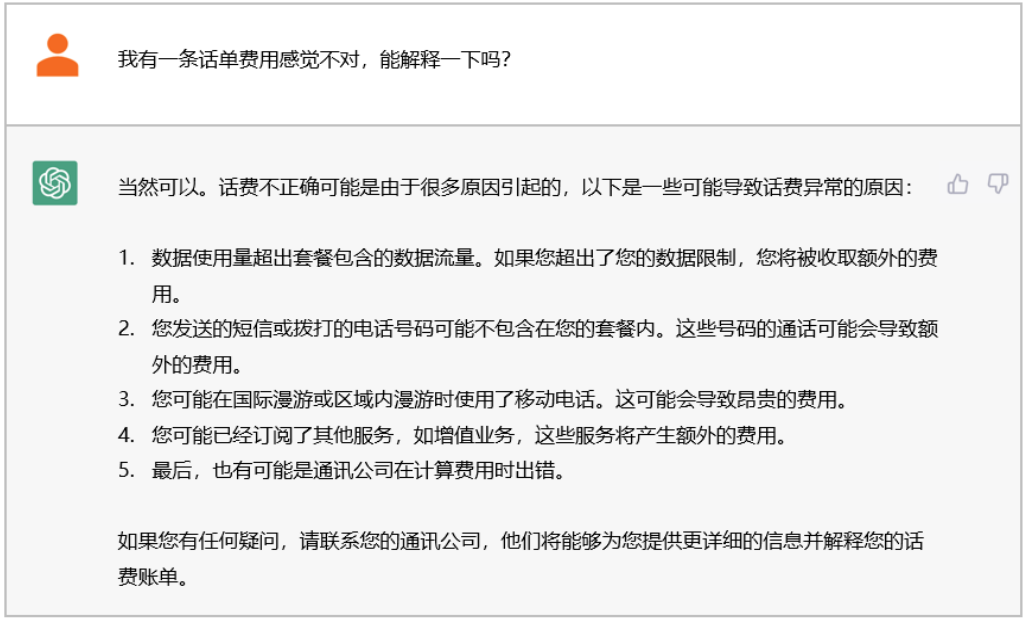

**图 3-22 GPT 大模型根据计费数据给出情况解释**

**场景二：业财合拢助力精益企业运营**

电信运营企业是典型的重资产投入企业，在整个"资金 -> 资产 -> 资源 -> 服务 -> 资金"的过程中，由于网络投资与用户收入的财务统计方法和口径的天然差异，很难做到资源的精准投放，进而影响电信运营企业的运营效率。引入 GPT 大模型相关技术，将用户的计费数据（使用记录）与网络资源投资情况（利用效率）相结合，建立多类、多模数据的融合分析，可以帮助电信运营企业实现更加精准的资源投放，进而提升运营效率。

例如：传统上，由于网络建设是基于网络规划、监测、优化等数据通过项目方式推进建设，而设备采购是按照财务预算投资节奏进行采购，用户收费又是基于用户单笔使用记录，因此在统计口径上很难形成一致。将网络数据、用户使用及费用数据导入，通过建立企业财务数据、用户使用及费用数据与网络数据之间的概率关系，实现财务与业务的有效结合，提高企业的盈利能力和市场竞争力，从而实现可持续发展。

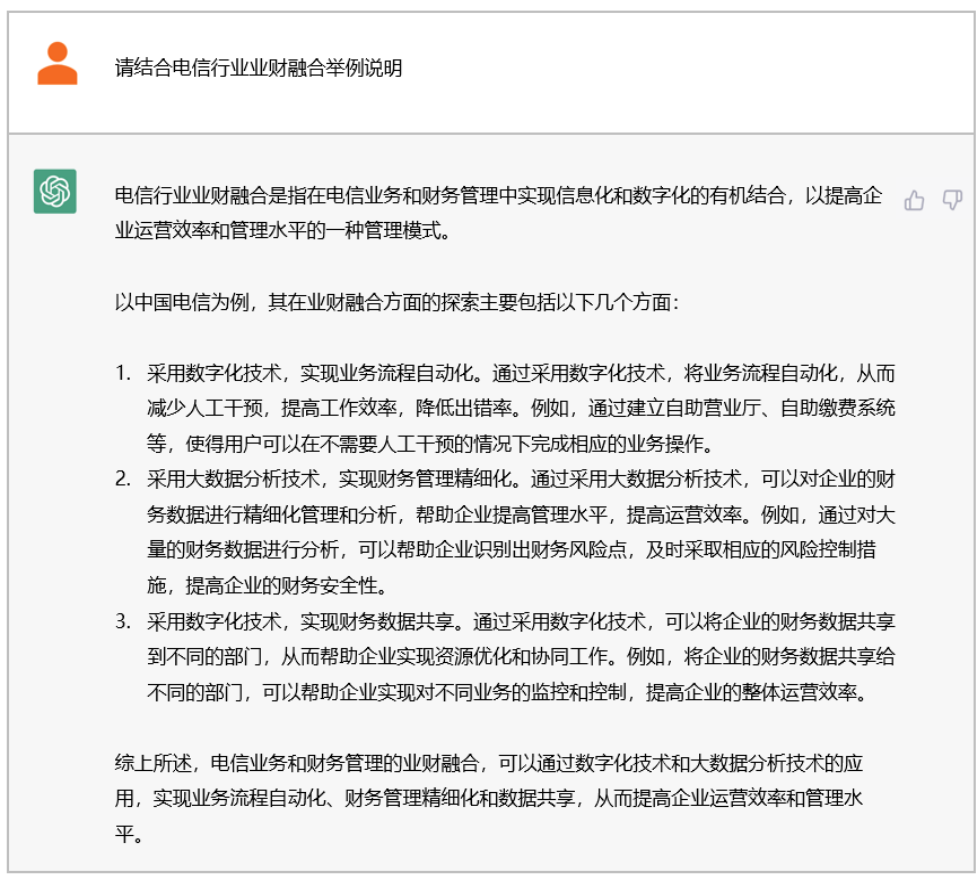

**图 3-23 GPT 大模型智能将企业业财融合助力运营**

## 场景三：多边平台模式下的风险控制

电信运营企业随着业务发展，无论在基于内容与服务业务的多边平台，还是在未来的算力网络的多边平台模式下，要想建立一个健康的平台生态，就需要建立以足够强大的风险控制体系以确保整个多边平台的信用基础。引入 GPT 大模型相关技术，针对多边平台的合作方建模超长历史交易行为，充分利用每次交易记录信息，利用思维链进行多步推理，逐步完善一种低成本的风控体系，可以有效确保多边平台下的信用基础。

例如：基于 SP/CP 的业务，对所有最终用户、SP/CP 等各合作方的基础信息、内容与服务产品信息建立标注，同时对每一笔交易记录建立跟踪说明，基于最终用户与合作伙伴在申请、鉴权、使用、分摊、结算等过程建立风险控制判断，

在 GPT 中训练风险控制评价模型即对风险控制判断进行奖励优化。正式使用中可以将每一笔交易请求通过模型评估该笔交易的风险等级，进而规避风险。

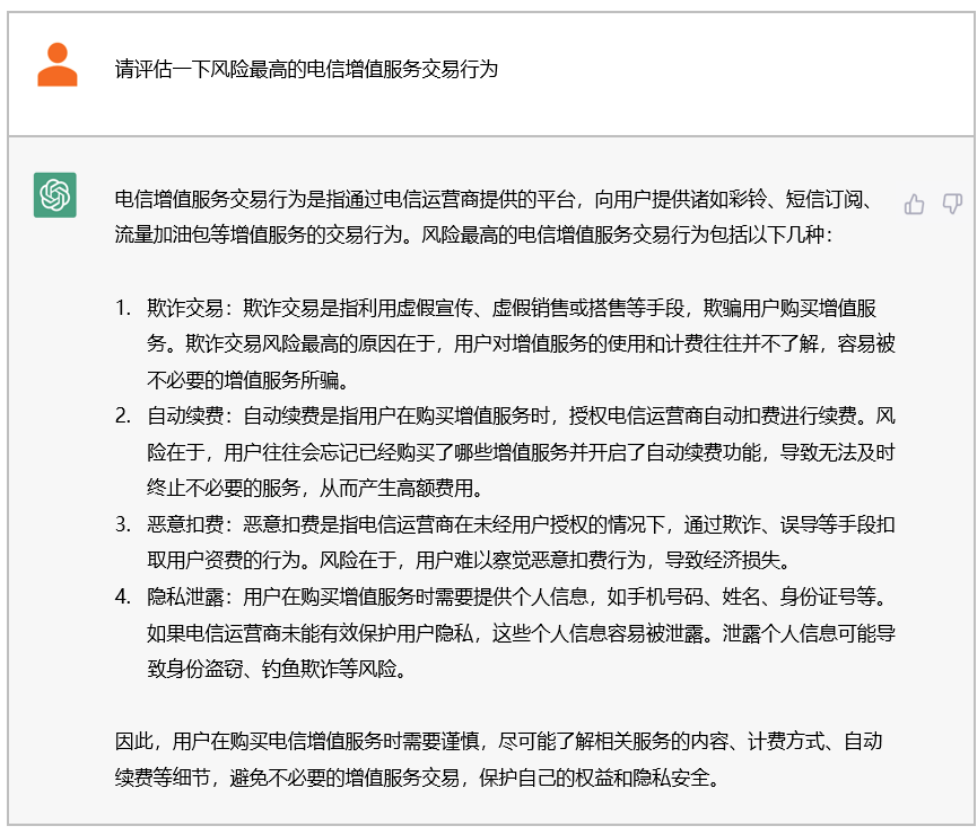

**图 3-24 GPT 大模型举例风险交易行为**

**场景四：基于意图的算网方案推荐**

随着运营商在算力网络方向的努力，传统基于简单资源成本及服务水平的定价模式很难适应算力网络中复杂场景下，多资源、多服务的融合服务模式。引入 GPT 大模型相关技术，以具体使用场景为意图基础，同时整合各类资源、各类服务能力的信息，利用思维链进行多步推理，给出符合客户场景意图、电信运营企业效益的最佳建议方案，可以帮助电信运营企业建立可持续性发展的算网生态。

例如：在算网资源管理阶段，对每一个节点的算力和网络资源、服务能力、成本、预期收益等建立标注，同时基于客户在不同场景（基于六何分析法构建场景模型）下的资源、服务请求建立跟踪说明，基于客户在不同场景的跟踪说明，

以及算力、网络资源的标注，建立同时满足运营商成本、效益、客户预期的调度方案与资费方案，在 GPT 中训练最优解评价模型即对调度方案与资费方案进行奖励优选。正式使用中可以根据算网请求识别客户意图（成本最优、效率最优、效能最优等）自动推荐算网资源调度方案并形成自动化订购、分摊、结算等过程。

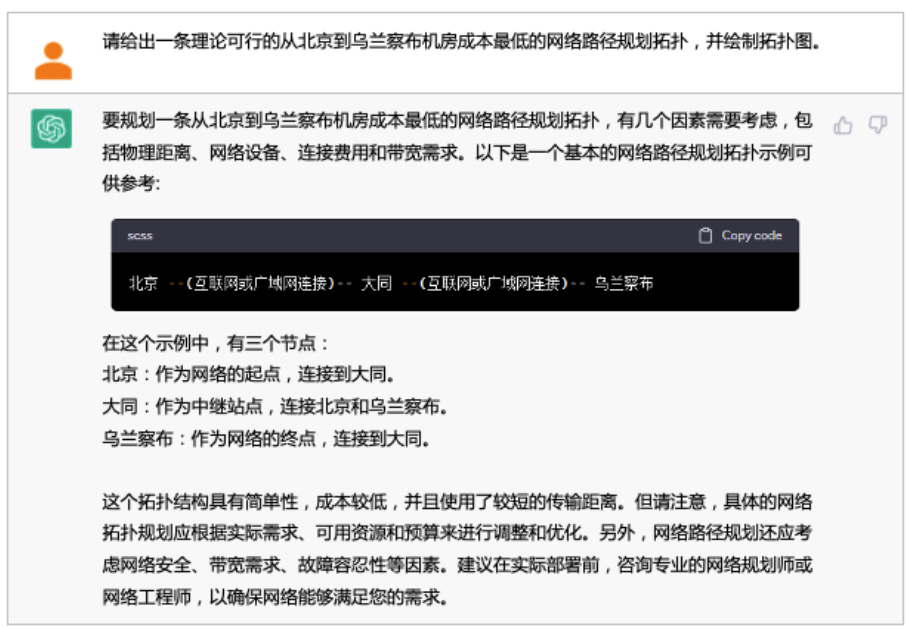

**图 3-25 GPT 大模型基于意图的算网方案推荐**

## 3.1.5 智能客服

客服系统旨在为通信运营商的客户提供一种高效、灵活、可定制的解决方案，用于管理、维护运营商与客户之间的交互和支持。包括一个集成的多渠道服务平台，支持通信运营商在多个渠道（例如电话、邮件、短信、社交媒体和在线聊天）上与客户进行互动。

智能客服系统可以通过自动化处理、智能机器人和自助服务等方式来降低人工客服中心的负载，减少客服中心的成本，提高客户满意度。此外，系统还提供实时分析和监控工具，以便电信运营商可以追踪客户反馈和服务质量，及时解决问题并改进服务质量。

将智能客服系统和 GPT 大模型结合起来，可以发挥两者的技术优势，在智能语音助手、坐席助理、智能推荐、自助服务、社交媒体管理、个性化服务等多个场景方面，提高客户服务的质量和效率，满足客户日益增长的个性化需求，从而帮助企业更好地服务客户，提高竞争力和盈利能力。

**场景一：增强智能客服的语义理解、情感识别**

智能客服系统在处理客户请求方面具有很大的优势，可以自动分析和处理大量的客户请求，提高客户服务的效率和质量。然而，智能客服系统有时可能难以理解和回答复杂的问题，或者无法识别客户的情感和语气，导致客户体验下降。因此，结合 GPT 大模型这种自然语言处理技术，可以弥补智能客服系统的不足之处。GPT 大模型可以识别用户提问的主题和关键词，并将其转化为机器可读的形式，帮助智能客服系统更好地理解用户的需求，从而提供更准确、个性化的服务，也可以帮助智能客服系统识别用户的情感状态。GPT 大模型可以分析用户输入的语言，判断其中的情感色彩，比如是否表达了愤怒、焦虑或快乐等情感状态。通过分析用户情感状态，智能客服系统可以更好地理解用户的情感需求，并提供更加智能化、贴心的服务。

例如：当一个用户在智能问答系统中输入"我的账单有误，这已经是第三次了"，GPT 大模型可以识别到用户的不满和不满意情绪。这时，智能客服系统可以自动回应，并表达理解和同情，并提供更快速和有效的解决方案，以缓解用户的不满情绪。

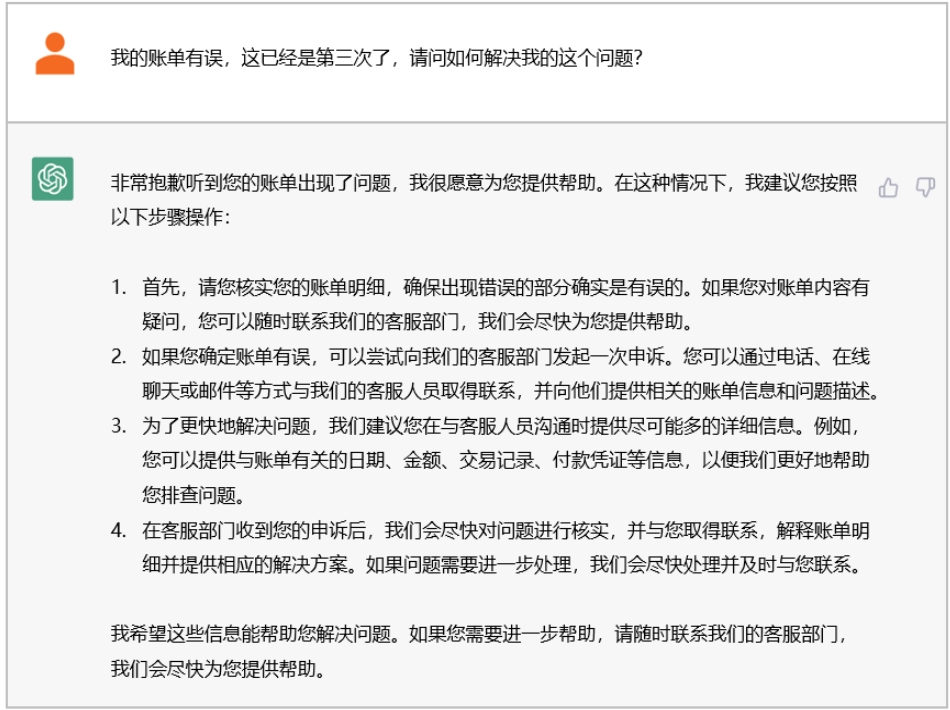

**图 3-26 GPT 大模型基于客户情绪，智能化处理客户请求**

结合 GPT 大模型的智能客服系统可以更好地理解用户的问题和需求，识别用户的情感状态，并提供更智能化、贴心的服务，从而提高客户的满意度和忠诚度。

### 场景二：使智能客服知识搜索定位更精确

有时候客户可能并不清楚如何描述他们所遇到的问题，或者他们可能使用了不正确的词汇来描述问题。这时，智能客服系统也会出现困惑，无法找到准确的解决方案。结合 GPT 大模型后，GPT 大模型可以帮助智能客服系统在这种情况下提供更精确的搜索定位，用户提供最相关和精确的回答和解决方案。

例如：客户向智能客服系统中提出了"我的电脑无法上网了，什么情况？"。由于客户的输入比较简略，智能客服系统可能会出现困惑，无法准确理解用户的问题，也无法为用户提供最准确的解决方案。

通过与 GPT 大模型的结合，智能客服系统可以更好地理解用户的问题并提供更精确的搜索定位。GPT 大模型提供了更大的语境窗口，处理和记住更多的信息，从而提高它的精度和细节处理能力，可以利用上下文和历史数据生成更准确的问题描述，针对"我的电脑无法上网了，什么情况？"问题，智能客服系统可以将这个更加准确的问题描述用于搜索相关解决方案，为用户提供更具体、个性化的建议和指导。

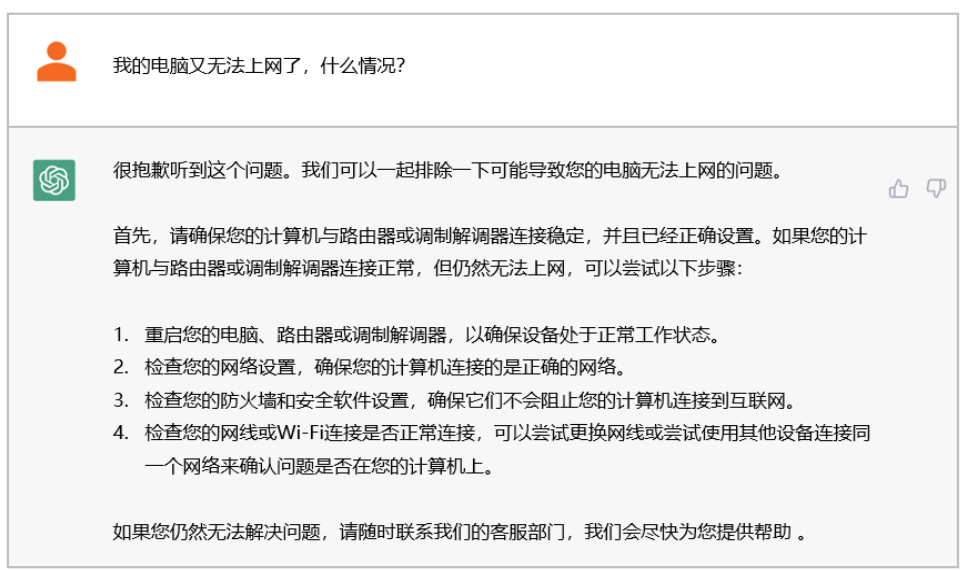

**图 3-27 GPT 大模型根据客户提问，智能化生成解决方案**

GPT 大模型还可以通过理解用户的意图，进一步优化搜索结果。如果 GPT 大模型理解到用户需要的是如何解决电脑无法启动的问题，而不是如何更换硬件设备，那么智能客服系统就可以提供更准确的解决方案，使客户更容易找到所需的帮助和支持。结合 GPT 大模型的智能客服系统可以更好地理解用户的问题和意图，提供更精确、个性化的搜索结果和解决方案，从而提高客户的满意度和忠诚度。

### 场景三：增强智能客服的交互式服务体验

智能客服系统结合 GPT 大模型，可以为客户提供更加个性化、准确和自然的回答和解决方案，从而增强智能客服系统的交互性和用户体验。在智能客服系统中，客户可以通过语音或文本与机器人进行交互。机器人通过分析客户的问题，自动检索相关信息，并生成回答或解决方案。然而，机器人有时无法理解或解决复杂的问题，或者无法识别客户的情感和语气，导致客户体验下降。在这种情况下，GPT 大模型可以理解客户的意图和语境，生成更加个性化、准确和自然的回答和解决方案。GPT 大模型还可以识别和回应客户的情感和语气，以提供更加细致的服务。另外，GPT 大模型还可以通过学习历史对话和客户信息来持续优化智能客服系统，提高回答准确率和交互性。通过这种方式，智能客服系统可以不断适应和优化自己，提供更好的服务体验，增强客户满意度，维护客户关系，提高企业的竞争力和盈利能力。智能客服系统与 GPT 大模型结合，提供更加个性化、准确和自然的回答和解决方案，以解决客户的投诉问题。

假设 GPT 大模型担当通信运营商的客户代表，一个客户通过智能客服系统投诉手机质量问题，GPT 大模型为这个客户提供服务。客户可能会表达情绪，描述手机的故障和影响，提供订单号码和其他相关信息等等。智能客服系统可以分析客户提供的信息，自动识别问题，并通过事先编写好的规则或模板，生成一份回答或解决方案。然而，如果问题比较复杂或客户的情感比较激烈，没有 GPT 大模型支持的智能客服可能难以理解和回答客户的问题，导致客户体验下降。结合 GPT 大模型的智能客服系统可以提供更加个性化、准确和自然的回答和解决方案，以解决客户的投诉问题，并提高客户满意度和忠诚度。

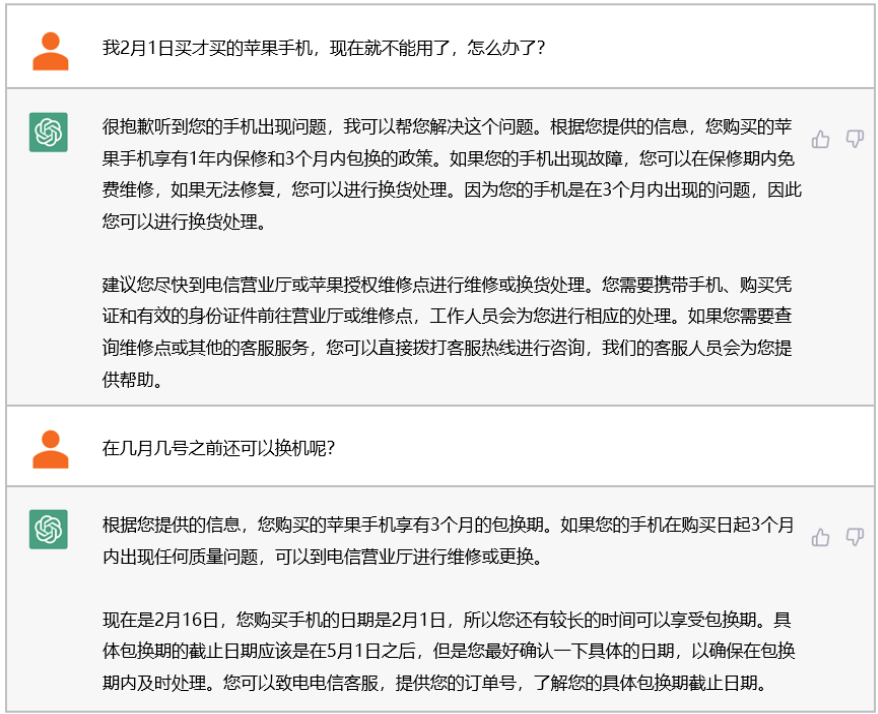

**图 3-28 GPT 大模型与客户交互式对话能力展示**

# 3.1.6 智能客户关系管理

客户关系管理系统（Customer Relationship Management，CRM），是企业为提高核心竞争力利用软件、硬件和网络技术，建立的一个客户信息收集、管理、分析和利用的信息系统。以客户数据的管理为核心，记录企业在市场营销和销售过程中和客户发生的各种交互行为，以及各类有关活动的状态，提供各类数据模型，为后期的分析和决策提供支持。

在客户关系管理系统中，引入 GPT 大模型可以更好地协调企业与顾客间在销售、营销和服务上的交互，从而提升管理方式，向客户提供创新式的个性化的客户交互和服务的过程。促进最终目标"吸引新客户、保留老客户以及将已有客户转为忠实客户，增加市场"的达成。

**场景一：人性化智能的交互方式，促进客户信息收集**

CRM 系统通过客户注册、客户主动填写个人资料、问卷调查等方式收集客户信息，虽然可以通过奖励积分等方式激励客户填写个人资料或进行问卷调查，但传统冗长的表单填写模式会让很多人望而却步。GPT 大模型采用更为人性化的提问方式进行问卷调查，同时可以根据客户的给出的答案，更为智能地选择向客户询问的问题，在问卷调查的过程中，若发现客户出现不满情绪，能及时进行安抚，最大程度保证问卷调查的顺利完成。

例如：为了解 5G 网络用户普及和使用情况，收集对 5G 发展的建议，开展了 5G 服务的问卷调查活动，根据客户是否使用了 5G 网络，GPT 大模型会聪明地选择不同的问题进行调研。

**图 3-29 GPT 大模型人性化提问方式进行问卷调查**

**场景二：知识图谱与 GPT 大模型结合，更深入的分析，提供更准确的推理**

GPT 大模型是一个强大的语言模型，但它在上下文、背景知识和推理方面存在局限性。知识图谱本质上是语义网络，把所有不同种类的信息连接在一起而得到的一个关系网络，因此知识图谱提供了从"关系"的角度去分析问题的能力。两相结合，一方面可以弥补 GPT 大模型推理方面的局限性，另一方面可以用更人性化、友好且智能的方式将知识图谱的分析结果呈现给用户。

例如：知识图谱根据收集到的某客户购买的电信套餐、消费记录、社交圈（通过分析话单获取）、调查问卷结果等信息，通过数据挖掘、信息处理、知识计量等处理，推理出此客户现在虽然使用的是 4G 网络，但对 5G 网络有一定兴趣，并且与之通话频繁的三个用户最近均开始使用 5G 网络，GPT 大模型提醒客户经理，可联系此客户，推荐使用 5G 业务。

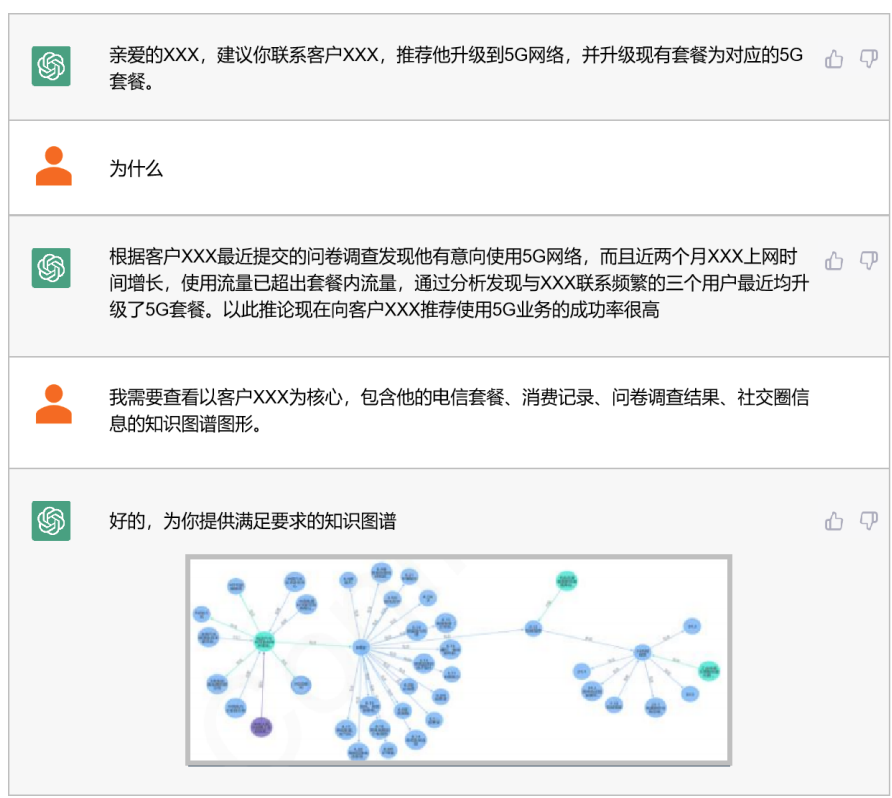

**图 3-30 GPT 大模型结合知识图谱信息进行推理，为客户经理挖掘潜在客户**

### 场景三：GIS 与 GPT 大模型的结合，提供智能化地址管理与服务

客户地址信息是客户关系管理的基础性关键信息，可以利用 GPT 大模型辅助对地址进行采集、治理和管理，结合了 GIS 技术和 GPT 大模型的优势，可在多个应用场景下实现智能化的地址管理与服务，提高数据准确性、业务效率和用户满意度。

例如：多源地址合并：在采购的地址库与运营商自有地址库合并任务中，可以利用 GPT 大模型 的强大文本逻辑推理能力及地理知识库，辅助地址清洗人员实现地址库的准确合并，而不必出门踏勘。这一应用场景充分发挥了 GPT 大模型 的智能文本处理能力，减轻了地址清洗人员的工作负担，提高了地址库合并的准确性。

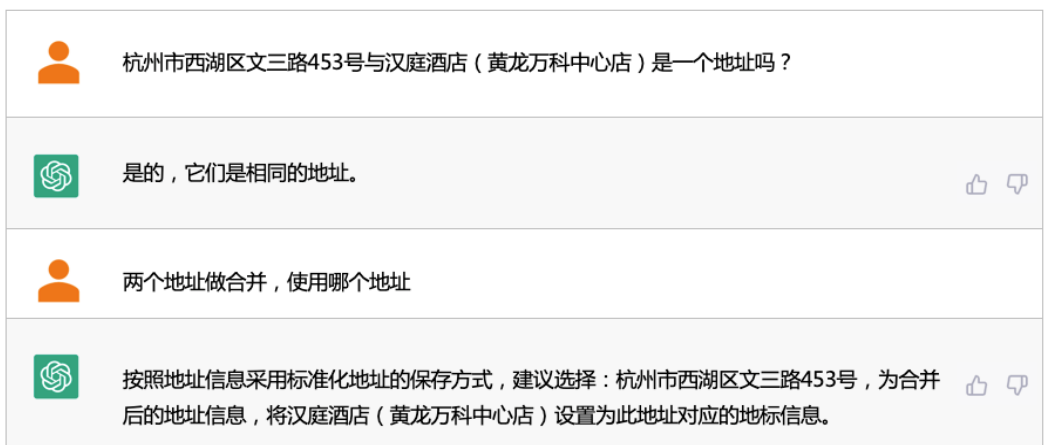

**图 3-31 GPT 大模型智能文本处理实现多源地址合并**

例如：用户地址匹配：在用户报装过程中，当用户口头报出自家地址时，可以利用 GPT 大模型 的能力，将其识别匹配为标准地址，从而快速实现报装地址的精确选择，提高业务办理速度和用户满意度。这一应用场景充分发挥了 GPT 大模型 的智能文本处理和地理知识库能力，实现了对用户需求的快速响应和解决。

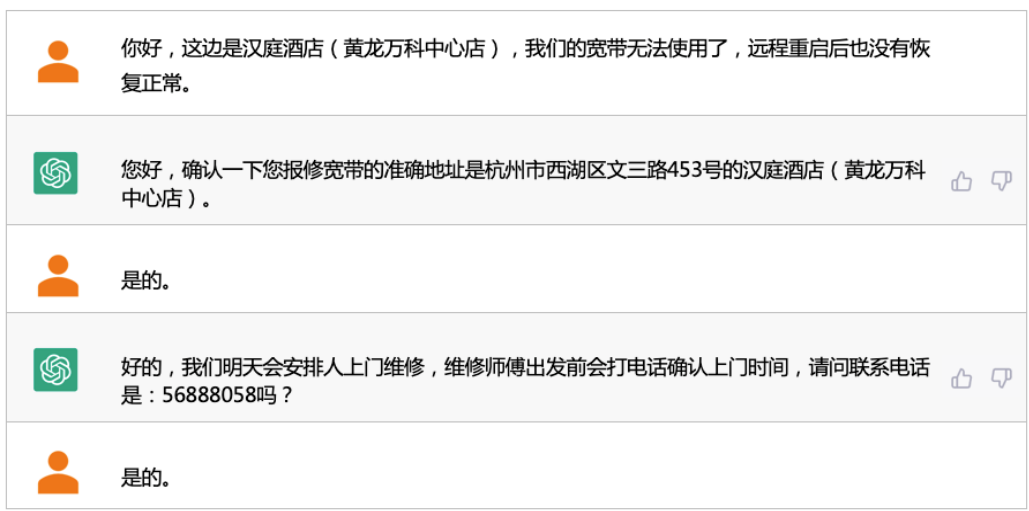

**图 3-32 GPT 大模型智能文本处理实现用户地址匹配**

**场景四：GPT 大模型虚拟智能同伴，促进客户关系的建立与维系**

利用 GPT 大模型创建一个虚拟的智能同伴，能够帮助客户经理以更加智能、自然和快速的交互方式分析潜在客户群和预测市场发展需求，为寻找新客户提供商业策略。

GPT 大模型虚拟智能同伴基于收集、整理和分析的客户信息，在客户经理与目标客户交流的过程中，通过客户对话内容和说话语气，识别客户情绪，为客户经理提供可使用的对话策略，可推介的产品信息，可激发客户兴趣的话题以及可激发购买冲动的建议，并提醒调整交谈的语气和方式，从而更好了解客户的需求，进行响应，达到拓展新客户、维系老客户的目的。

例如：一个新入职的电信政企客户经理，使用 GPT 大模型了解工作职责，获取近期需要拜访的客户信息。

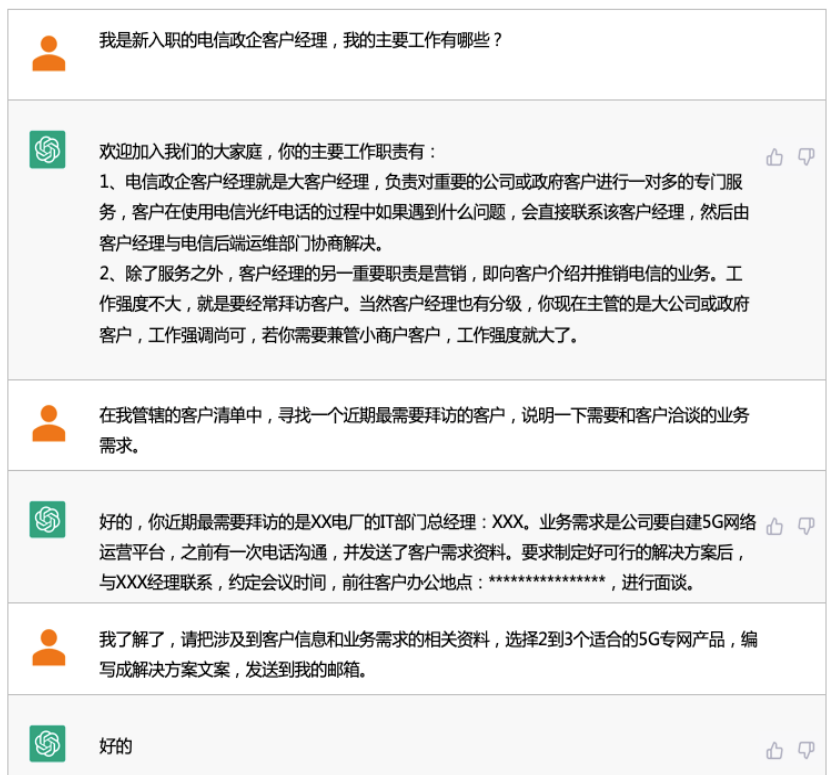

**图 3-33 GPT 大模型虚拟智能同伴应用案例 1**

例如：为维系客户关系，客户经理通过电话交流的方式，主动为客户提供积分兑换服务，GPT 大模型虚拟智能同伴从旁协助，根据双方对话内容，及时提供应对策略。

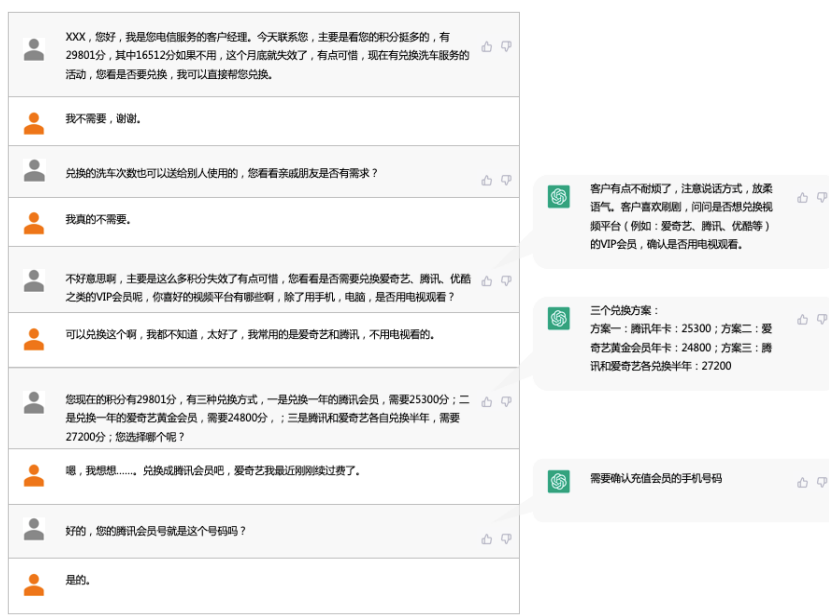

**图 3-34 GPT 大模型虚拟智能同伴应用案例 2**

### 3.1.7 商业智能

商业智能（Business Intelligence，简称 BI），又称商业智慧或商务智能，指用现代数据仓库技术、线上分析处理技术、数据挖掘和数据展现技术进行数据分析以实现商业价值。

1996 年，BI 的概念被 Gartner 提出，并前后经历了三代发展。BI1.0 即传统 BI 诞生于上世纪 90 年代。作为面向 IT 侧的报表平台，传统 BI 技术门槛较高，以 IT 为主导，业务需求匹配度较低。BI2.0 即敏捷 BI，以自助式分析为卖点，在 2010 年以后进入快速发展阶段。敏捷 BI 面向业务人员提供自助式分析服务，但仍然存在一定技术门槛，需要耗费较大的时间和人力成本，不利于知识的沉淀和经验分享。2020 年以后，国外头部厂商掀起 BI3.0 浪潮，即智能 BI，结合人工智能技术，打造一种纯业务侧主导，基于自然语言处理能力的数据搜索与分析平台，开启报表零门槛、人人都是分析师的时代。

国内商业智能市场经历十来年的发展，目前正处于敏捷 BI 的阶段。敏捷 BI 提供包括图表模板、分析函数、可视化组件在内的部分工具供用户选择，虽较传统 BI 在自助化方面有所提升，但完全依靠人工经验进行选择和判断，费时费力。运营商在产品服务和渠道运营方面呈现出多元化的发展趋势，敏捷 BI 已无法满足数据分析诉求在数量、频率、交付周期、场景化的广度和深度以及准确性等方面的要求，亟需向智能 BI 演进。

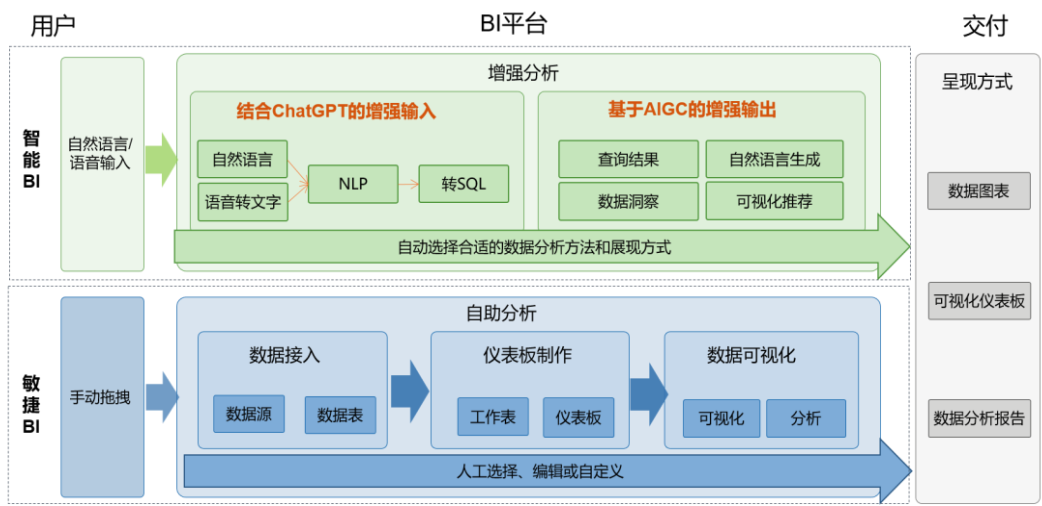

**图 3-35 智能 BI 增强分析 VS 敏捷 BI 自助分析**

增强分析是智能 BI 的核心能力，支持更多不同的业务角色全程参与数据分析，并结合实际业务诉求把关分析质量与效率。BI 系统通过集成 GPT 大模型能力，将从以下几个环节实现增强分析：

1. **需求输入方式的变更**：基于 GPT 大模型的自然语言理解和多轮对话能力，支持用户通过文字描述或语音输入的方式替代传统的手动拖拽。在大幅提升 Text2SQL 的准确性和及时性的基础上，自动将用户输入的自然语言转化为查询 SQL。

2. **分析方式的变更**：结合业务需求，基于大模型对运营商私有数据进行训练，针对指标异常检测、多维度和多指标关联分析、根因分析、趋势和结果预测等细分领域生成场景化模型，开展自动数据洞察。

3. **可视化呈现方式的变更**：将可视化比作是一种数据到设计的翻译，好的可视化也需要符合 "信达雅" 的标准。BI 系统基于 AIGC 的能力，结合用户意图和分析结果，自动生成并推荐可视化方案。一方面可以根据用户语言描述自动化生成报表，尤其是复杂的财务报表。另一方面 GPT 大

模型结合图形语法学，支持用户自定义任意图形，将自然语言所描述的绘制理念通过语法规则和复杂的语法结构生成代码表达；随着 AIGC 能力进一步发展，支持根据用户意图自动生成自定义的可视化结果。

4. **报告交付方式的变更**：基于 GPT 大模型的自然语言生成能力，对分析过程中获得的现象、原因、预测等要素进行文字描述，并结合可视化的图表进行合理布局，自动生成可交付使用的书面报告。

**场景一：GPT 赋能增强分析，实现智能化经营分析**

虽然很多企业部署了自助式的数据分析工具（敏捷 BI），但在实际经营分析场景中，业务侧用户的使用率并不高。大部分业务经营者和管理决策者仍然采用传统的经营分析模式，依赖于专业的分析团队，从提出需求到完成交付，流程冗长、参与协作难度大、效率和质量和预期存在较大差距。究其原因发现：找不到指标、看不懂数据以及分享不便捷成为阻碍用户使用分析工具的痛点。

结合大模型的能力优势，提出以下优化举措：

针对找不到指标：通过自然语言检索数据指标，除了明确的指标外，也可以通过泛化指标分析，自动关联识别指标名称，用户即使不知道指标名称，也能根据业务化的语言描述，找到想要分析的指标。

针对看不懂数据：通过意图识别和多指标关联分析，发现各指标之间的联系，将关联性较强的指标进行对比分析，自动推荐合适的图形表达方案。

针对分析不便捷：一键生成报告长图，对于长篇幅的分析报告，可以用一张长图完成，无需手动截图操作。

通过通信行业大模型 TelcoGPT 赋能增强分析，在指标检索场景下支持精确检索和泛化检索两类需求。

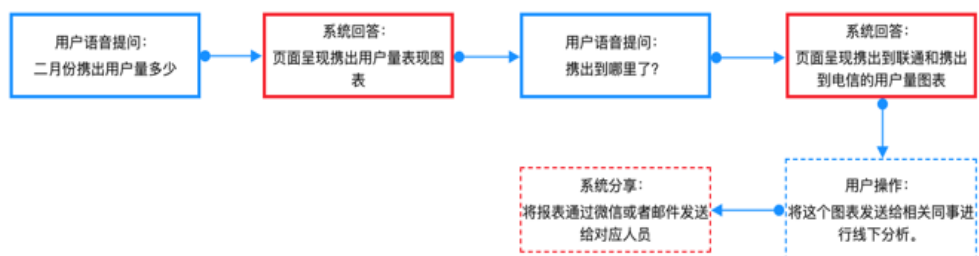

**图 3-36精确指标检索与分析**

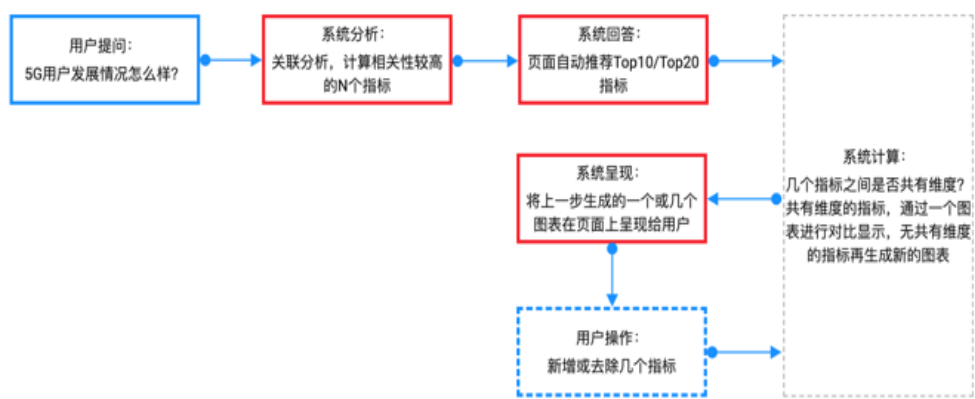

**图 3-37泛化指标检索与分析**

## 3.2 通信网络领域的典型应用场景

随着 5G 网络建设和业务发展逐步加速，在网络运营与运维领域，运营商正在面对巨大挑战。一方面，5G 在网络基础设施中引入 NFV、SDN、云原生架构等新技术，运营商的网络运营与管理工作的技术对象已经与传统的 2/3/4G 时代大不相同。另一方面，纷繁的业务场景组合要求 5G 能够提供按需分配，实时响应、端到端保障的网络体验，运营商的网络管理目标也已从"确保网络运行稳定"向"高效支撑业务发展"转变。这些变化为网络运营管理工作带来了前所未有的复杂度。电信网络运营商需要面向消费者和垂直行业客户提供全自动、零等待、零接触、零故障的创新网络服务与 ICT 业务，打造自服务、自修复、自优化的自智

网络。但是由于通信网络数据海量多维异构、数据质量不高；知识领域较多、规则多，专业化程度复杂，缺乏知识提炼或表达；网络复杂、动态可靠性要求极高导致目前基于图像视觉及自然语言处理的 AI 技术距离 L5 级的网络智能仍有较大差距，因此以 GPT 大模型为代表的 LLM 技术，在融合多模态后可能成为网络智能化实现革命性突破的重要机会，能够大幅提升网络自治的水平，助力 L5 级自智网络达成，同时也将为 6G 网络技术研发注入新的范式。下文围绕网络规建维优营全生命周期各领域梳理了 GPT 大模型的一些应用场景。

## 3.2.1 智能网络规划

网络规划包括规划目标建立、规划方案设计、规划仿真的全流程管理工作。涵盖从网络整体表现、产品运营战略、业务使用体验提升等角度建立规划目标。通过连通规划目标和规划方案的能力（包括拉通环境数据、业务需求数据、资源数据的多维度分析能力），实现业务覆盖、容量、带宽等规划目标。通过仿真能力，实现规划目标的仿真验证。

传统网络规划较为聚焦于网络覆盖、网络容量、拓扑结构、路径路由，缺乏对于网络价值、网络业务发展预测等多维度的综合分析，同时需要规划人员掌握较多的专业工具，如覆盖仿真工具、容量建模工具等。GPT 大模型知识引擎可以提供一站式综合性网络规划。规划人员仅需通过输入简单的规划意图并通过多轮交互的方式，较为高效的实现不同要求的网络规划，同时 GPT 大模型也可给出专业规划工具的代码、脚本等，大幅提升工作效率。

### 场景一：无线网络规划

无线网络规划是无线网络全生命周期管理中极为重要的一环，需要通过无线环境数据（如覆盖地形地貌，建筑物特征等）、业务需求数据、频谱数据等多维

数据来规划目标区域的覆盖、容量等。现有常规无线网络容量规划方法数据来源单一，且不能根据用户需求意图及现有网络情况、用户价值及网络口碑等多种因素综合进行自动化网络容量规划。通过 GPT 大模型可以将网络规划意图较为便捷转换为对区域覆盖、容量、价值收益的综合指标，从而得出规划方案。

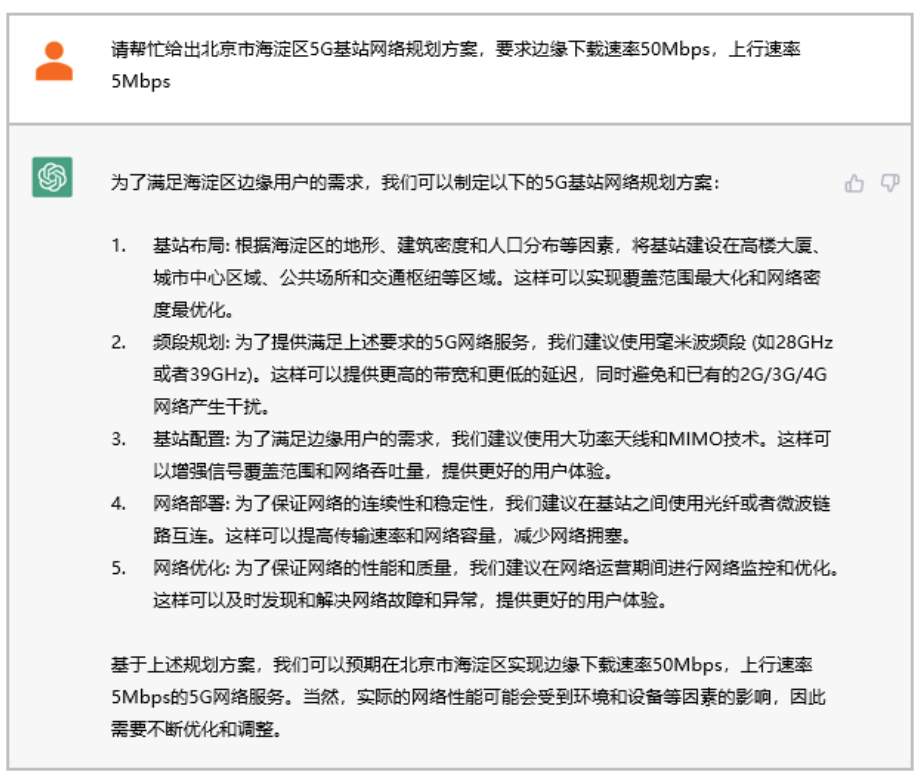

**图 3-38 GPT 大模型根据专业需求，生成无限网络规划方案**

### 场景二：切片自动勘察

5G 网络切片是为了达成对客户承诺的个性化 SLA 需求，对无线、传输和核心网等多个子域的网络资源进行隔离、预留、独占和配置的端到端网络。目前运营商评估网络能否满足用户对切片的要求，是通过开通前的资源勘查来完成的。现有常规的切片资源勘查，是通过人工方式将用户对切片的 SLA 需求转换为初步的切片方案后，通过系统派单给相应的专业的负责人（可能是大区核心网的运维人员，也可能是地市公司的无线和传输负责人），人工或半人工的确认资源满足情况。无线专业负责人需要确认业务区域相关基站和小区是否具备切片要求的

能力，如信号覆盖情况、基站能力（是否支持切片、是否支持 RB 预留能力等等）、上下行容量等等，核心网专业则需要检查 AMF 的用户容量、SMF 的会话数、UPF 的吞吐量、UDM 的开卡容量等是否满足需要。人工勘查的工作非常繁琐，而且容易出错，常导致勘查环节占用时间从几天到十几天不等，严重拖慢业务开通过程。通过 GPT 大模型能够通过多轮问答的方式对无线、传输和核心网等子域网络设备的配置数据、容量数据、性能统计数据的采集和分析，以及无线栅格数据、MR 测量数据的实时深度分析，配合不同切片类型的评估模型，实现自动化的资源勘查。

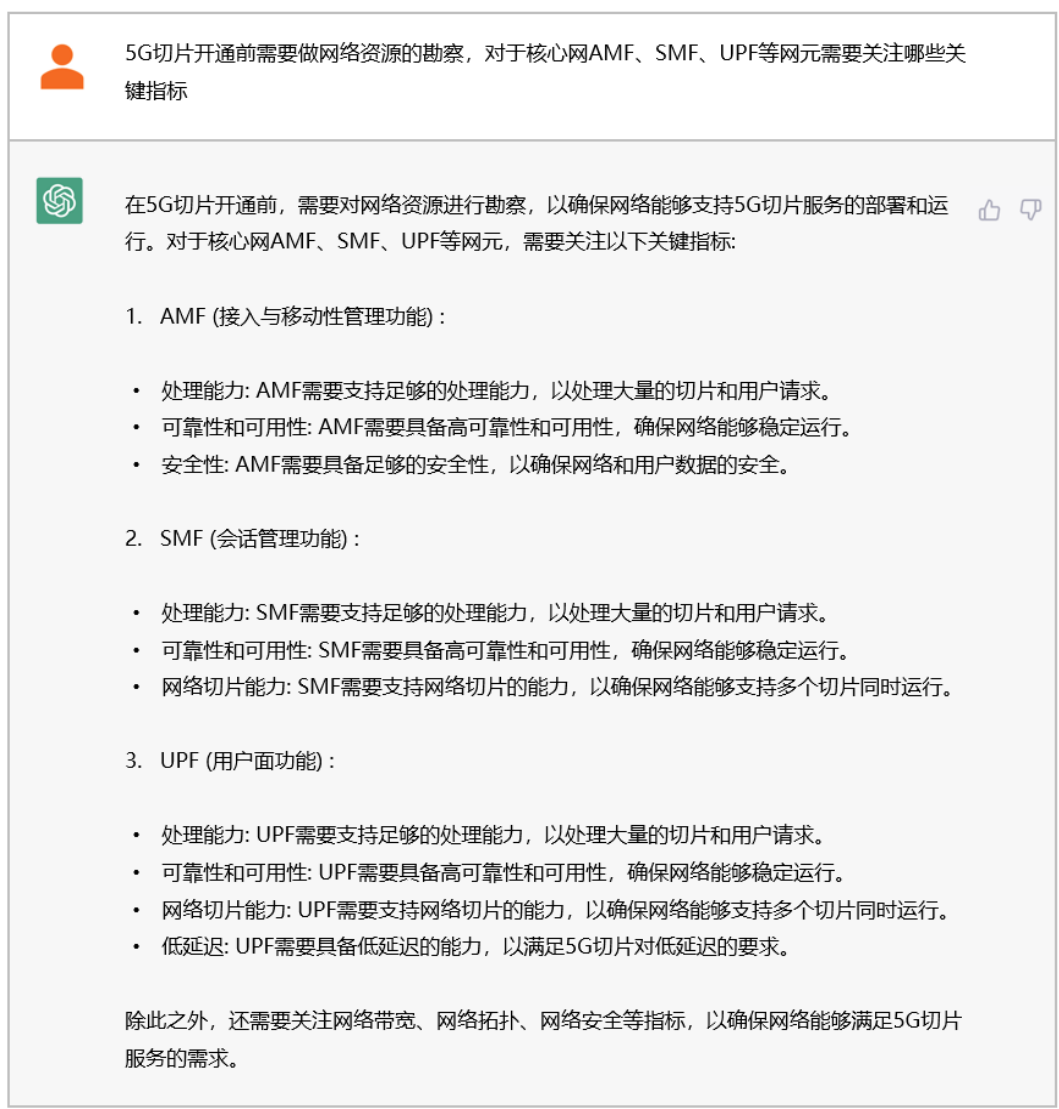

**图 3-39 GPT 大模型实现切片自动勘察**

## 3.2.2 智能网络部署

网络部署（建设）包括建设项目的立项、设计、验收的全流程管理工作。根据规划设计完成交付、配置、以建设目标为基准完成网络能力、可用性等具体指标的验收。除以上整体流程外，相关环节也可以通过自动化工具实现能力提升，包括自动化交付配置、自动化验收测试和闭环调整。

在网络部署建设的各个环节借助 GPT 大模型技术均可以实现工作效率的大幅提升，例如可通过专业领域知识模型以及网络设备专有知识实现包括 5G 专网、云网业务、边缘计算、SPN 专线、SD-WAN 等各类网络开通时间的大幅缩减，这对运营商提升市场竞争力至关重要。同时在结合多模态技术后，可以实现网络建设后质检验收的进一步自动化，节省大量人力资源。

### 场景一：基于意图的传输专线网元配置激活

随着 AI/ML 基于与网络运营运维结合的愈发紧密，基于意图驱动的网络成为网络自智发展的重要关键技术。在传统的网络管理中，运维人员需要手动配置和管理网络设备，例如路由器、交换机等。而在意图驱动网络中，运维人员只需要指定网络的目标和目的，而不需要关注底层设备的具体配置和操作。在意图驱动网络中，只需要指定网络的目标和目的，而不需要关注底层设备的具体配置和操作。其核心理念是将网络管理与业务需求相结合。通过 GPT 大模型知识引擎网络运维人员能够通过简单对话轻松实现设备开通或一键生成网元配置模板，通过与网管系统的集成实现自动配置激活。此外通过多轮对话可以高效查询网络资源位置、状态等信息降低操作门槛。

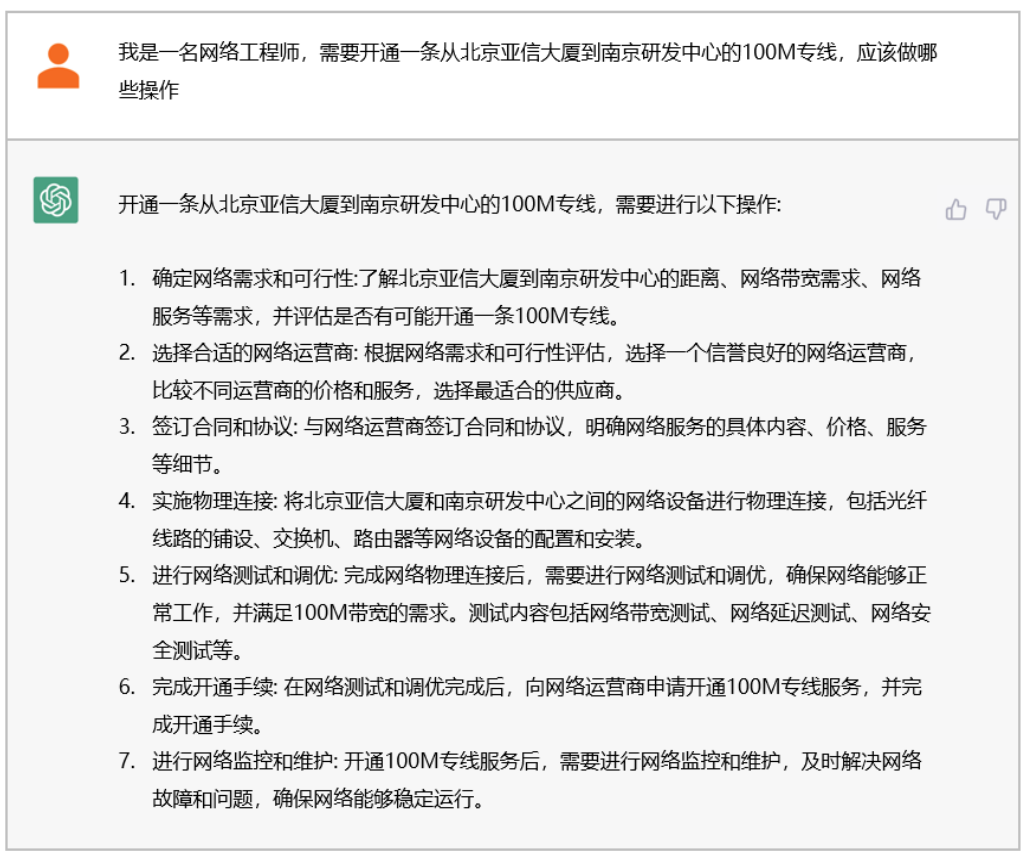

**图 3-40 GPT 大模型基于意图的传输专线网元配置激活**

**场景二：基站验收质检**

同时由于 5G 基站站型种类多，施工验收较 4G 基站更为复杂。传统的人工验收方式，验收的周期长，同时由于验收标准不一致，很难做到统一的质量管控。随着网络的大规模建设，基站验收的工作量将急剧增长，需要智能化的站点验收方式提高工作效率和质量。在融合多模态数据后，可以方便的将包括前端图片、各类测试数据、各类性能告警数据统一输入至智能引擎中，通过智能分析工程验收标准、各类质检标准比对核查，按照基站入网验收的要求，对数据进行自动解析并生成基站入网验收报告。

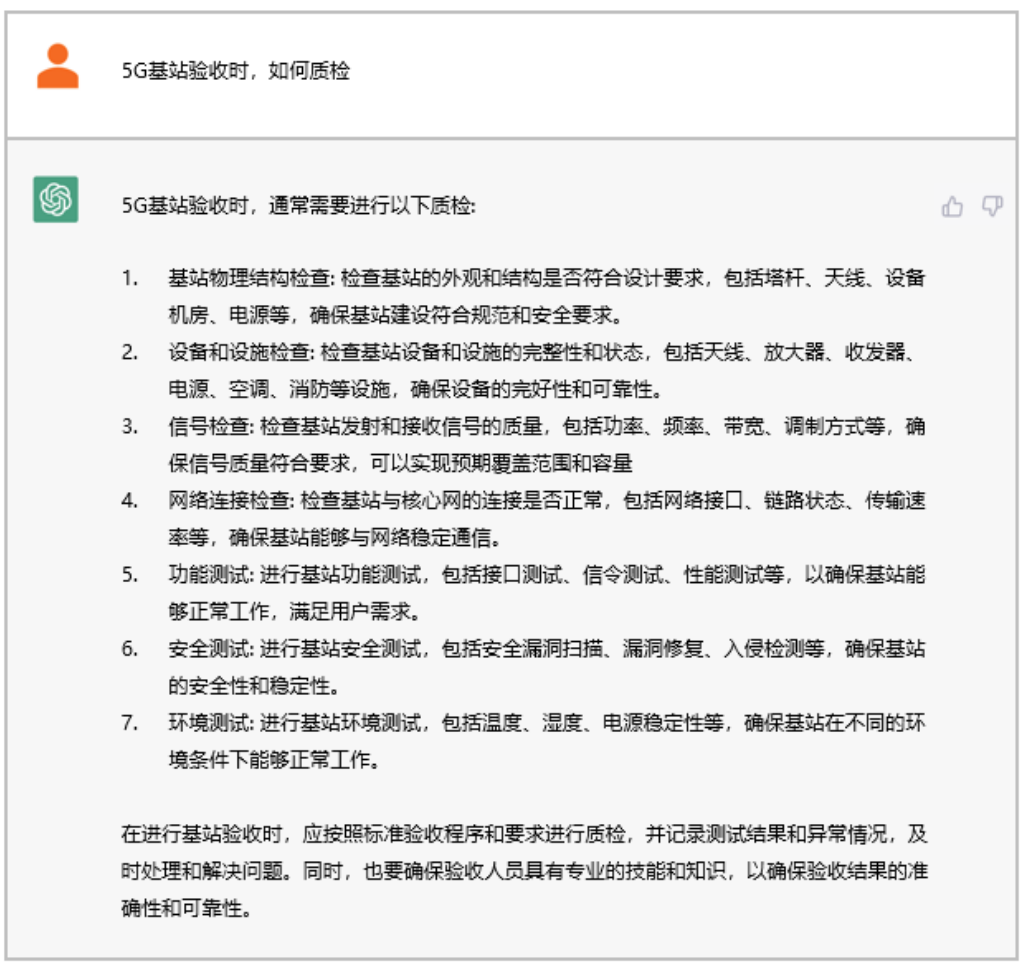

**图 3-41 GPT 大模型生成 5G 基站质检方案**

## 3.2.3 智能网络维护

网络维护包括对网络整体表现、产品运营表现、业务使用体验、资源健康度进行管理、监控、分析。通过被动的监控和处理，或者通过对故障告警和性能劣化的主动感知分析以及自动化的资源调整实现网络、业务的恢复。通过售前、售中、售后的端到端支撑能力，提供贯穿于运维各项生产环节的自动化运维感知和决策信息的流转能力。

在网络故障诊断中可使用 GPT 大模型生成故障诊断报告，提供对网络故障的快速分析。在告警、性能分析中可使用 GPT 大模型生成告警、性能、日志等关键网络数据分析，帮助网络运维人员快速识别告警的重要性并给出处理建议；

可快速自动生成各类报告及数据可视（代码），帮助运维人员更好的理解网络数据。

**场景一：家宽网络维护**

装维人员在客户现场说明遇到的问题，如"无空闲端口"等，产生工单，利用 GPT 形成的专家知识，给出解决该问题的推荐解决步骤，并结合运营商内部系统，自动分析出各个步骤所需的相关信息，如对应的 ONU、分光器、PON 口、OLT、是否欠费、账号状态、账号流量等等各种信息，推荐已离网但尚未释放的端口信息，并且自动调用资源接口修改装机单对应端口。利用 GPT 大模型能增强一线装维弱光分析、设备激活、资源端口自动更改等智能处理问题，可以大大降低综调人员学习成本，减少人工查询、人工定界定位的时间消耗，提升运维效率。

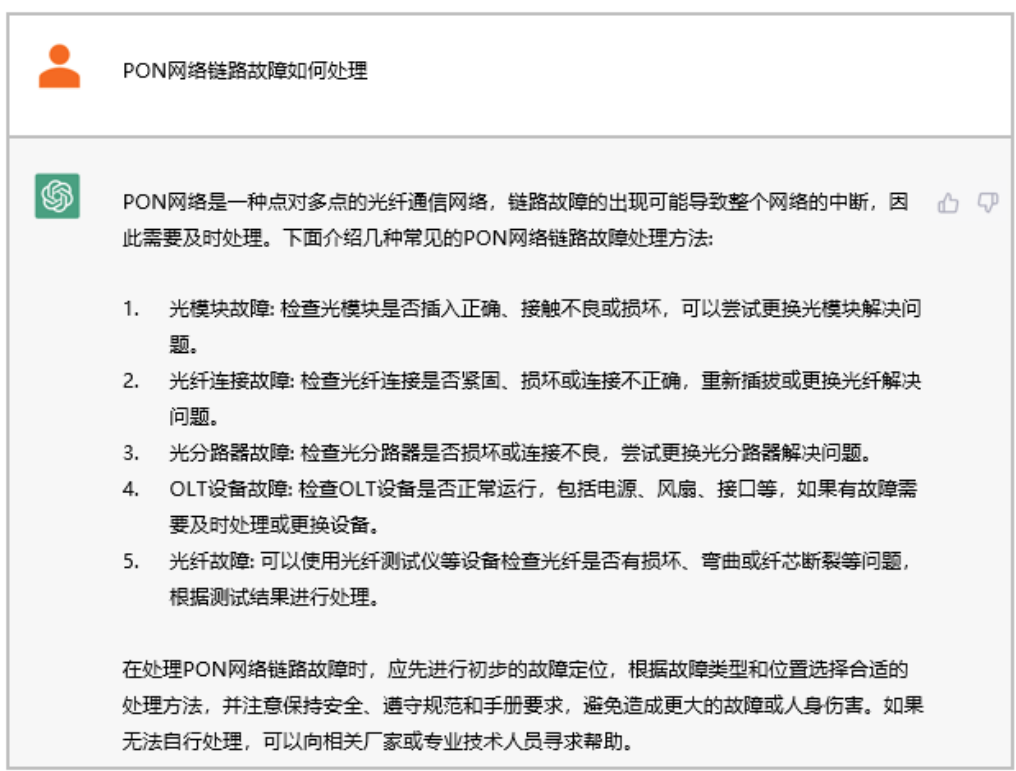

**图 3-42 GPT 大模型针对 PON 链路故障提供的解决方案**

## 场景二：基于 GPT 大模型的日志异常检测增强

借鉴 GPT 大模型的模型训练方法，可以对目前网络设备日志异常检测算法进行增强，主要步骤包括：

1. 结合专家经验选取部分日志数据，做出总结，构建提示符(Prompt)并给出答案。采用上述专家数据，在预训练语言模型的基础上，有监督地微调模型(SFT)；

2. 将日志数据输入 SFT 中，根据提示符，生成答复，基于专家经验对 SFT 生成答复进行排序；使用有序答复，在 SFT 模型上训练奖励模型(RM)；

3. 采用奖励模型，构建策略优化的强化学习模型，训练，并最终生成日志总结。

请分析如下日志，并判断是否异常: Failed to compute_task_migrate_server: No valid host was found. There are not enough hosts available.

Failed to bind port 575d6b82-f656-4d01-9360-f977bde0c229 on host h432-l13 for vnic_type normal using segments [{'segmentation_id': 20, 'physical_network': None, 'id': u'f5746dec-3bab-4396-8b97-da3652fb8397', 'network_type': u'vxlan'}]

AMQP server controller:5672 closed the connection. Check login credentials: Socket closed

AMQP server on controller:5672 is unreachable: [Errno 104] Connection reset by peer. Trying again in 1 seconds.

这些日志的内容是来自于一个计算任务迁移 (compute_task_migrate_server) 的操作，以及在此过程中涉及到的端口绑定和与AMQP消息服务器的通信。从这些日志中可以看出，操作遇到了一些问题，但不能确定是否异常。需要对每个日志进行分析:

1. Failed to compute_task_migrate_server: No valid host was found. There are not enough hosts available. (计算任务过移失败: 没有找到有效的主机。没有足够的可用主机。)

这个错误信息意味着系统尝试将一个计算任务从一个主机迁移到另一个主机时失败了，原因可能是因为系统找不到一个有效的主机进行迁移，或者所有可用的主机都没有足够的资源来处理任务。这个错误不一定是异常的，可能是由于系统负载过高或其他原因导致的，需要进一步分析。

2. Failed to bind port 575d6b82-f656-4d01-9360-f977bde0c229 on host h432-l13 for vnic_type normal using segments [{'segmentation_id': 20, 'physical_network': None, 'id': u'f5746dec-3bab-4396-8b97-da3652fb8397', 'network_type': u'vxlan'}] (无法在主机h432-l13上绑定端口575d6b82-f656-4d01-9360-f977bde0c229，使用的vnic类型是normal，使用的segment是[{'segmentation_id': 20, 'physical_network': None, 'id': u'f5746dec-3bab-4396-8b97-da3652fb8397', 'network_type': u'vxlan'}]

这个错误信息表示在给定的主机上无法绑定一个端口，其中，575d6b82-f656-4d01-9360-f977bde0c229是端口的ID，vnic_type是端口的类型，normal表示普通的虚拟网络接口。segment则是该端口所使用的虚拟网络分段的详细信息。这个错误可能是由于网络配置不正确或者主机资源不足等原因导致的。

3. AMQP server controller:5672 closed the connection. Check login credentials: Socket closed (AMQP服务器controller:5672关闭了连接。检查登录凭证: 套接字已关闭)

这个错误信息表示与AMQP服务器的连接已经关闭了，建议检查AMQP服务器的登录凭证以确定连接是否正确。可能的原因是由于AMQP服务器本身的问题，也可能是因为连接被意外断开。

4. AMQP server on controller:5672 is unreachable: [Errno 104] Connection reset by peer. Trying again in 1 seconds. (无法访问AMQP服务器controller:5672: [Errno 104]连接由对等方重置，将在1秒钟后重试。)

这个错误信息表示系统无法访问AMQP服务器。其中，controller:5672是AMQP服务器

**图 3-43 GPT 大模型日志异常检测**

### 3.2.4 智能网络优化

网络优化指规划部门、市场部门、服务部门、运维部门的需求建立优化目标、输出优化方案、执行优化流程。支持基于网络整体表现、业务使用体验、资源健康度等不同维度设定优化目标。主要手段包括通过优化方案设计能力输出常态化或专题类优化方案、通过优化分析工具执行优化方案等。

基于网络设备硬件告警、性能指标、用户投诉等多维数据，通过 GPT 大模型知识引擎给出隐患预警及并识别网络存在的问题；对于无线网络覆盖、容量、干扰、能耗、IP 路由及业务路径调整、网络云计算存储优化等常见问题根据网络的各类数据通过知识引擎给出较为精确智能的优化方案。

#### 场景一：网络能耗优化

网络优化的一项重要工作是通过用户业务的潮汐效应及网络覆盖的结构分析，完成网络结构的动态控制，进行单基站不同层级的资源休眠（如芯片级、板卡级、站点级）以及多模多制式网络的协同节能，达到网络降本增效的目标。但是在动态跟踪用户分布，跨设备厂家、多制式网络协同等方面存在较大挑战，通过 GPT 大模型智能引擎可以通过问答方式，或数据交互方式快速得到网络节能优化策略，以及网络能耗 VS 网络性能 VS 用户感知的分布地图，帮助运营商有效决策改如何动态调整网络资源达到能效&性能最优。

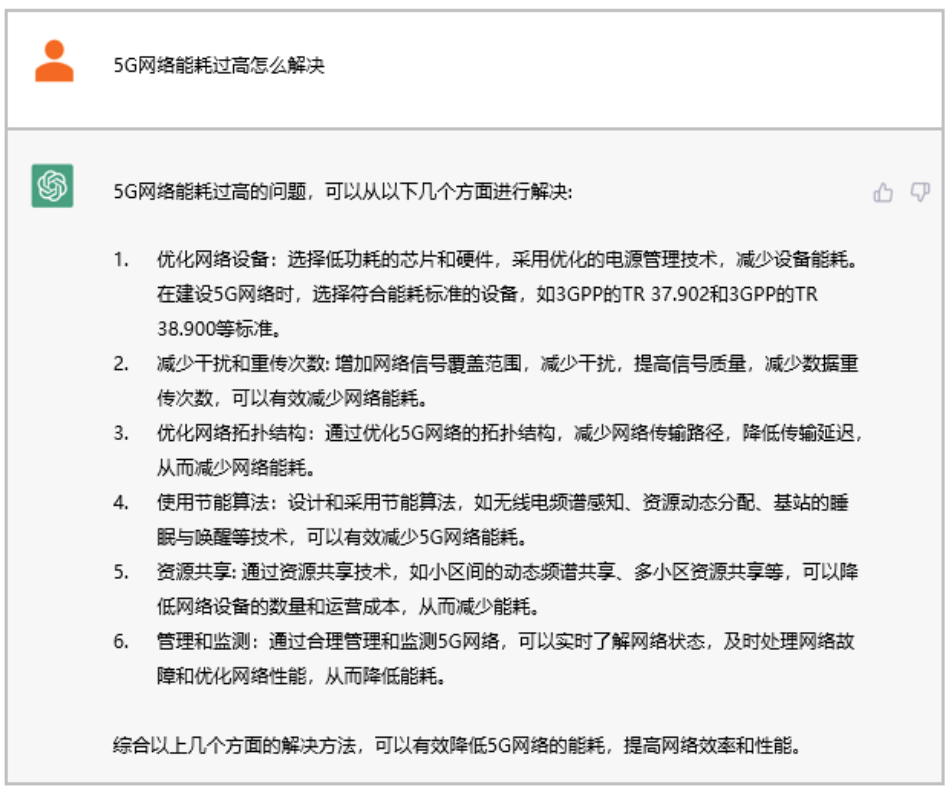

**图 3-44 GPT 大模型智能网络能耗优化应用 1**

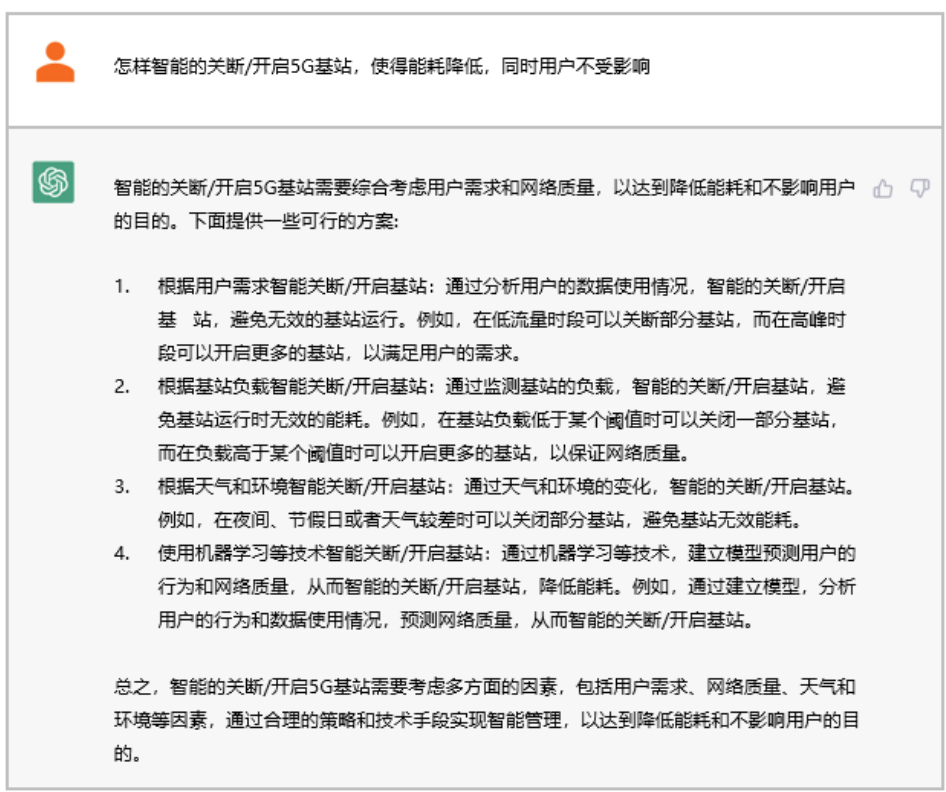

**图 3-45 GPT 大模型智能网络能耗优化应用 2**

场景二：优化专业问答及培训

随着运营商业务种类的激增，以及网络结构日趋复杂，5G 以及后续的网络中，网络优化工作，以及面向不同业务的端到端质量分析、优化的难度将成倍增加，需要运维工程师一专多能，具备较宽的知识面，GPT 大模型可以较为完美的代替目前的知识百科、专家经验库、甚至知识图谱等系统及技术。通过问答的方式展示云网优化的专业知识和优化案例，用户可以自由就关心的问题提问，并得到较为准确的回答。同时可以较为高效的成为专业技能培训系统，短时间快速提升员工技能。

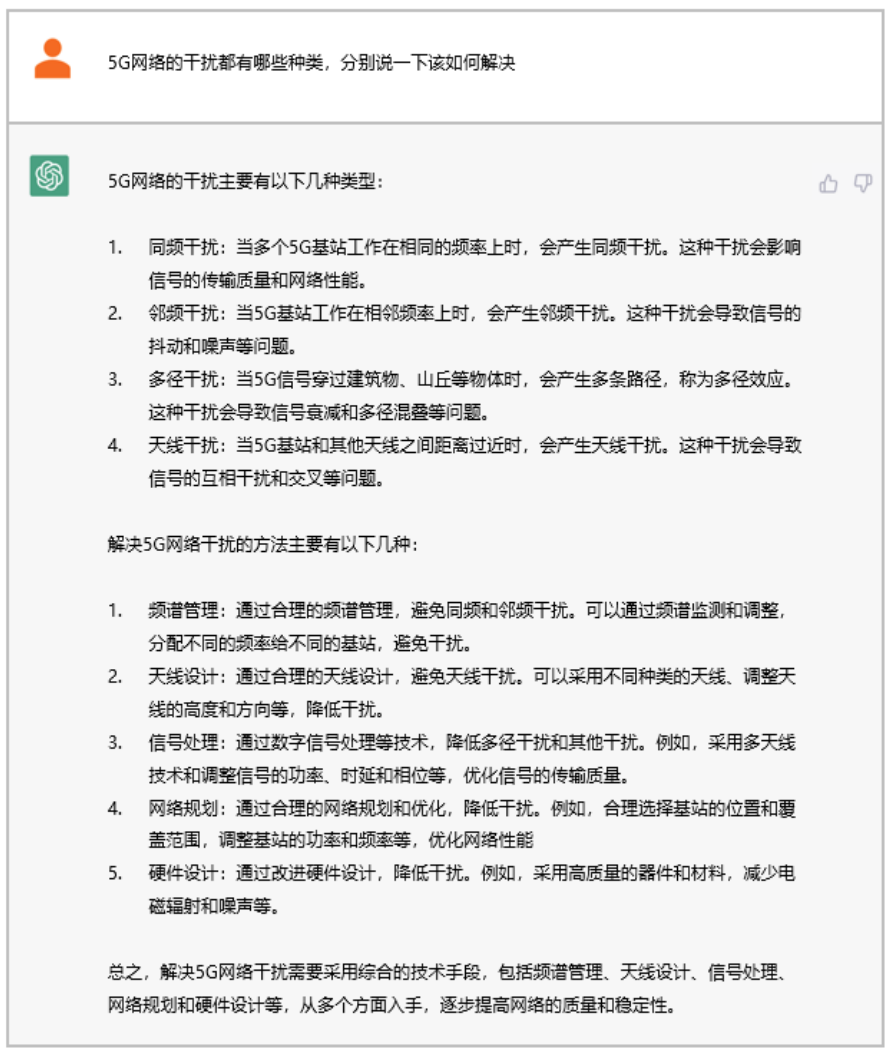

**图 3-46 GPT 大模型关于云网优化的专业知识储备**

## 3.2.5 智能网络运营

网络运营指支持市场部门设定的优化市场战略，支持产品设计、订单接收、流程分析以及业务在网络中的配置、激活、上线、扩缩容以及变更等全生命周期管理业务工作，同时也包含业务上线、变更带来的业务保障、端到端测试、质量监控、投诉预处理、客户服务、用户满意度保障等工作内容，保障网络资源对业务的诉求提供可靠的能力支撑。

新基建背景下，垂直行业对 5G 专网需求迫切，行业客户会出现建设需求井喷。结合后续建网技术瓶颈降低，建网模式出现多样化。随着 UPF 接口开放，企业对于用户面下沉，MEC 建设需求突增。针对业务上云及业务中台建设成为后续行业专网应用主要目标，这些新兴业务都需要运营商更加智能的动态高效的进行网络运营。GPT 大模型可以有效管理各类用户需求，并及时导入专网运营平台，提升用户感知。面向来自客户域的投诉工单，基于 GPT 大模型形成面向投诉的智能问答处理系统，识别用户投诉的语义、情感并形成处理策略，结合RPA 技术进入后续处理、关闭等自动化流转。

### 场景一：云网融合业务运营

随着网络演进，在多业务、跨领域的环境中形成了多样化的应用场景，网络业务调度开通环节多导致个性化业务需求匹配程度低、流程中存在人工操作的断点、业务端到端监控及可视化不足、定界定位能力不足等问题，难以满足用户"零等待"的诉求。通过 GPT 大模型知识引擎可以较为方便的实现云网用户意图的转译、理解和分析将业务订单自动转化为网络认知的 SLA 需求以工单的形式发送到各个网络环节。根据业务意图，自动识别并进行应用场景、内容、SLA 等规

格的配置并进行编排、创建、激活以及测试，完成云网业务的生成、部署和调测，确保上线质量。

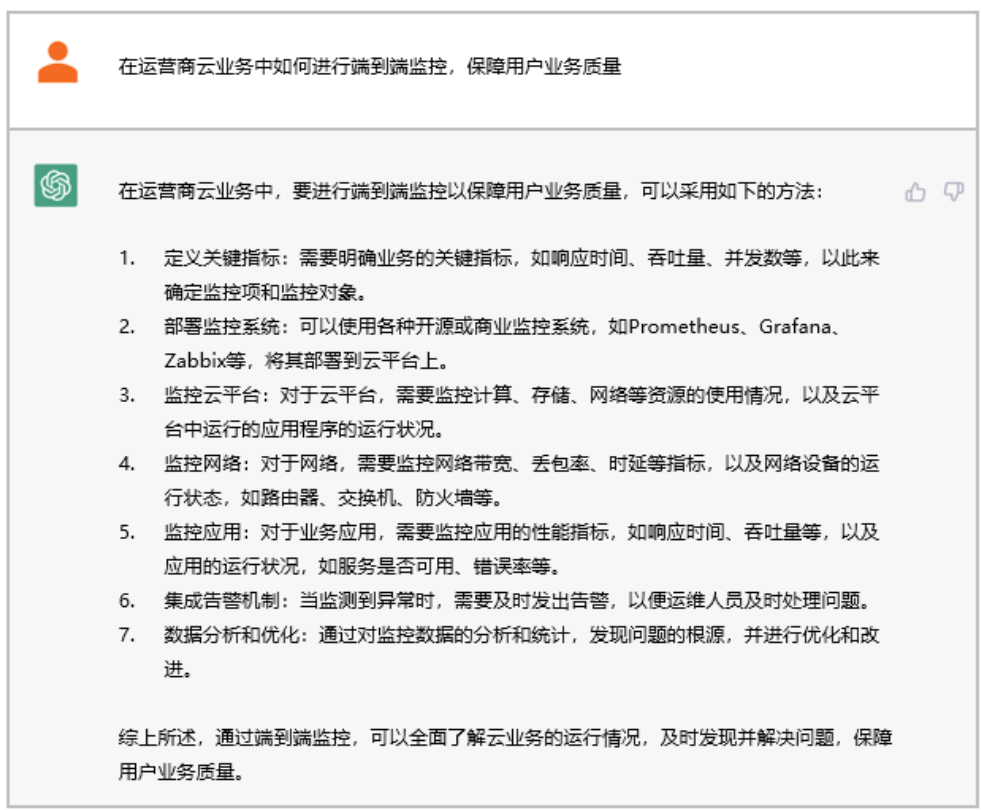

**图 3-47 GPT 大模型实现云网业务融合运营**

**场景二：网络投诉处理**

随着网络的变化以及业务运维模式的更新，网络投诉问题也将随之增多，且随着 NFV、切片等技术的应用，投诉问题的定界和处理的难度也必然增大，所以投诉预处理对自动化和智能化的需求更加强烈。现有的投诉处理流程着重在投诉发生后的一线应对，从分析和管理层面应对客户的原始投诉，减少后续一线运维团队的维护工作量。面向来自客户域的投诉工单，基于 GPT 大模型形成面向投诉的智能问答处理系统，识别用户投诉的语义、情感并形成处理策略，结合 RPA 技术进入后续处理、关闭等自动化流转。根据客户投诉内容安抚客户。对

于用户的投诉，识别出网络问题后给出较为精确的定界定位的分析。并进行针对

性的处理，如派单修复，套餐推送等，切实解决问题的投诉问题。

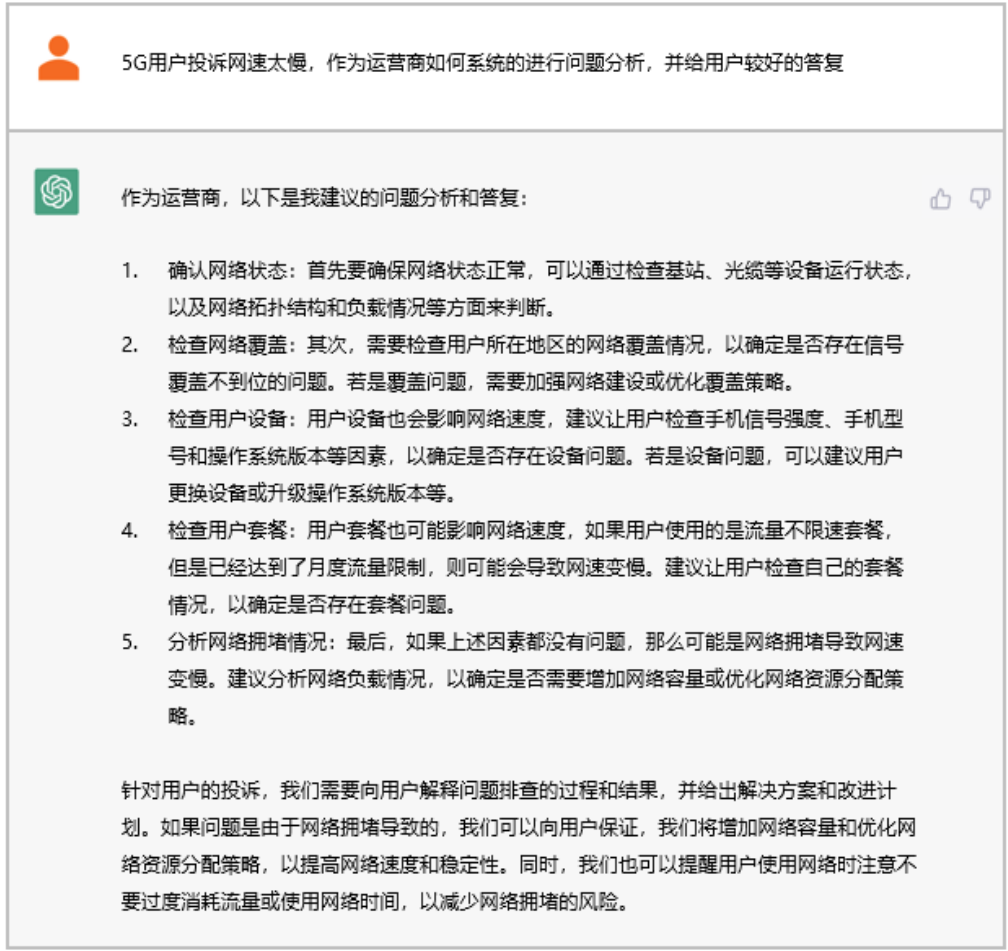

**图 3-48 GPT 大模型面向客户投诉的智能问答**

# 四. 构建通信业务 TelcoGPT

GPT 大模型具备通用知识，但缺少通信业的行业专有训练数据，难以满足专业场景。大模型深度赋能通信业，就需要构建通信业务 GPT。通信是为了实现互联互通，因此标准化是通信业的行业特征。标准化的体系结构为通信业 GPT 构建提供了相对高质量的数据，有助于行业知识构建。通信业拥有大规模的信息通信基础设施，也为行业 GPT 构建提供了算力基础优势。

## 4.1 通用大模型与通信业务之间的鸿沟

通用大模型基于海量互联网数据进行训练，缺乏通信业务专有数据，因此模型具备世界知识但不具备专业知识；通用大模型训练需要大规模算力，迭代更新慢，难以适应业务快速升级；通用大模型基于人类反馈进行了必要的文化和安全对齐，但面向行业应用缺少专业性质的人工校准；通用大模型以日常对话等通识场景为主。因此，通用大模型在面向通信业务应用时存在鸿沟。

### 通用大模型的数据要素更多为互联网数据

以 GPT-4 为例，其使用维基百科、书籍、期刊、Reddit 链接、Common Crawl 和其他数据集，并融入图表推理、物理考试、图像理解、论文总结、漫画图文等不同类型数据，在超过 4000 亿 token 上进行训练。

### 通用大模型的训练推理需要大规模算力

ChatGPT 训练使用了微软专门建设的 1 万个 V100 GPU 组成的高性能网络集群，总算力消耗约 3640PF-days (即每秒一千万亿次计算，运行 3640 个整日)。据报道，其前代 GPT-3 的训练单次成本就高达 460 万美元，预计 ChatGPT/GPT-

4 训练成本更高。训练大模型的成本大、时间周期较长，短期内难以有效地更新学习即时性的知识信息。大模型推理的并发性、推理实时性等也对部署成本提出了更高要求。

### 通用大模型校准做了必要的文化和安全对齐

GPT-4 训练团队组织了 50 多位来自长期从事 AI 生成一致性、网络安全、生物风险和国际安全等领域的专家来对模型进行对抗性测试；GPT-4 提出了一种基于安全规则的奖励模型，加入了安全相关的 RLHF 训练；采用细粒度个性化的风险信息响应，根据实际情况生成细粒度的健康建议。但大模型仍存在事实错误、有害响应等重要问题，对于某些确定的事实，GPT-4 给出的答案无法达到较高的置信度，对于常识性的知识问题不能高效地给出符合日常认知的答案。

### 通用大模型应用场景更偏向于通识场景

GPT 大模型能力覆盖广，但专业深度不足。模型复杂度，丰富性和通用性足够高，但对于某个专业领域方向的资深知识信息结构认知不足。在章节 2.4 给出了一系列通用大模型的应用场景，目前直接使用大模型效果比较好的场景主要聚焦于普通对话等通识场景。

### 最先进的 GPT-4 存在构建和应用问题

GPT-4 代表了 GPT 大模型的最先进形态，在 LLM 构建的理念和技术方面已经领先了一大步，但在使用 GPT-4 时，存在"技术开源开放问题"。本次 AI 浪潮的蓬勃发展依赖于开源开放，而 OpenAI 自 GPT-3 开始模型已不开源，chatGPT/GPT-4 更是不对中国大陆开放注册。同时，由于 OpenAI 不开展本地服务，按照中国相关法律法规和企业数据管理要求，存在跨国跨平台的"数据安全

/隐私保护"问题。因此，中国通信行业应用无法使用最先进的 GPT-4，必须考虑可私有化部署的大模型。

通信业务在 B 域、O 域、M 域以及融合域对大模型有多种场景需求，在数据要素、算力资源、文化和安全以及应用场景较通用有明显差异。

B 域主要实现对电信业务、电信资费、电信营销的管理，以及对客户的管理和服务的过程，它所包含的主要系统包括：计费系统、客服系统、帐务系统、结算系统以及经营分析系统等，业务场景包括开发、运维、营销、计费、客服、客户关系管理、商业智能等方面。

O 域主要是面向资源（网络、设备、计算系统）的后台支撑系统，包括专业网络管理系统、综合网络管理系统、资源管理系统、业务开通系统、服务保障系统等，为网络可靠、安全和稳定运行提供支撑手段。业务场景包括网络规划、网络部署、网络维护、网络优化、网络运营等方面。

M 域包括为支撑企业所需的所有非核心业务流程，内容涵盖制订公司战略和发展方向、企业风险管理、审计管理、公众宣传与形象管理、财务与资产管理、人力资源管理、知识与研发管理、股东与外部关系管理、采购管理、企业绩效评估、政府政策与法律等。业务场景包括财务分析、知识管理、采购管理等方面。

融合域指 BOM 三域融合，基于业务域、网络域、管理域的融合数据，利用通信人工智能技术实现跨领域业务与场景的融合智能。

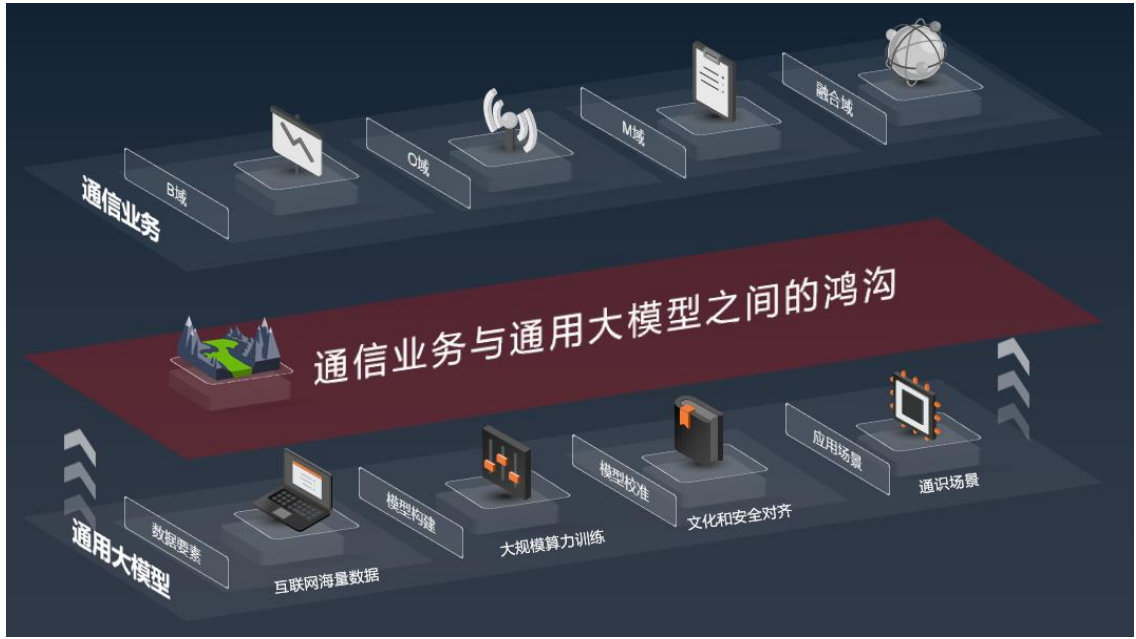

**图 4-1 通用大模型与通信业务之间的鸿沟**

# 4.2 TelcoGPT-基于通信认知增强弥合鸿沟

针对通用大模型与通信业务之间的鸿沟，我们首次提出了通信认知增强能力体系，构建通信业务 GPT（简称 TelcoGPT），实现 TelcoGPT 深度赋能通信行业。

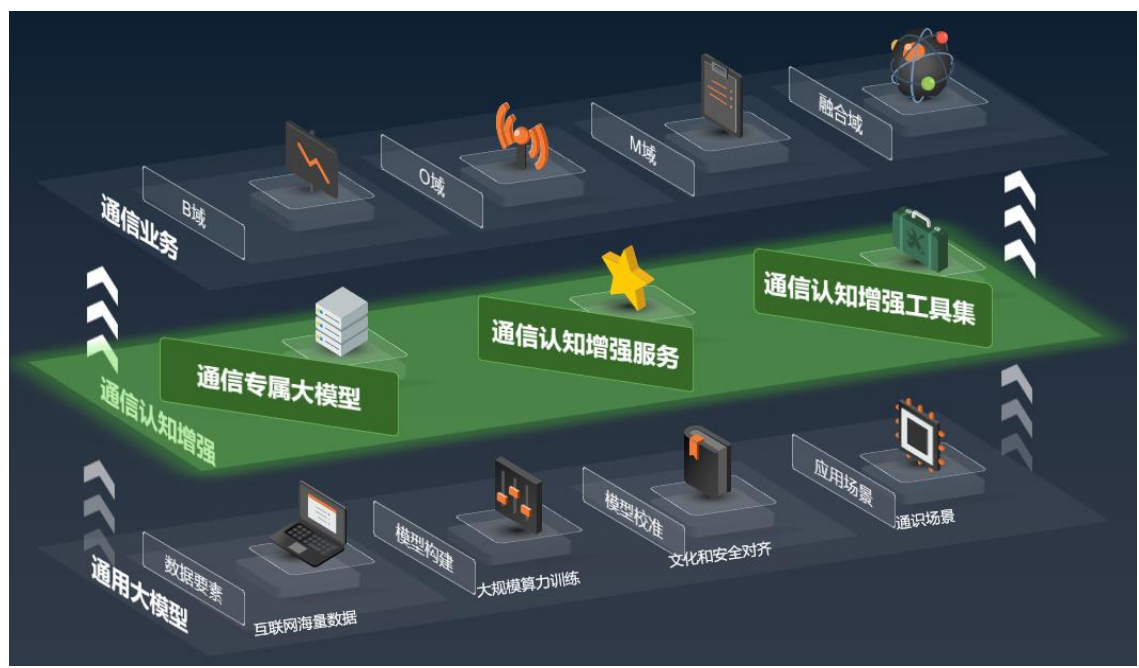

**图 4-2 通信认知增强弥合鸿沟**

TelcoGPT 主要包括三方面的通信业务增强能力：

### 通信专属大模型

由于通用大模型缺乏专业知识，需要使用通信专业数据训练通信专属大模型。通信专业数据，包括通信行业属性数据、客户服务对话数据、行业标准和法规文件、通信技术数据、运维装维日志数据、研发开发代码数据等等。通信专属大模型有两种实现方式，一种是基于海量互联网数据和通信专业数据从零开始训练自主大模型，另一种是基于开源或商用通用大模型进行通信专业数据精调。通信专

属大模型已经具备了较好的通信业务知识，但由于大模型迭代周期长、训练成本高，具有静态特征，在实际通信业务中仍需要结合通信认知进行动态增强。

### 通信认知增强服务

通信认知增强服务包括通信业务知识融合增强、通信业务大小模型融合进化以及通信业务决策执行能力融合增强。由于前述通信专属大模型仍然具有静态性特征，面向实际通信业务应用，通信专属大模型需要与外部知识进行融合增强，提供实时动态的知识更新；通信业务传统小模型具有场景化能力和专业效果，可以与大模型的泛化理解能力相结合，使小模型的单点智能面向通信业务全流程实现高等级智能化闭环；通信专属大模型还需要与外部业务工具相结合，使大模型具备通信业务的决策执行能力。

### 通信认知增强工具集

通信认知增强工具集将通信认知增强服务能力注入到通信业务工具中，提供通信业务开箱即用的工具集，如智能开发工具、智能运维工具等，便捷快速集成，最大化通信业务研发时效。

以 3GPP 标准知识问答为例，说明通信认知增强如何赋能通信业务。

1. 基于大模型纳管，使用开源大模型作为通用大模型；

2. 使用 1992.12 – 2023.6 期间 3GPP P1 – R17 的标准文档进行精调，得到具有 3GPP 标准知识的通信专属大模型；用户可以对截止到 R17 的 3GPP 标准进行知识问答，但无法获取在研的 R18 标准知识；

3. 使用 3GPP R18 在研文档进行融合增强，通过向量嵌入和检索的方式，获取最新标准知识并使用大模型提供自然的问答；

4、开发 3GPP-GPT 组件工具；

5、3GPP-GPT 组件嵌入 OSS 生产系统，为 OSS 提供便捷智能的标准知识问答。

## 4.3 TelcoGPT 如何构建？

TelcoGPT 通信认知增强体系自底向上包括通信专属大模型、通信认知增强服务、通信认知增强工具集三层，如下图所示。

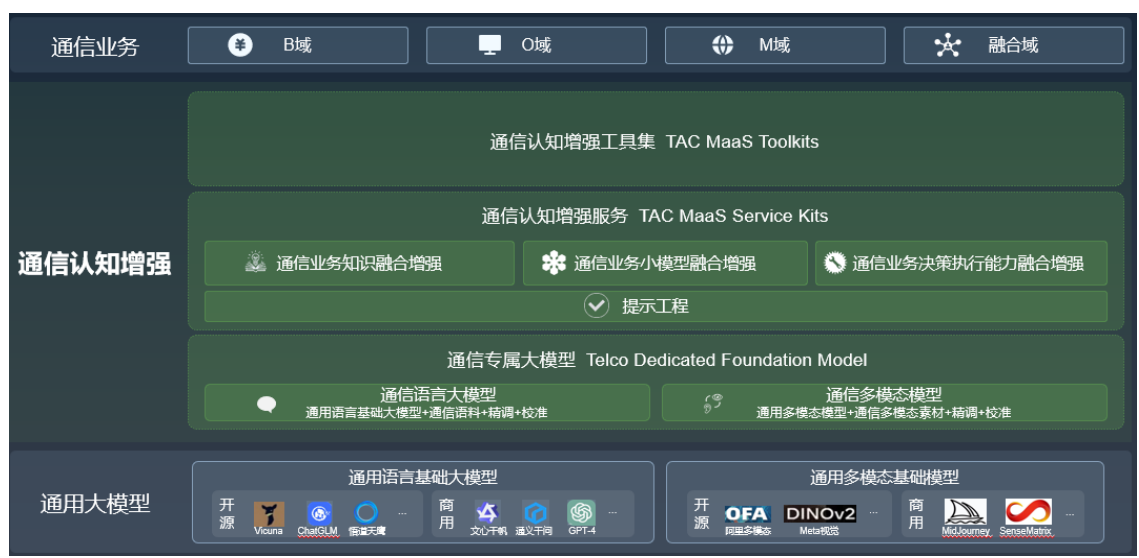

**图 4-3 TelcoGPT 通信认知增强能力体系**

通信专属大模型包括通信语言大模型和通信多模态模型。基于通用大模型纳管，包括基于开源/商用的通用语言基础大模型与通信语料，进行数据继续训练、指令精调、通信专家校准、评估部署等工作，构建通信语言大模型；基于开源/商用的通用多模态基础大模型和通信多模态素材，进行数据继续训练、指令精调、通信专家校准、评估部署等工作，构建通信多模态模型。

通信认知增强服务以提示工程为基础，实现通信业务知识融合增强、通信大小模型融合进化、通信业务决策执行能力融合增强。提示工程是通信认知与通信

专属大模型结合的基础手段，通过零样本提示、少样本提示、思维链、情景提示、角色扮演等手段，使通信专属大模型能够更好的理解、管理和执行任务。通信业务知识融合增强，通过知识向量嵌入、向量检索、知识图谱等，赋予 TelcoGPT 的动态实时性知识。通信大小模型融合进化，以大模型实现任务意图分析，并作为模型自主代理调用小模型的专业处理能力，实现全流程智能化闭环，最大程度减少人工干预。通信业务决策执行能力融合增强，以大模型为工具自主代理，实现通信业务工具的编排管理，进行业务决策分析和自主执行，打通模型与业务工具之间的壁垒，强化业务决策执行能力。

建设 TelcoGPT 通信认知增强能力体系，需要构建 1 个通信认知增强模型服务平台（Telco Augmented Cognition MaaS, TAC MaaS）、N 个通信认知增强工具集（TAC MaaS Toolkits）。

面向通信 B 域、O 域、M 域和 BOM 融合域，打造 N 个通信认知增强工具集，快速赋能通信各领域业务。如，面向 B 域的经分增强分析 chatBI 工具、智能运维 OpsGPT 工具；面向 O 域的自智网络智能副驾 AN Co-Pilot 工具；面向 M 域的 OA GPT 办公工具；面向融合域的 RPAGPT 流程机器人工具、CodeGPT 低代码工具、Chatbots 交互工具等。

模型即服务（MaaS）是链接基础大模型和行业领域任务迁移过程重要环节，是实现更好专业化服务的关键路径。TelcoGPT 的通信专属大模型和通信认知增强服务由通信认知增强模型服务平台 TAC MaaS 构建。TAC MaaS 围绕通信业务 TelcoGPT，提供一系列优化过的基础服务和原子能力，统一为各相关产品提供低门槛、轻量化、可组装的技术组件，并通过 API 接口进行赋能。TAC MaaS 包括模型推理平台和训练平台。TAC MaaS 推理平台提供 MaaS 开箱即用的功能及轻量化承载，可以通过提供 API 或作为子模块的方式和其他产品集成。TAC

MaaS 训练平台包括领域适配训练、精调、对齐和模型压缩等较高资源消耗型服务，可以与推理平台组装或分别交付。

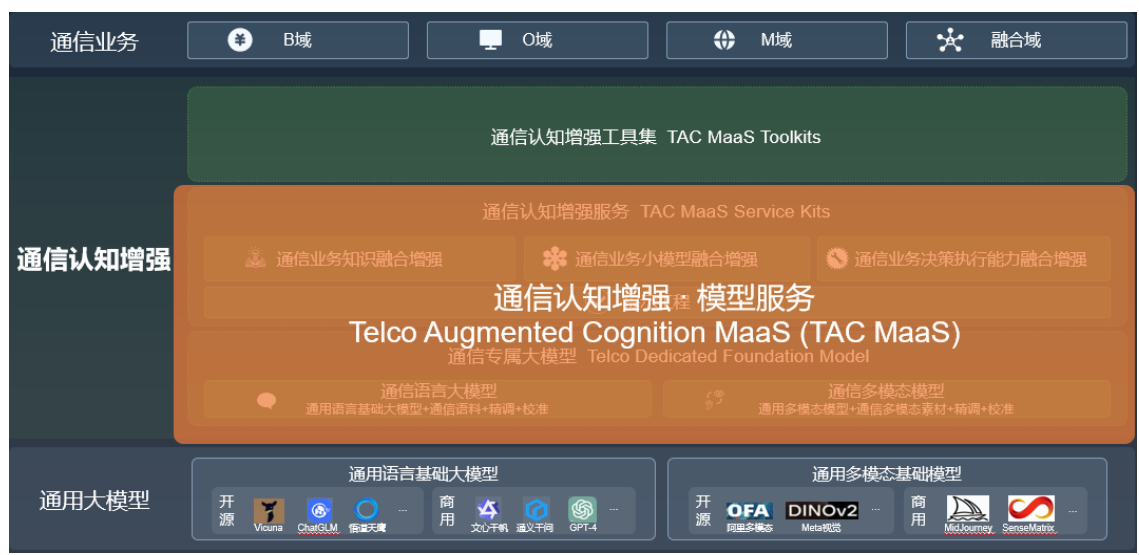

图 4-4通信认知增强模型服务 TAC MaaS

# 4.4 TelcoGPT 实践

通信行业数据要素和生产资料积累，以及 BOM 域专业通信服务场景沉淀，为 TelcoGPT 通信认知增强提供了核心能力，通过研发通信认知增强模型服务平台（TAC MaaS）与通信认知增强工具集（TAC MaaS Toolkits），快速推动 TelcoGPT 实践落地。以下介绍典型实践案例。

## 4.4.1 TAC MaaS

通信认知增强模型服务平台分为推理平台和训练平台。训练平台提供较高资源消耗型服务，支持 AI 模型预训练、监督精调、人机语义对齐、评估优化与部署发布等。推理平台提供 MaaS 开箱即用的功能及轻量化承载，提供大模型的继续训练、指令微调、对齐、评估部署、提示工程，实现模型纳管/托管，向量嵌入管理、模型推理运算和模型治理等，为通信认知增强服务层提供智能模型编排。

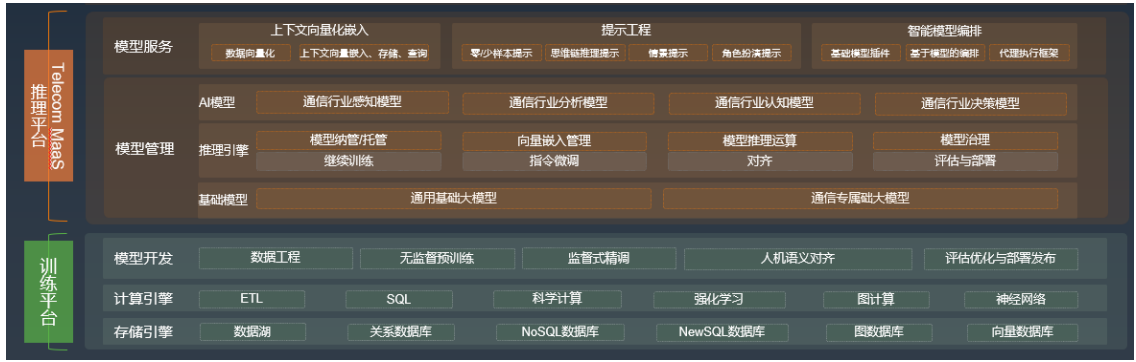

**图 4-5 通信认知增强模型服务功能**

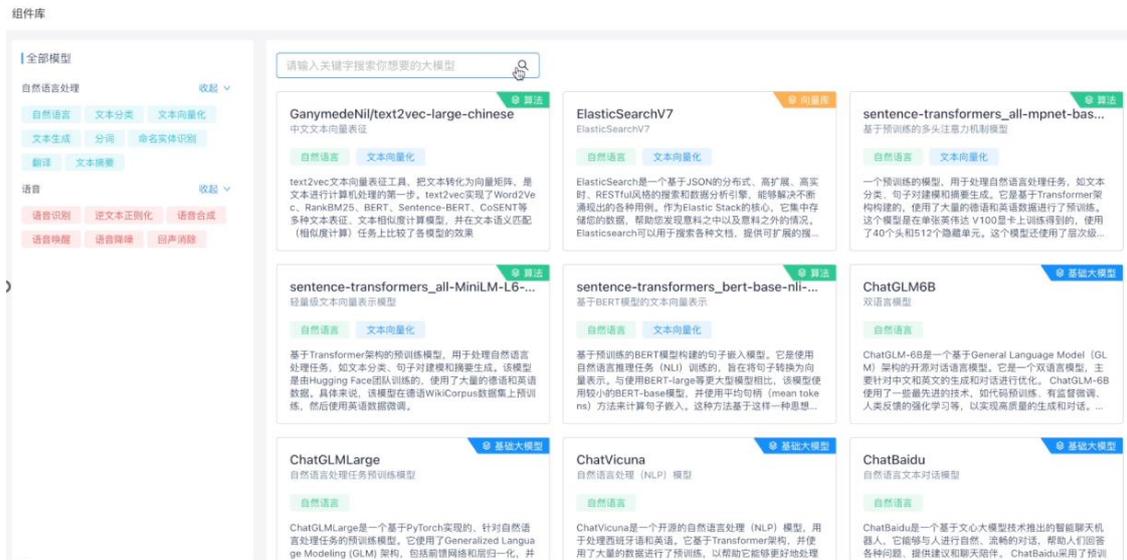

**图 4-6 通信认知增强模型服务平台**

## 4.4.2 TelcoGPT-CodeGPT

　　CodeGPT 是基于 TelcoGPT 提供的大模型在应用开发领域的实践。运用 TelcoGPT，将复杂的代码编写过程进行智能化处理，能够大大提高代码开发的效率和质量。当前，CodeGPT 在应用开发领域的主要实践有代码补全、代码注释和用例生成。

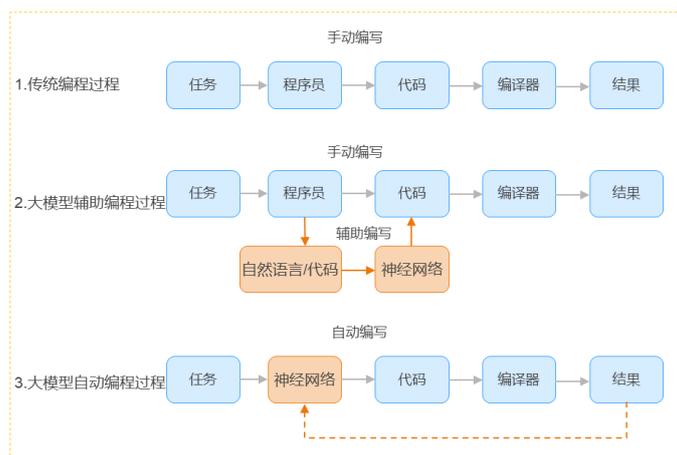

**图 4-7 引入大模型后的代码开发过程**

- 代码补全：传统的代码编写过程中，程序员往往需要记住大量的代码库和函数。CodeGPT 可以实时地对代码进行智能补全，减轻了程序员的记忆负担，同时也大大提高了代码编写的效率。程序员只需要输入部分代码，CodeGPT 就能提供相应的代码补全建议。此外，还可以基于自然语言的功能描述生成代码片段。

```java
import java.util.Map;

/**
 * @author Juergen Hoeller
 * @author Mark Fisher
 * @author Ken Krebs
 * @author Arjen Poutsma
 */
@Controller
class VetController {

    private final VetRepository vets;

    public VetController(VetRepository clinicService) {
        this.vets = clinicService;
    }

    @GetMapping("/vets.html")
    public String showVetList(Map<String, Object> model) {
        // Here we are returning an object of type 'Vets' rather than a collection of Vet
        // objects so it is simpler for Object-Xml mapping
        Vets vets = new Vets();
        vets.getVetList().addAll(this.vets.findAll());
        model.put("vets", vets);
        return "vets/vetList";
    }

    @GetMapping( "/vets" )
    public @ResponseBody Vets showResourcesVetList() {
        // Here we are returning an object of type 'Vets' rather than a collection of Vet
        // objects so it is simpler for Json/Object mapping
        Vets vets = new Vets();
        vets.
        vets.getVetList().addAll(this.vets.findAll());
        return vets;
    }
}
```

**图 4-8 代码行级补全**

**图 4-9 代码片段补全**

● 代码注释：代码注释是保证代码可读性的重要方式，但是编写注释的过程却往往耗时且乏味。CodeGPT 可以根据代码的内容和逻辑，自动生成相应的注释，大大提升了代码的可读性，同时也节省了程序员编写注释的时间。

**图 4-10 代码注释生成**

● 用例生成：在软件开发过程中，用例测试是保证代码质量的重要手段。然而，编写用例的过程往往需要大量的时间和精力。CodeGPT 可以根

据代码的功能和需求，自动地生成相应的用例，从而大大提高了测试的效率，同时也保证了代码的质量。

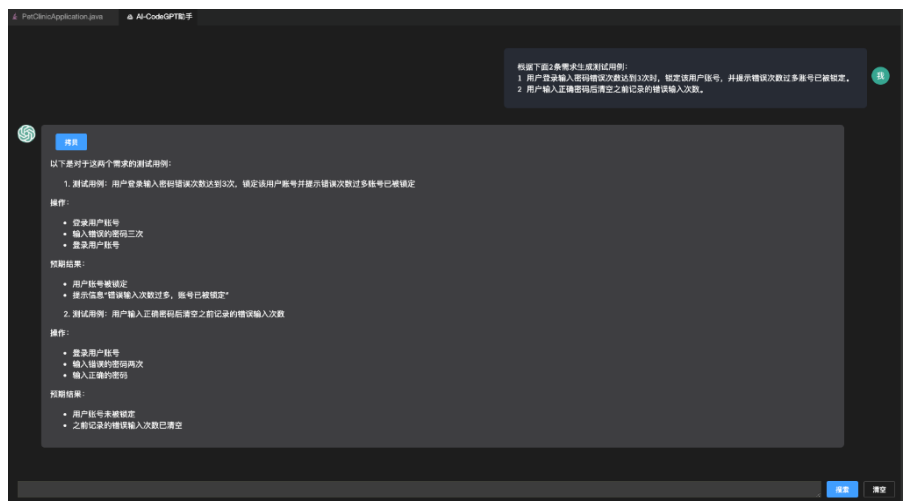

**图 4-11 CodeGPT 生成测试用例**

CodeGPT 不仅仅是一个智能化的代码辅助开发工具，更是一种全新的开发理念。它将大模型的人工智能技术与代码开发完美融合，为程序员提供了一种更加高效、便捷的开发方式，为软件工程领域带来了革命性的创新。在未来，随着 AI 大模型的更新迭代，将会对软件开发生命周期的各个环节赋能，完成从 CodeGPT 向 DevOpsGPT 的演进。

### 4.4.3 TelcoGPT-OpsGPT

TelcoGPT-OpsGPT 基于通用大模型的底座，通过自研的大模型轻量化微调能力，提供了面向运维知识问答和智能工单分析两个应用场景的实践。

OpsGPT可与本地私域运维知识融合实现运维知识智能问答，首先通过拖拽的方式将本地运维知识文档（文档类型包括不限于 pdf、word、csv 等）进行导入，OpsGPT对文档内容进行预处理并自动分段、切片，实现与大模型运维知识的融合增强。当用户问询运维知识时，OpsGPT可以基于对导入的知识文档的理

解生成答案，除了生成答案之外，还可以提供答案的来源以及答案参考的知识清单，实现答案的可追溯。

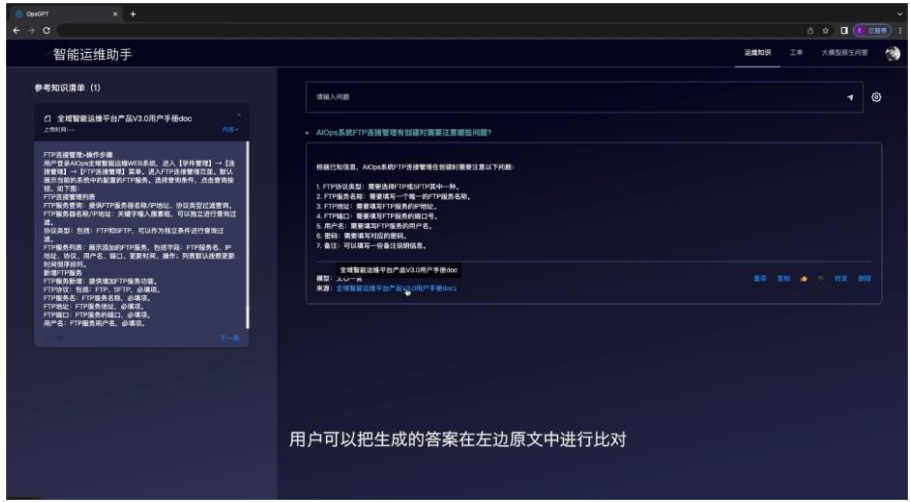

**图 4-12 运维知识智能问答**

进一步的，OpsGPT 也可以基于历史工单数据实现相似工单去重、工单自动填充、推荐工单解决方案等能力。首先将历史工单数据上传导入，通过对工单数据的预处理分析，实现工单数据与大模型的融合。针对新的工单，可以通过 OpsGPT 搜索最相似的故障工单的内容进行语义理解和答案生成，同样做到工单答案的可溯源。

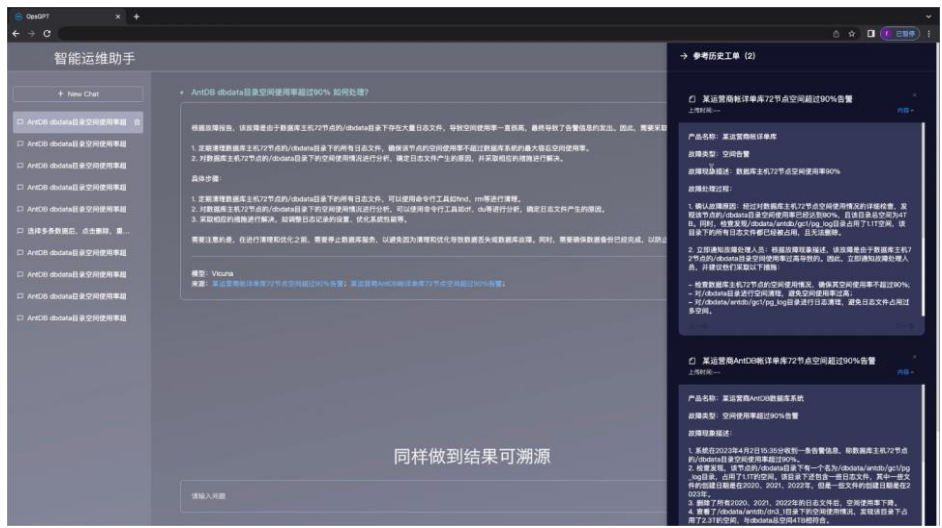

**图 4-13 智能工单分析**

### 4.4.4 TelcoGPT-ChatBI

利用增强分析技术，在 BI 工具中新增 TelcoGPT-ChatBI 功能，面向业务用户提供会话式的指标检索和分析服务，支持用户通过自然语言快速检索业务指标，并对指标进行深度剖析。

ChatBI 主要功能包含：指标检索、指标关联分析、报告生成和分享。

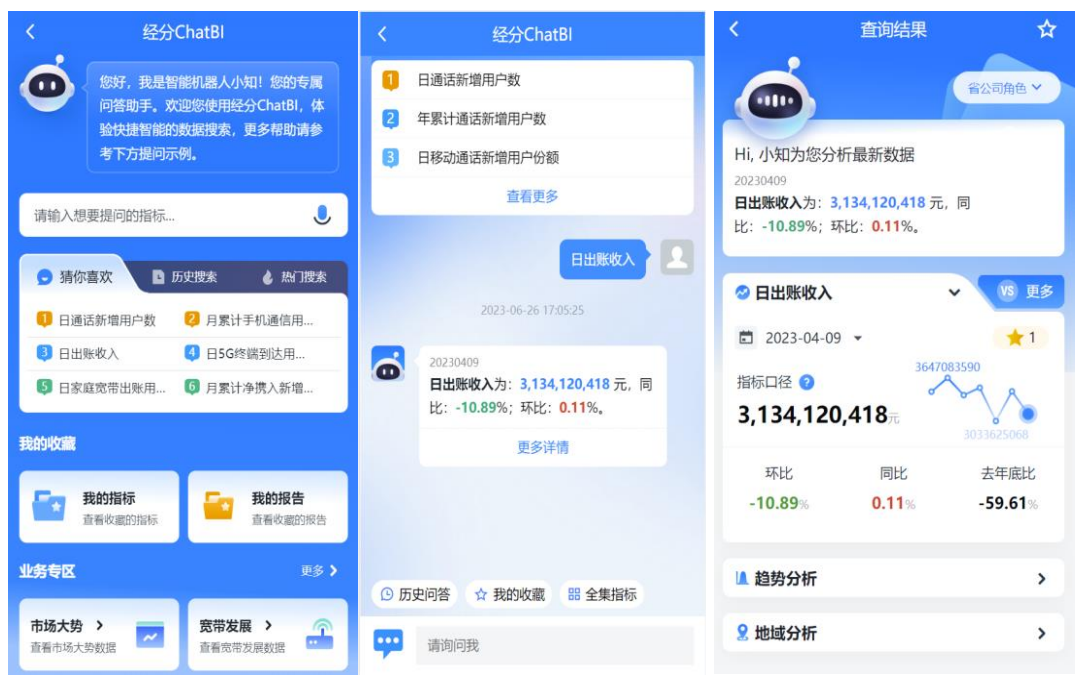

**图 4-14 精确指标检索与分析**

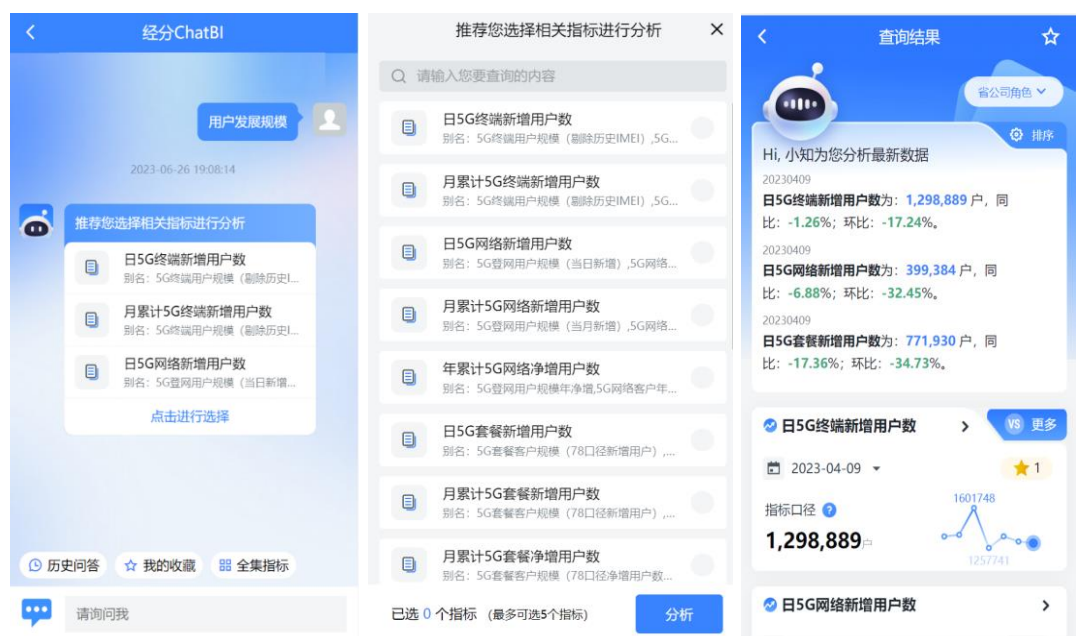

图 4-15 泛化指标检索与多指标关联分析

ChatBI 帮助经营管理者轻松掌握关键业务指标，摆脱对专业分析团队的依赖，经营分析在时效性和自主性方面得到了空前提升。对专业分析团队而言，大量来自事务性工作的压力得以释放，团队价值在更高级别的数据需求上得以体现。

## 4.4.5 TelcoGPT-Chatbots

TelcoGPT-Chatbots 是通信行业级、覆盖 B、O、M、S 多个领域的问答平台，支持 FAQ、KBQA、多轮对话等多种会话引擎，支撑包括智能客服、数字党建、融合推荐、智能运维等应用场景。

大模型为 Chatbots 的发展注入了新的技术力量。一方面，将大模型能力融入会话技能，发挥大模型在语义理解和表达方面的优势，更好地理解用户意图，灵活地排序、筛选、组织生成答案，从而提升问答的准确率。另一方面，大模型赋予开放域的对话能力，提升交互感知。除此以外，Chatbots 通过知识图谱补全事实知识，弥补大模型在领域知识储备、数据推理方面存在的问题。

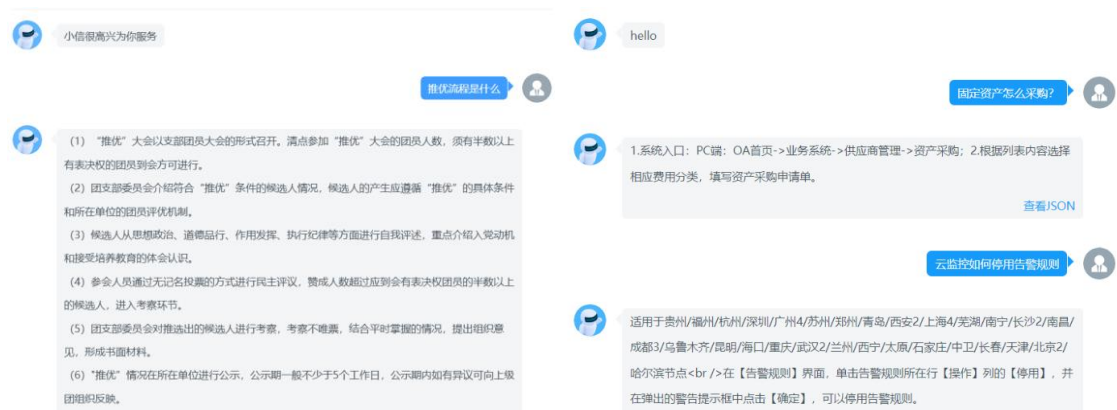

图 4-16 Chatbots FAQ 问答

### 4.4.6 TelcoGPT-AN Co-Pilot

TelcoGPT-AN Co-Pilot 是以通信认知增强模型服务为基础底座，帮助行业提升自智网络等级的运维辅助工具集。借助大模型的理解和生成能力，AN Co-Pilot 能够覆盖网络规建维优营全生命周期的主要场景，赋能自智网络向高阶演进。目前实现功能包括：

AN Co-Pilot 网络实时分析：利用 GPT 大模型意图理解、外部能力调用和信息总结生成能力赋能网络数据平台，为用户提供灵活智能的数据处理、数据洞察等功能，同时通过 API 调用保证数据隐私。用户可以使用自然语言查询，AN Co-Pilot 自动解析查询意图，通过标准的 API 接口和数据格式，提供相关的数据集、数据源的检索能力。大大提高了数据发现和获取的效率，帮助用户快速找到所需的数据资源。

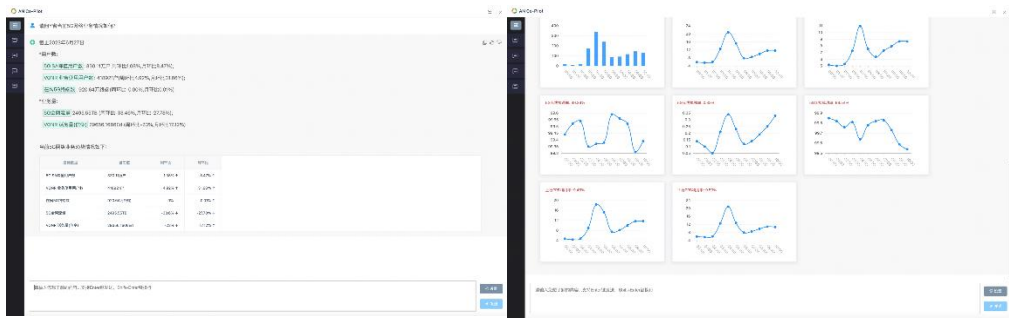

**图 4-17 网络数据查询及分析**

AN Co-Pilot 投诉助手：利用 GPT 大模型语言理解能力结合无线网络分析功能，为投诉处理人员提供网络投诉智能处理辅助。投诉人员可以通过自然语言提问，大模型将自动识别语义，利用前期知识，实现标准专业的投诉应答、处理方法。无线网络分析功能通过大模型反馈的应答进一步实现投诉相关数据查询呈现、

提供投诉基本处理方案。后续会进一步将相应模块解耦被大模型统一调用，最终

实现投诉问题自动闭环处理。

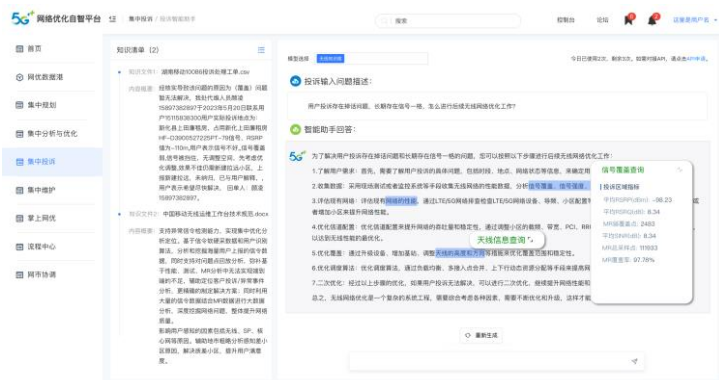

**图 4-18 网络投诉助手及深度分析**

- AN Co-Pilot 网络设备配置辅助：IP 网控制能够基于专用大模型自动生成

  适配不同设备商的南向网络配置报文，并自动调用下发。有望解决网络

  业务配置自动化开通流程里的难题。

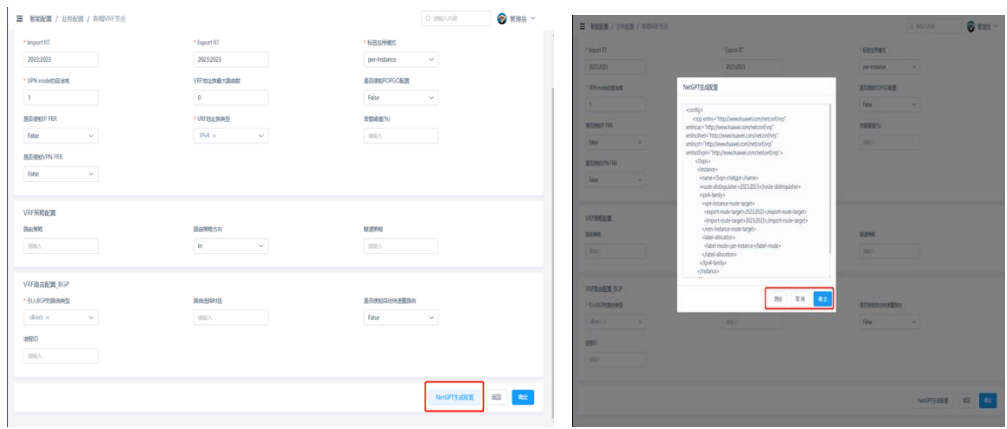

**图 4-19 自动生成南向网络配置报文并检测下发**

## 4.4.7 AIGC 平台

AIGC 平台面向多模态内容生成，通过预集成文本生成、图像生成、数字人

形象生成、交互动作生成、语音生成、素材生成、3D 驱动生成等生成式 AI 工具

集，可对外提供虚拟人生产、虚拟商品生产、虚拟场景制作、文案/图片/语音生产等智能内容生产能力。这些能力将成为公司"数智化全栈能力"基座的重要组成。

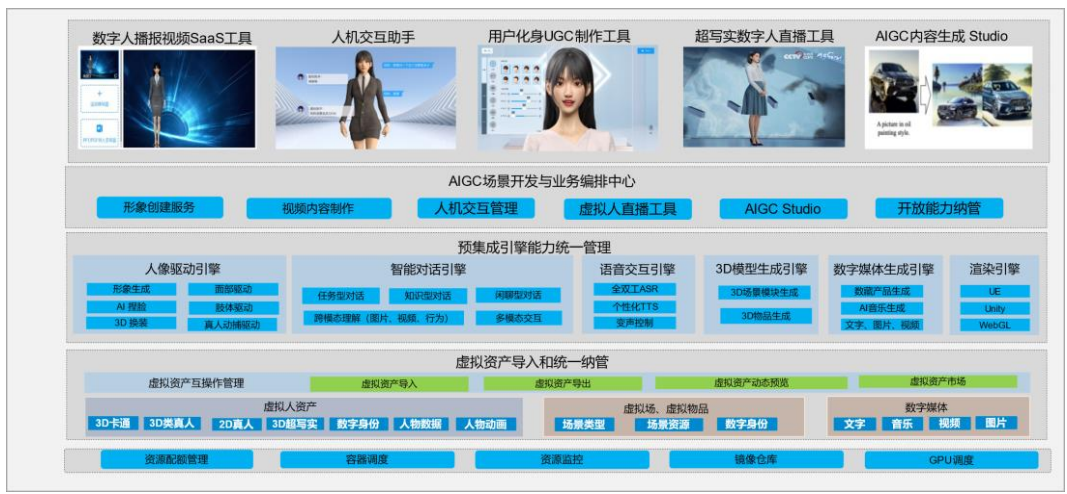

**图 4-20 AIGC 平台功能架构图**

AIGC 平台是有效实现多种生成式 AI 技术能力、多类虚拟资产复用和共享的 AIGC 平台，支持通过标准化的方式，有效地纳管客户已有的三方能力和资产。以上功能架构图分五个层次，描述了各类功能的解耦和组织方式。

● 弹性计算层：基于云原生的方式提供资源配额管理、容器调度、资源监控、镜像仓库管理、异构 AI 计算资源调度的能力，为上层提供高安全、高效率的计算资源管理；

● 资产层：提供了一整套标准化的，面向各类虚拟资产的数据文件格式定义，包括各类 2D、3D 虚拟人资产；各类虚拟场景、虚拟物品的资源和资产；文字、音乐、图片、视频在内的各种数字媒体资产。同时提供了虚拟资产互操作管理，各类虚拟资产导入导出，虚拟资产市场以及各类

虚拟资产动态预览的能力。通过这种资产层开放和标准化的方式，能够有效地帮助客户扩充资产库，保障客户的已有资产投资；

- 能力层：研发并预集成了各类上层业务编排和应用所需要的，业界领先的原子能力，包括各类面向驱动、对话、语音、3D 建模、数字媒体生成和渲染引擎等，并可通过开放 API 的方式对上层提供开箱即用的服务能力。同时可以有效支持对 API 的权限管控、访问认证、流量控制，以及对真实服务地址的屏蔽等；

- 业务编排层：业务编排层可以通过低代码的方式，面向特定 AIGC 应用场景的需要，灵活组装和调用底层的各项能力，实现应用的快速打造。目前已经预置的业务模板包括形象创建服务、视频内容制作、人机交互管理、虚拟人直播工具。此外在业务编排层还支持开放能力纳管功能，方便用户在进行编排时灵活引入自身已有的或成熟的第 3 方基础能力；

- 应用层：当前重点打造的开箱即用的样板间主要有数字人播报视频 SaaS 工具、具备 3D 虚拟人形象的人机交互助手、用户可以自服务订制的数字化身制作工具、超写实数字人直播间以及 AIGC 内容生成 Studio。更多更丰富的 AIGC 应用场景还将进一步陆续推出。

# 五. 总结和展望

以 GPT 大模型为代表的 AIGC 迈出了通用人工智能的第一步，开启了 AI 新代际，打开了诸多行业变革的想象天窗。AIGC 大模型将使机器可以大规模参与知识类和创造性工作，成为继移动互联网之后未来最大的技术平台。ChatGPT 首次实现了人与机器之间自然的交互界面，GPT-4 具备更高的交互理解力和准确率。通信行业作为人与人、人与自然、人与机器之间进行信息交流与传递的行业，天然具有与 GPT 大模型的结合力。基于白皮书的分析可以看到，AIGC 可作为先进生产力工具改进信息通信服务能力，从各个层面深度赋能通信业，构建为千行百业数字化转型赋能的智慧通信。与此同时，AIGC 也将推动通信业从人与人之间的沟通交流扩展到人与机器之间的沟通交流，支撑人机物三元融合。从内涵和外延看，通信行业必须积极拥抱大模型。

算力发展是人工智能的核心要素之一。大模型、基础模型在开启人工智能技术新代际和应用新范式的条件下，对于算力的庞大需求将形成巨大挑战，也必将带动算力基础设施的发展。算力建设将是决定人工智能赋能各行各业的广度与深度的重要因素。

通信行业运营商作为国家信息通信基础设施建设者和运营者，也是让大模型实现普惠智能服务的承载者。随着摩尔定律的逐步放缓，数据中心、边缘节点、智能终端等单点算力正在面临瓶颈。以 5G 为代表的通信基础设施和以云计算为代表的信息基础设施的高质量发展，为网络与算力融合提供了基础条件。在算力作为数字经济的关键生产力，已形成新的全球竞争格局的状态下，中国通信行业率先提出了算力网络技术并主导了这一新型信息通信基础设施的发展建设，在产业方面已初步形成算力网络软硬件基础设施及平台与应用服务的上中下游完整产

业链。算力网络的内涵就是要以网调算、以网融算、以网强算，通过网络对算力的感知、触达、编排、调度能力，在算网拓扑的任何一个接入点，为用户的任何计算任务可灵活、实时、智能匹配并调用最优的算力资源，从而实现云-边-端任何时间与任何地点的多方算力需求。基于算力网络对先进算力进行集中化建设使用和集约化发展，提升大模型构建和使用的算力能效，为大模型普惠智能服务提供高效基础支撑，也是在先进半导体面临严格管制的情况下通信行业节约算力投资、提升计算产业安全的重要举措。

面向 2030 的通信系统，6G 的竞争已经开始。中国 IMT-2030（6G）推进组提出"万物智联、数字孪生"的 6G 总体愿景。6G 将与先进计算、大数据、人工智能、区块链等信息技术交叉融合，成为服务生活、赋能生产、绿色发展的基本要素。6G 将构建人机物智慧互联、智能体高效互通的新型网络，具备智慧内生、多维感知、数字孪生、安全内生等新功能；将实现物理世界人与人、人与物、物与物的高效智能互联，打造泛在精细、实时可信、有机整合的数字世界。可以展望，通信认知增强的 TelcoGPT 将实现 GPT 大模型与通信的深度融合，加快 6G 智慧内生、全自智网络的能力演进，即 AI4Network；6G 万物智联，将加快普惠智能的实现，即 Network4AI。

# 参考文献